\documentclass{article}

\PassOptionsToPackage{numbers, compress}{natbib}
 \usepackage[preprint]{neurips_2026}

\usepackage[utf8]{inputenc} % allow utf-8 input
\usepackage[T1]{fontenc}    % use 8-bit T1 fonts
\usepackage{hyperref}       % hyperlinks
\usepackage{url}            % simple URL typesetting
\usepackage{booktabs}       % professional-quality tables
\usepackage{amsfonts}       % blackboard math symbols
\usepackage{nicefrac}       % compact symbols for 1/2, etc.
\usepackage{microtype}      % microtypography

\usepackage{graphicx}%
\usepackage{multirow}%
\usepackage{amsmath,amssymb,amsfonts}%
\usepackage{amsthm}%
\usepackage{mathrsfs}%
\usepackage[title]{appendix}%
\usepackage[usenames,dvipsnames]{xcolor}%
\usepackage{soul}
\usepackage{textcomp}%
\usepackage{manyfoot}%
\usepackage{algorithm}%
\usepackage{algorithmicx}%
\usepackage{algpseudocode}%
\usepackage{listings}%
\usepackage{latexcolors}

\newcommand{\mycolorbox}[2]{%
\sethlcolor{#1}\hl{#2}%
}

\definecolor{pdred}{HTML}{d62728}
\definecolor{pdorange}{HTML}{ff7f0e}
\definecolor{pdgreen}{HTML}{2ca02c}

\newcommand{\thickhline}{\specialrule{0.15em}{0.20em}{0.50em}}

\let\cline\cmidrule

\title{The Illusion of  AI Expertise Under Uncertainty: A Probabilistic  Paradigm to Interpret Results Under  Ground Truth Variation}

\author{%
  Aparna Elangovan \\
  Independent Researcher\\
  \And
Lei Xu \\
 Independent Researcher\\
 \AND
Mahsa Elyasi \thanks{Work done outside Amazon} \\
  Amazon \\
 \And
 Ismail Akdulum \thanks{Equal Contributions} \\
   Gazi University \\
   \\
\And
Mehmet Aksakal $^\dagger$ \\
   University of Colorado \\
\And
Enes Gurun  $^\dagger$ \\
   Samsun University \\
\And
Brian Hur \\
   The University Of Melbourne \\
   \And
Saab Mansour *\\
   Amazon \\
  \And
Ravid Shwartz Ziv\\
   New York University \\
 \And \And
Karin Verspoor\\
   RMIT University \\
\And
Dan Roth\\
   University of Pennsylvania \\
}

\begin{document}

\maketitle

\begin{abstract}

Benchmarking the  capabilities of AI systems, including Large Language Models (LLMs) and Vision Models, typically ignores the impact of uncertainty in the underlying ground truth answers from experts. This ambiguity is not just limited to human preferences, but is also consequential  even in safety critical domains such as medicine where uncertainty is  pervasive. In this paper, we introduce a probabilistic paradigm to theoretically explain how  \textit{high certainty in ground truth answers is almost always necessary for even an expert to achieve high scores}. In contrast, in datasets with high variation in ground truth answers, the expert performance drops substantially  resulting in  \textit{almost no difference between a random labeller and an expert}. This characteristic also manifests when comparing models, where uncertainty  obfuscates differences between poor and high performing  models. Therefore, ignoring uncertainty in ground truth evaluation data can result in the misleading conclusion that a non-expert has similar performance to that of an expert.  Given ground truth answer variability and commonly used F1 and accuracy metrics, we bring forth a simple and explainable probabilistic \textit{coin toss} model to estimate \textit{expected accuracy} and \textit{expected F1}  and compare the estimated scores with those that  experts   (human or model)  achieve across multiple domains -- including radiology and  models such as GPT-5 and Gemini-3.

The results lead to the recommendation that
stratification by the probability of the ground truth answer becomes critical when expert performance is relatively low. Under stratified evaluation, performance comparison becomes more reliable in high certainty bins, mitigating the effect of the key confounding factor --- uncertainty.

\end{abstract}

\section{Introduction}\label{sec1}
Artificial Intelligence (AI) systems, including large language models (LLMs), are typically evaluated against ground truth (GT) human labels under the simplistic assumption that  a single correct  or a preferred answer can be associated with a test question. Conventional evaluation
simply compares the performance of an AI model or algorithm against that of experts, e.g.\  an individual doctor or a consensus gold standard, to benchmark the relative capabilities of the model, ignoring the impact of uncertainty in the underlying ground truth labels. In this paper, through a probabilistic paradigm, we demonstrate why  datasets that have
variation in ground truth answers show two
characteristics:
\textbf{(a)}  high certainty in ground truth answers  is almost always necessary for an expert to achieve high scores on an evaluation set, as measured by typical ground truth uncertainty agnostic metrics such as  accuracy or F1 scores, and
\textbf{(b)} when certainty in ground truth answers is low (or equivalently, uncertainty is high),
there may not be any difference between a random labeller and an expert.

Humans can rely on heuristics or mental shortcuts to make   decisions which are often noisy under uncertainty \cite{kahneman_noise_2021, kahneman1982variants}.  Uncertainty is pervasive even in safety critical applications such as in medicine, where uncertainty remains a hidden component in  the diagnosis, treatment, and prognosis of disease \cite{Lichtstein2023, 10.1001/archinte.168.7.741, Kim2018-wz, Han2021-la, gawande2003complications, Esty_2024, Bhise2018-ku}, due to a range of factors   from lack of scientific data for certain conditions to lack of time \cite{Kim2018-wz}.
Yet, when comparing AI models to experts such as doctors, uncertainty is often ignored.  %Medical uncertainty can be the result of many factors, ranging from incomplete scientific data for certain conditions,  lack of comprehensive information about a patient, complexity of clinical information, the probabilistic nature of certain  outcomes, practical constraints such as time and cost \cite{Kim2018-wz, Helou2020-uv,Bhise2018-ku} and cognitive biases  \cite{Saposnik2016}.  %As an example, when patients present with   undifferentiated symptoms it is difficult for clinicians to identify a satisfactory explanation of the patient’s  problem, especially under finite pragmatic constraints such as  time and cost \cite{Helou2020-uv}. Despite the role of uncertainty in  medicine, popular data sets such as MedQA \cite{jin2020disease} have a single definitive answer associated with each question in the dataset, which are then used to evaluate LLMs \cite{Singhal2025-kf}. Such datasets  rarely capture the medical ambiguity encountered in practice.
Recent works have begun to investigate the role of human label uncertainty in model training and evaluation \cite{fc4b0d04b6c84ca18949b746f604eb0b, plank-2022-problem,  ICLR2025_87983214, fc4b0d04b6c84ca18949b746f604eb0b},
 identifying that the approximation of the human majority by LLM-as-a-judge may be an illusion, an artifact of high uncertainty in ground truth human labels \cite{ICLR2025_87983214}.  Stutz et al.\ \cite{STUTZ2025103556}  show how ignoring uncertainty leads to overly optimistic estimates of model performance in dermatology. While such studies provide empirical evidence of the existence of problems as a consequence of ground truth uncertainty, they \underline{do not explain why these problems occur under uncertainty} and if they are universal or simply artifacts of a specific dataset or model.

Our  contribution  provides a probabilistic model (\textit{\underline{as simple as a coin toss}}) for  \textit{expected accuracy} and \textit{expected F1},  to  estimate the impact of uncertainty on evaluation provides a theoretical explanation of why uncertainty obscures  performance problems of the underlying system regardless of the dataset, humans or models. We measure certainty through considering agreement on ground truth labels between multiple experts --- with high agreement implying high \textit{certainty}, and lower agreement corresponding to \textit{\textbf{un}certainty} in the data.
Our  simple closed-form probabilistic paradigm supports  empirical results as well as simulations to demonstrate :
d
{\setlength{\leftmargini}{0.5cm}
\begin{enumerate}
    \item  Wherever ground truth tends to be subjective, including domains like medicine,  \textit{high \textbf{certainty} in  ground truth is almost always necessary, but not sufficient, for human experts and models to achieve high performance} as measured by   ground truth uncertainty agnostic metrics such as F1 or accuracy. Here,  the ground truth correct answer is implicitly or explicitly based on a majority label and is not objective.

    \item  Under \textit{high \textbf{\textul{un}certainty} in ground truth answers, the performance of an independent expert drops. Thus  high uncertainty datasets cannot effectively discriminate between an expert and novice}. Hence, when the uncertainty is high or simply not quantified, it can lead to the incorrect conclusion that a weak and a strong performer have similar capabilities.  As an example, if two expert doctors disagree on the presence of a certain disease, even an answer selected by a random coin flip cannot be wrong, as its answer will match  one of the two experts and its performance will be similar to one of the doctors.
\end{enumerate}
}

\textbf{Empirical validation:} We  demonstrate the paradigm empirically using  the  CheXpert \cite{Irvin2019-im} dataset which consists of chest radiograph images of patients from Stanford Health Care between  2002  and 2017, where each X-ray in the evaluation set (500 X-rays) is annotated by 5 radiologists providing ground truth, and the performance of  3 additional human radiologists is also captured, allowing us to systematically measure and capture human ambiguity and compare human experts as well as model performance with ground truth. In addition, we use existing published results with over 200000 samples, from AI mammogram screening results \cite{Eisemann2025-gq}, as well as 3 NLI datasets\cite{williams-etal-2018-broad, snli:emnlp2015} and G-Eval \cite{liu-etal-2023-g}.

%KMV-20260425 ( thank you looks good)
Our work has significant implications for interpretation of model performance evaluation,  illustrating the need for explicit consideration of certainty of ground truth labels in evaluation.

\section{Theory --  A coin toss probabilistic paradigm for ground truth uncertainty}\label{sec:statisticalexplanation}
In order to explain why it is highly unlikely even for an expert  to achieve over 90\% performance on a dataset that has high disagreement rates among experts themselves, we present the following statistical analysis.

Let us assume that in general doctors have a probability $p_d$ of following the consensus of other doctors on the majority label. For instance, if for a given diagnosis,  3/5 doctors indicate  agreement on a finding while 2/5 disagree on that finding, here $p_d=\frac{3}{5}=0.6$ on the majority finding.  %\dr{The following sentence isn't clear to me}.
When  $p_d$ is computed empirically from the dataset by counting how many votes the majority answer gets compared to the  total number of answers for that item, it is the observed  $p_d$.   The observed  $p_d$ tends to be a closer approximation to true underlying $p_d$ as more opinions or answers are collected per item. When only one answer or opinion is collected per item, it does \textit{not} imply that $p_d$ is 1.0 (1/1), but rather that  the underlying  $p_d$  simply cannot be computed empirically.  In the case of a consensus answer determined by an independent adjudicator when 2 doctors disagree, the $p_d$ on the adjudicator's label is not 1.0, since the  majority label may  change when the adjudicator changes. That is, we should treat the adjudicator's label as  simply another expert's opinion. In the remainder of the paper, we assume that the observed $p_d$ is a close enough approximation of the true underlying $p_d$.

%Computing $p_d$ empirically from the dataset is the observed $p_d$, hence in the case of single label where multiple consensus is not collected, the agreement or uncertainty is simply not quantified and cannot be computed as a result of a consensus of 1.
%\dr{This analysis is important to the paper but it is too verbose and not explained cleanly.} Given datasets use the majority label  as the ground truth against which models or other doctor's performance is compared to, the probability of a doctor selecting the majority label is also  $p_d$. Assuming that the label that an expert selects is independent of what the others select (this is the case where they are blind to what the others have selected),  we can model the problem of selecting the right label as a biased coin toss problem (where the probability of a head is $p_d$), so the probability of obtaining heads is equivalent to selecting the majority label. Then,  directly following the Binominal distribution, in a sample of $N$  examples, the probability  of getting \textit{at least} $r_c$ items correct $P(Score >= r_c)$ is

We make the following 3 assumptions:

{\setlength{\leftmargini}{0.5cm}
\begin{enumerate}

    \item The ground truth answer against which any automatic system or another expert is compared to \textit{is a majority label} from people trained to perform the task. This is typically the case for subjective problems, where the ground truth  is either explicitly or implicitly an assumed majority label.
    \item The probability of an expert following the consensus of similarly trained peer group to select the majority label is $p_d$, where $p_d$ is also the probability of agreeing with the majority label.  This assumption is a key  contribution of this paper on how we model the probability of an expert selecting the  majority answer.  This assumption is also validated  empirically against real-world dataset in  subsequent sections (Section~\ref{sec:results}).
    \item The label assigned to one item  does not influence the label assigned to a different item, therefore label assignments are  independent of other labels  in the dataset.

\end{enumerate}
}

Using the above assumptions, we can model the problem of an expert selecting the majority (correct) label as a biased coin toss problem, where the probability of a head is $p_d$, so the probability of obtaining heads is equivalent to selecting the majority label. Then,  directly following the Binomial distribution, in a sample of $N$  examples, the probability  of getting \textit{at least} $r_c$ items correct $P(Score >= r_c)$ is
{\small
\begin{align}
   \intertext{Follows from Assumption 2}
  \text{Expected Accuracy score} &=  E(Accuracy) = p_d
  \label{equ:expectedacc} \\
   \text{Standard Error} &= SE = \sqrt{\frac{p_d*(1-p_d)}{N}}  \label{equ:expectedaccstandarderror}
   \intertext{Binomial distribution follows from above equation \ref{equ:expectedacc} and Assumption 3}
   P(Accuracy >= r_c) &= 1- \Sigma_{k=0}^{r_c-1} \binom{N}{k}p_d^k(1-p_d)^{(N-k)} \label{equ:biomial}
\end{align}
}
%
% {\small
% \end{equation}
% }

Applying the above equation, the probability that a doctor obtains at least 50 of 100 correct (in other words, the majority label matches), i.e., \ ($r_c >=50$), given that the probability of following the majority label  $p_d = 0.6$  is 0.98.  At the same time, when $p_d = 0.6$  the probability of obtaining more than 80\%  correct \ ( $r_c > = 80$ out of 100 samples) drops to near zero $1.6e-5$ as shown in Table~\ref{tab:probablitypfgettinghighscore}.

On the other hand, assume that  $p_d$ is 0.9, where 9 out of 10 doctors arrive at the same finding, then the probability of any other doctor who follows the same distribution obtaining a score of  80\% score or higher increases from near zero (when $p_d = 0.6$) to 0.99 as shown in Table~\ref{tab:probablitypfgettinghighscore}. Therefore, an individual rater's observed score can fall below or exceed $p_d$ through sampling variation and the magnitude depends  depends on the sample size, where at high sample sizes the magnitude becomes close to zero as per standard error in equation~\ref{equ:expectedaccstandarderror}.

\begin{table}[h!]
         \centering

{\tiny
    \begin{tabular}{r|rr|rr}
    \hline
     & \multicolumn{2}{c|}{$N=100$} & \multicolumn{2}{c}{$N=10$}  \\

        \textbf{$p_d$} & \textbf{$r_c$} & $P(Score >= r_c)$ &  \textbf{$r_c$} & $P(Accurcy  >= r_c)$  \\
        \hline
           \textbf{$p_d=p_r=0.25$} & \multicolumn{4}{c}{Random labeller baseline for 4 classes}\\
             \hline
           0.25 & 50 & 6.6e-8 & 5 & 0.08\\
         0.25 & 80 & 1.2e-30 & 8 & 0.00\\
         \hline
           \textbf{$p_d=p_r=0.5$} & \multicolumn{4}{c}{Random labeller baseline for 2 classes }\\
             \hline
           0.5 & 50 & 0.54 & 5 & 0.62\\
         0.5 & 80 & 5.6e-10 & 8 & 0.05\\
         \hline
         0.6 & 50 & 0.98 & 5 & 0.83 \\
         0.6 & 80 & 1.6e-5 & 8 & 0.17 \\   \hline
         0.9 & 50 & 1.00 & 5 & 0.99 \\
         0.9 & 80 & 0.99 & 8 & 0.92\\

        \hline

    \end{tabular}
    }
    \caption{Examples of how given  probability  $p_d$ of following the majority label, the corresponding probability of scores over 50\% or 80\% accuracy changes out of $N$ 100  vs.\ 10 examples.}
    \label{tab:probablitypfgettinghighscore}.

\end{table}
%\vspace{-1em}

Further comparing this behavior with a weak labeller, let us take a random labeller who randomly chooses to either agree or disagree with the majority label. In that case,  random labeller probability is $p_r = 0.5$ while a doctor who follows majority consensus in a highly uncertain dataset  $p_d = 0.5$ and  then  there is 54\% chance that the doctor and the random labeller will achieve the at least an average score of 50\%. On the other hand, on a dataset with $p_d = 0.9$, the average performance of a random labeller is only 50\% while that of a performer who follows a typical doctor's distribution has a 50\% probability of achieving more than 90\% accuracy. \textbf{Note:} As shown in Table~\ref{tab:probablitypfgettinghighscore}, when the sample size is small (say $N=10$), even a random labeller has a 5\% chance of  over 80\%, compared to a chance of near zero when the sample size is 100.
\textbf{Case of multi-class problems with  $L$ labels:} Here $p_r=\frac{1}{L}$, say for $L=4$, $p_r=0.25$ becomes the random labeller baseline.  Hence as $p_d \rightarrow 0.25$ the expert's performance will be no different to random labeller.

In summary, in a dataset that contains high uncertainty where agreement on what the correct answer is near borderline (say $p_d=0.5$ and that of  random labeller $p_r=0.5$), then there is   0.5 probability that the strong and weak labeller will achieve a score greater than or equal to 50\% and therefore, may not be able to effectively differentiate between a strong and a weak performer.  High agreement datasets are therefore almost always required for even an expert doctor to reach high scores  where the  correct answer is either implicitly or explicitly the   majority label. %We further demonstrate this phenomena empirically in real world datasets in Section~\ref{sec:results}, where the performance of human experts generally tends to drop at $p_d=0.6$ compared to  $p_d=1.0$.

%KMV-20260425 reworded
\textbf{Note:} We do not consider cases where the ground truth label is not based on a simple majority, such as where maximum likelihood estimation is utilized \cite{10.2307/2346806}. See Limitations in Appendix section~\ref{sec:limitations}.

\subsection{The impact of class imbalance}
Real-world datasets, especially in medicine, tend to be skewed, where there may be very few positive examples especially when the disease or condition has low prevalence.
Assume that given $N$ samples, the fraction  $m$, where $0<m<1$, represents the proportion of positive samples according to the majority label. For example, $N=100$, $m=0.2$ indicates that 20 ($0.2*100=20$) examples are positive according to the majority ground truth label.

To understand how $m$ impacts precision and recall given $p_d$, the probability of selecting the majority label, we introduce \textit{Expected} (\textbf{E}) metrics, reflecting an estimate of the number of True Positives (TP), False Positives (FP), False Negatives (FN) and True Negatives (TN) as a function of $m$ as follows:
%
%then out of $m*N$ positive samples, the expected (average) number of true positives $TP$ would be $m*N*p_d$, while the False negatives $FN$ would be $m*N*(1-p_d)$. Similarly, we calculate the expected number of False Positives (FP) and  True Negative (TN) as shown below to derive the expected precision and recall.
%
{
\small
\begin{align}
    E(TP) &= m*N*(p_d)  \\
    E(FN) &= m*N*(1-p_d)\\
    E(FP) &= (1-m)*N*(1-p_d) \\
    E(TN) &= (1-m)*N*(p_d) \\
    E(Precision) &= \frac{E(TP)}{E(TP)+E(FP)} \\
    &=\frac{m*p_d}{m*p_d + (1-m)(1-p_d)} \\
    E(Recall) &= \frac{E(TP)}{E(TP)+E(FN)} \\
    &=p_d\\
     E(F1) &= \frac{2*E(Precision)*E(Recall)}{E(Precision)+E(Recall)}\\
     &= \frac{2*m*p_d}{2*m*p_d + 1-p_d} \label{equ:equationf1}\\
     E(Accuracy) &= \frac{E(TP)+E(TN)}{E(TP)+E(FP)+E(TN)+E(FN)}
     \\ &= p_d
\end{align}
}
%\vspace{1em}
%

From the above equation, for the labeller who follows the distribution $p_d$ to select the majority label, \textbf{the recall is independent of  $m$}. Precision is  impacted by $m$ or the class balance ratio as shown above. If $m$ is very low at 0.01 (or high class imbalance which is often the case, e.g.\ rare diagnoses), at even high certainty datasets, say $p_d = 0.9$,  the expected precision  would be as low as 0.08 as shown computationally in Table~\ref{tab:expectedprecisionwithclassimbalance}. As a result,  experts at $p_d >= 0.9$, will have low F1 score when $m$ is low.

\begin{table}[h!]
    \centering
    {\tiny
    \begin{tabular}{cp{0.2\linewidth}cc}
    \hline
         \textbf{$p_d$}(Expected Recall) & \textbf{$m$}  & Expected Precision & Expected F1 \\
         \hline
          $p_d=p_r=0.5$ & \multicolumn{2}{c}{Random labeller baseline for 2 classes}\\
           \hline
           0.5 & 0.01 & 0.01 & 0.02 \\
            0.5 & 0.10 & 0.10 & 0.17 \\
         0.5 & 0.50 &  0.50 & 0.50 \\
        \hline
      \hline
      0.6 & 0.01  & 0.01 & 0.03  \\
         0.6 & 0.10  & 0.14 & 0.23  \\
         0.6 & 0.50  & 0.60 & 0.60 \\
         \hline
          0.9 & 0.01 & 0.08 & 0.15  \\
         0.9 & 0.10 & 0.50 & 0.64  \\
         0.9 & 0.50 & 0.90  & 0.90 \\
         \hline

        \hline
    \end{tabular}
    }
    \caption{Examples showing how given probability $p_d$ following the majority label, the positive class  ratio  $m$, the expected precision and F1 increases with increasing $m$.}
    %\caption{Examples of how given probability  $p_d$ following the majority label,  given the positive class  ratio of  $m$, how the expected precision and F1 increases with increasing $m$.}
    \label{tab:expectedprecisionwithclassimbalance}
\end{table}
An extension to the above equation is the case where in a dataset of $N$ samples, there are $k$ different agreement distributions, where $p_d \in {p_1.. p_k}$, with ${n_1,..n_k}$ samples where $N=\Sigma_k n_k$  and the corresponding positive class ratio $ {m_1..,m_k}$ (note: $m$ is wrt $n_k$).
{
\small
 \begin{align}
     E(Precision) &=\frac{\Sigma_1^k n_k*m_k*p_k}{\Sigma_1^k n_k ( m_k*p_k + (1-p_k)*(1-m_k)) }\\
     E(Recall)&= \frac{\Sigma_1^k n_k*m_k*p_k}{\Sigma_1^k n_k * m_k  }\\
     E(Accuracy) &= \frac{\Sigma_1^k n_k * p_k}{\Sigma_1^k n_k}
 \end{align}
}
\section{Simulation}
\begin{figure}[h!]
    \centering
    \includegraphics[width=1.0\linewidth]{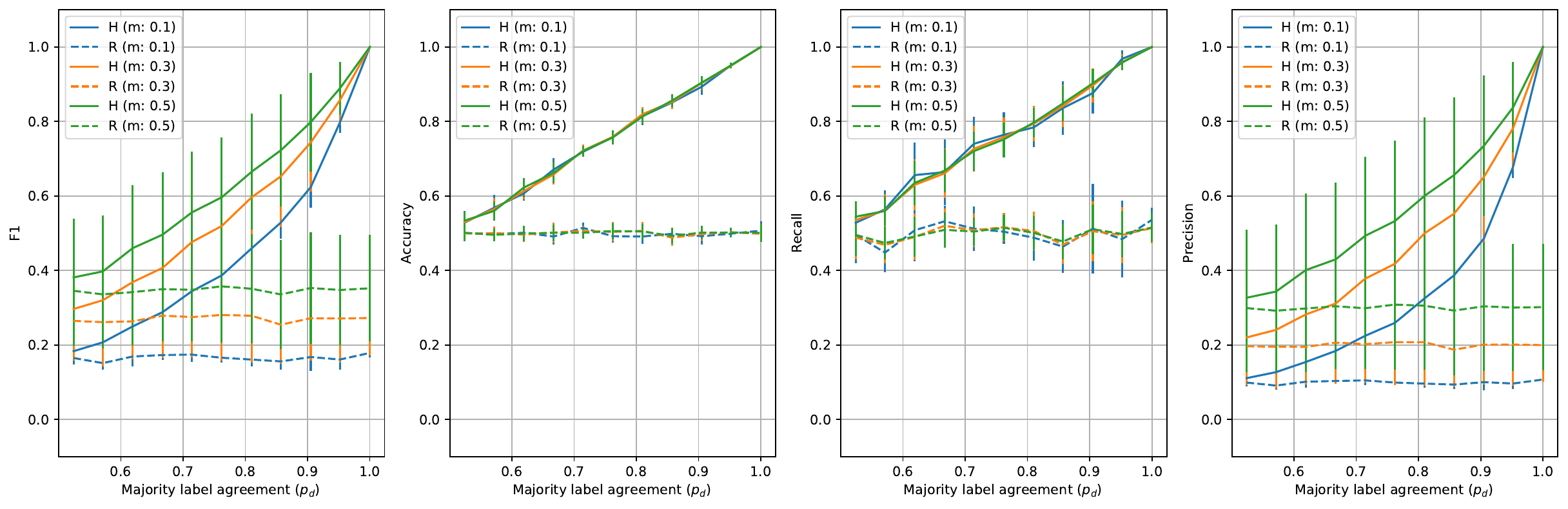}
    \caption{Simulated  impact of agreement for the majority label (X axis).  Y axis is the performance of a simulated human \textbf{(H)} who follows the majority label distribution ($p_d$) vs.\  the random labeler \textbf{(R)} performance across different positive label ratios $m \in {0.1, 0.3,  0.5}$.  At low certainty, the difference between human labeler  and random labeler is relatively low compared to high certainty.  Recall and Accuracy are unaffected by the positive label ratio. The vertical bars indicate standard deviation.}
    \label{fig:agreement_imulation}
\end{figure}

We simulate the behavior of a random labeler (who randomly picks one of the 2  binary labels with $\frac{1}{2}$ probability) vs.\ an expert (who tends to follow the rest of the experts' label distribution) using synthetically generated labels. Say, when a group of experts have  $\frac{3}{5}$ probability of selecting the same label, then another expert is likely to also have the same chance ($\frac{3}{5}$) of selecting the same label. We also vary positive sample ratios in the simulation in  Figure~\ref{fig:agreement_imulation}. The difference between the random labeller and the simulated human narrows during high uncertainty (Figure~\ref{fig:agreement_imulation}).
%KMV-20260425 what does "in addition" mean here? (  good point let me clarify)
In addition to low $p_d$, another factor  -- positive class ratio,  decreases the simulated human performance on F1  even further.

\section{Empirical Results}\label{sec:results}
We compare the performance of the state-of-the-art (SOTA) multimodal-LLMs, Gemini 2.5 pro, Gemini-3-Preview, GPT-5.1 and Gemini-3.1  to 3 independent human radiologists for the CheXpert dataset \cite{Irvin2019-im}. Please note that  the 3 radiologists and all models are disjoint from the 5 GT annotators, thus empirically validating assumption 2 and equations \ref{equ:expectedacc}, \ref{equ:equationf1}.
Each X-ray in the CheXpert dataset has 9 pathologies  (e.g.\ Atelectasis, Cardiomegaly, and Edema)  annotated by 5 ground truth radiologists independently.   Each  pathology is treated as a binary classification problem (positive finding, no finding).   The extent of uncertainty in ground truth is shown in Appendix Figure~\ref{fig:uncertaintyhexagon}, where samples with high agreement on positive findings drop substantially to less than 1\%. %The  pathologies curated in CheXpert dataset include Atelectasis, Cardiomegaly, and Edema. % and we compare the results for 9 of the 12  pathologies as the rest of the pathologies have very few positive samples (pathologies excluded  are inline with a previous SOTA   results \cite{Lee2025-yl, Tiu2022},  full results made available in Appendix~\ref{tab:appendix:fullperformancechexpert}).
\begin{figure*}[h!]
        \centering
        \includegraphics[width=0.48\linewidth]{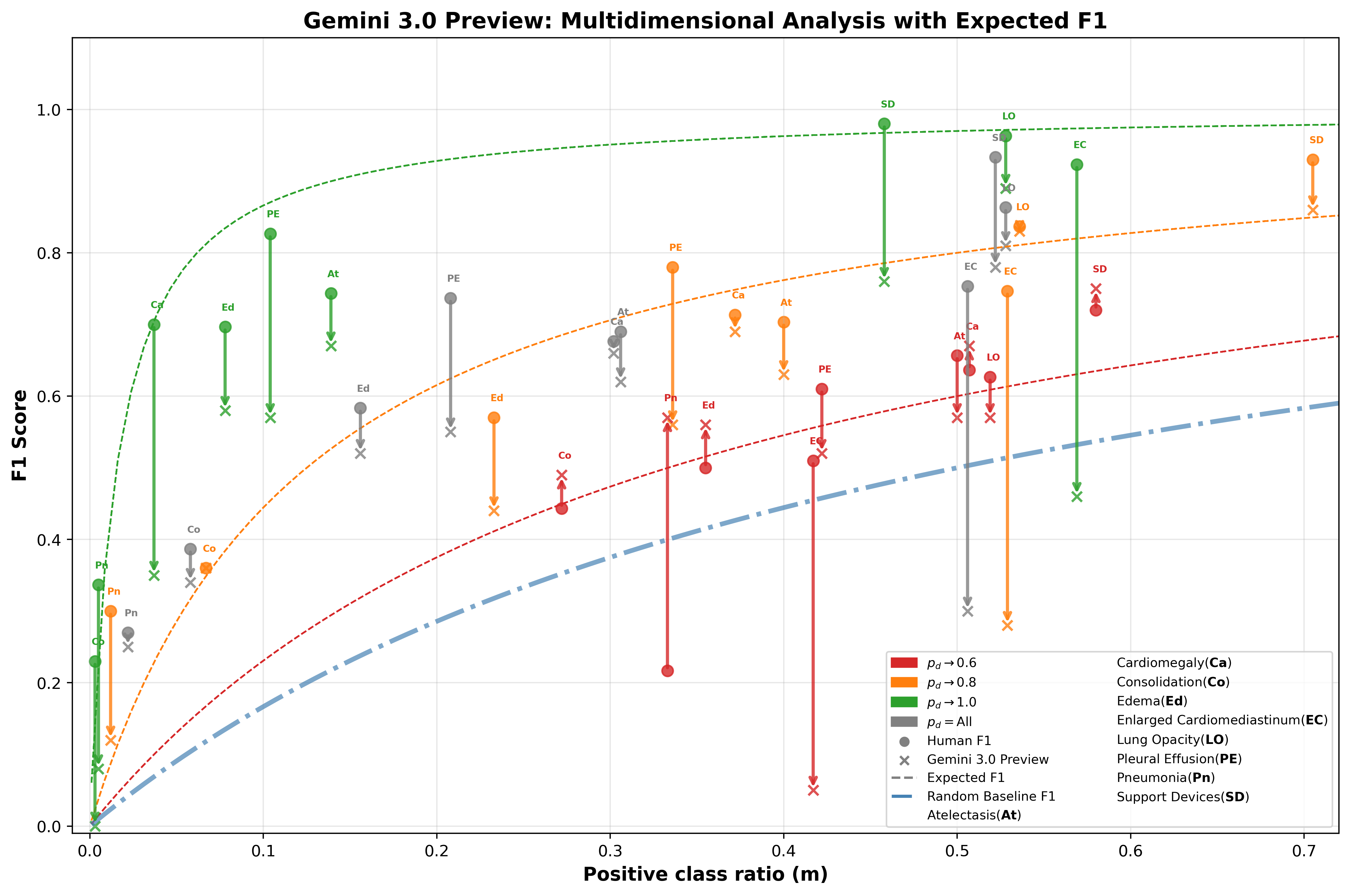}
         \includegraphics[width=0.48\linewidth]{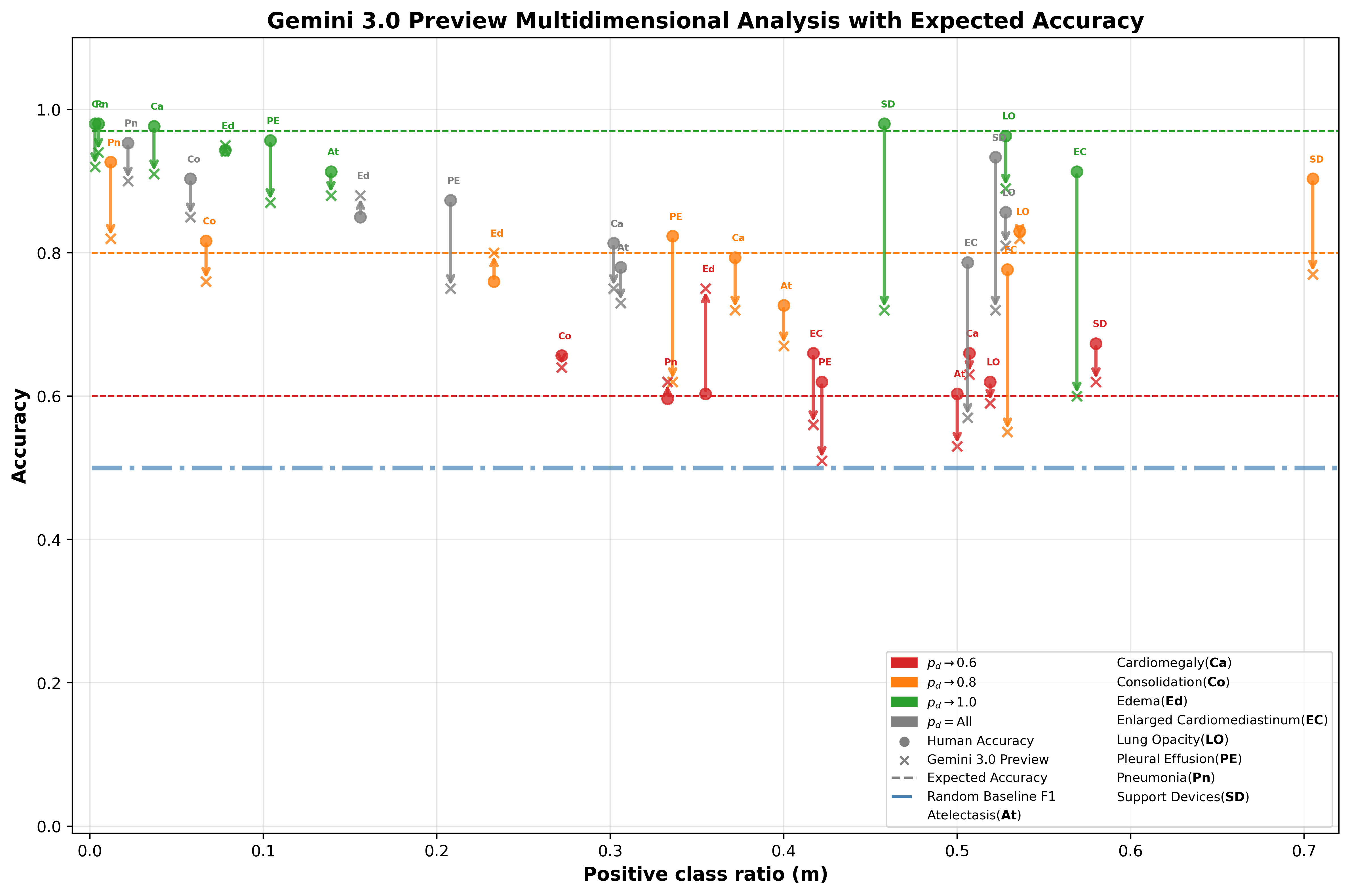}\\

          \includegraphics[width=0.48\linewidth]{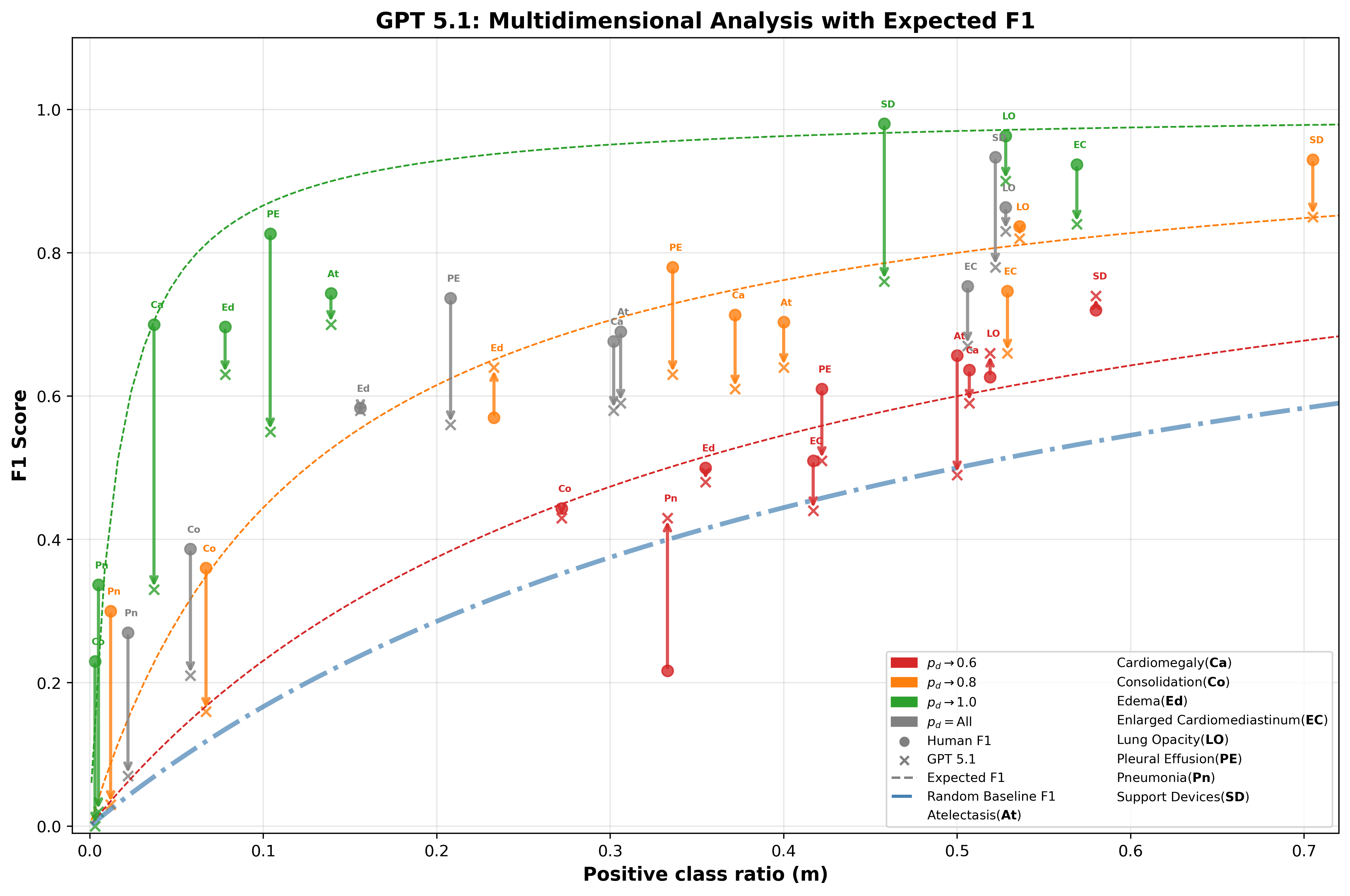}
         \includegraphics[width=0.48\linewidth]{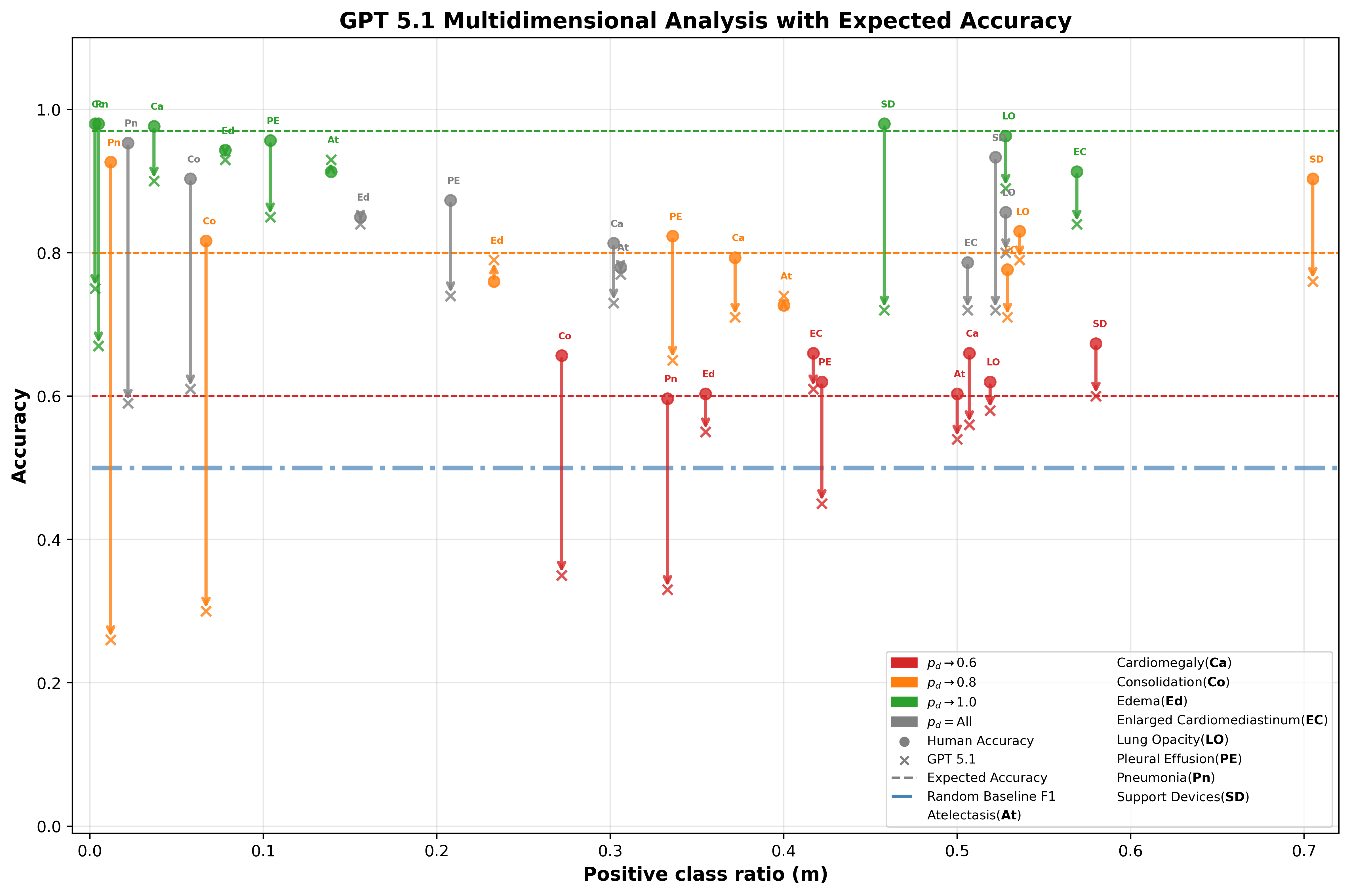}
        \caption{Comparison of theoretical vs.\ empirical performance of Models (\textbf{M}) and human (\textbf{H}) radiologists,  compared against ground truth (majority label across 5 human radiologists). \textbf{H} is the average performance of additional 3 radiologists.   Dashed lines are theoretical expected performance at given $p_d$ and $m$  according to equations~\ref{equ:expectedacc} and \ref{equ:equationf1}. Theoretical performance closely matches human expert performance ($p_d\rightarrow1.0$ is approximated as $p_d=0.97$).   Humans  achieve a relatively high F1-score ($> 0.8$) as  $p_d\rightarrow1.0$, whereas  $p_d\rightarrow0.6$ the peak  performance drops close to a random labeller baseline. At low positive class ratio ($m<0.01$), even at $p_d\rightarrow1.0$  (see Pneumonia and Consolidation), F1 is $< 0.35$. $\Delta = H - M$ is the vertical distance  illustrated by the length of the vertical line connecting the human and the model performance.     See   Appendix Tables~\ref{tab:appendix:gemini_profull} and  ~\ref{tab:gpt5.1} for detailed results with sample size. Similar results for Gemini-2.5 \& Gemini-3.1 are in Appendix Figure~\ref{fig:gemini2.5vshuman} and Tables ~\ref{tab:appendix:gemini25pro} ~\ref{tab:appendix:gemini31pro}.}
        \label{fig:modelvshumansunderuncertainity}
\end{figure*}

Given the extent of uncertainty in ground truth labels, in addition to a typical aggregate results that does \textit{not} account for uncertainty,  we also stratify the results by the probability ($p_d$) of agreement among experts  for the majority label, where $p_d$ is computed empirically from the dataset. As an example,  say out of the 5 ground truth set of experts, 4 have selected the same label to form the majority, then $p_d=\frac{4}{5}=0.8$.
As shown in Figure~\ref{fig:modelvshumansunderuncertainity}, there are 3  key observations where expert  absolute performance is  high at high $p_d$ and as $p_d$ drops the relative \textit{lack of difference} to a non-expert.
{\setlength{\leftmargini}{0.5cm}
\begin{enumerate}
  \item %The best absolute performance human radiologists achieve across all pathologies is 0.98 F1 and 0.98 accuracy. This high performance occurs in bins with high ground truth agreement as $p_d$ tends to 1.0 ($p_d\rightarrow1.0$).
As $p_d\rightarrow1.0$ (\textcolor{pdgreen}{\textbf{\textit{green lines}}}  in chart~\ref{fig:modelvshumansunderuncertainity}), the average human F1 across  pathologies with $m>=0.1$ is  $0.89\pm0.1$,  while the F1 including the 4 pathologies that have   $m<0.1$ is  $0.71\pm0.27$. As $p_d\rightarrow1.0$, the average accuracy (independent of $m$) is $0.95\pm.02$  and the expected accuracy $\rightarrow$ 1.0  according to Equation~\ref{equ:expectedacc}.  In pathologies where the models do perform well, the    peak  performance is also as $p_d\rightarrow1.0$, with Gemini-3 and GPT-5 scoring an F1 0.89 and 0.90 respectively for Lung Opacity as shown in Table~\ref{tab:exampleperfvar}. \textit{\textul{In short, this absolute high performance, regardless of model or human, occurs in bins with high ground truth agreement as $p_d\rightarrow1.0$, and  the expected theoretical performance  approximates the empirical performance of experts}}. Thus, high  certainty is a necessary condition for high expert performance.

    \item  As a consequence \textbf{(a)},  only when the agreement in ground truth  is high, does a   \textit{model's poor performance} become more apparent. As a result,  in the bins with high certainty is where the  largest performance  gap can be observed between human experts and models. For instance, for Support Devices, humans achieve F1 0.98 ($p_d\rightarrow1.0$), and the $\Delta=H-M$ compared to models is over 22 points for both Gemini-3 and GPT-5.1 as shown in Table~\ref{tab:exampleperfvar}.

    \item As a consequence \textbf{(b)}, the performance of humans drops substantially at low agreement ($p_d\rightarrow0.6$ \textcolor{pdred}{\textbf{\textit{red lines}}} in chart~\ref{fig:modelvshumansunderuncertainity}) with an average F1 of $0.55\pm.15$, accuracy $0.63\pm.02$ and expected accuracy of $0.6$ according to Equation~\ref{equ:expectedacc}. Here, the relative performance difference narrows the most between humans and models even in pathologies \textit{where models do not perform particularly well}, with models outperforming the radiologists in some cases. For instance, for Support Devices, human performance drops from F1 0.98 ($p_d\rightarrow1.0$)
    to F1 0.72 ($p_d\rightarrow0.6$), whereas the   performance of all 4 models remains largely the same  $0.76\pm0.03$, outperforming radiologists by $4\pm2$ points. \textit{Sample size note:} Akin to a random coin flip achieving 100\% hit rate on very few samples,  $p_d\rightarrow0.6$ with very few samples ($N<=12$), humans can achieve an accuracy of 0.8. At $N=12$, this has a $\approx22\%$ chance according to Equation~\ref{equ:biomial}.

\end{enumerate}
}

\begin{table}[h!]
    \setlength{\tabcolsep}{4pt}
    \centering
    \resizebox{1.0\columnwidth}{!}
    {
    \begin{tabular}{p{0.1\linewidth}rrr||rrrrrr||rrrrrr}
\toprule
  \multicolumn{4}{c||}{  $|GT|= 5, |L|$=2}  & \multicolumn{6}{c|}{F1} & \multicolumn{6}{c}{Accuracy}\\
 \hline
 Pathology & $p_d$ & S & $m$  & AI & $\text{H}$\ \   & $E$  & $\Delta_{\text{H}-AI}$ & $\Delta_{E-\text{H}}$ & $\Delta_{E-M}$ & AI & $\text{H}$\ \ \  &   $E$  & $\Delta_{\text{H}-AI}$ & $\Delta_{E-\text{H}}$ & $\Delta_{E-M}$\\

\thickhline
 \multicolumn{16}{c}{Gemini-3.0 Preview}\\

\thickhline

\multirow[c]{4}{*}{\shortstack[l]{Cardio-\\megaly}} & 0.60 & 142 & 0.507 & 0.67 & 0.64$\pm$0.10 & 0.60 & -0.03 & -0.04 & -0.07 & 0.63 & 0.66$\pm$0.03 & 0.60 & 0.03 & -0.06 & -0.03 \\
\cline{2-16} \cline{3-16}
 & 0.80 & 196 & 0.372 & 0.69 & 0.71$\pm$0.04 & 0.75 & 0.02 & 0.04 & 0.06 & 0.72 & 0.79$\pm$0.03 & 0.80 & 0.07 & 0.01 & 0.08 \\
\cline{2-16} \cline{3-16}
 & 1.00 & 162 & 0.037 & 0.35 & 0.70$\pm$0.12 & 0.71 & 0.35 & 0.01 & 0.36 & \mycolorbox{paleblue}{0.91} & \mycolorbox{paleblue}{0.98$\pm$0.02} & 0.97 & 0.07 & -0.01 & 0.06 \\
\cline{2-16} \cline{3-16}
 & All & 500 & 0.302 & 0.66 & 0.68$\pm$0.07 & 0.68 & 0.02 & 0.00 & 0.02 & 0.75 & \mycolorbox{paleblue}{0.81$\pm$0.03} & 0.80 & 0.06 & -0.01 & 0.05 \\
\thickhline
\multirow[c]{4}{*}{\shortstack[l]{Lung\\Opacity}} & 0.60 & 106 & 0.519 & 0.57 & 0.63$\pm$0.07 & 0.61 & 0.06 & -0.02 & 0.04 & 0.59 & 0.62$\pm$0.03 & 0.60 & 0.03 & -0.02 & 0.01 \\
\cline{2-16} \cline{3-16}
 & 0.80 & 125 & 0.536 & \mycolorbox{paleblue}{0.83} & \mycolorbox{paleblue}{0.84$\pm$0.03} & 0.81 & 0.01 & -0.03 & -0.02 & \mycolorbox{paleblue}{0.82} & \mycolorbox{paleblue}{0.83$\pm$0.03} & 0.80 & 0.01 & -0.03 & -0.02 \\
\cline{2-16} \cline{3-16}
 & 1.00 & 269 & 0.528 & \mycolorbox{paleblue}{0.89} & \mycolorbox{paleblue}{0.96$\pm$0.01} & 0.97 & 0.07 & 0.01 & 0.08 & \mycolorbox{paleblue}{0.89} & \mycolorbox{paleblue}{0.96$\pm$0.01} & 0.97 & 0.07 & 0.01 & 0.08 \\
\cline{2-16} \cline{3-16}
 & All & 500 & 0.528 & \mycolorbox{paleblue}{0.81} & \mycolorbox{paleblue}{0.86$\pm$0.02} & 0.86 & 0.05 & -0.00 & 0.05 & \mycolorbox{paleblue}{0.81} & \mycolorbox{paleblue}{0.86$\pm$0.01} & 0.85 & 0.05 & -0.01 & 0.04 \\
\thickhline
\multirow[c]{4}{*}{\shortstack[l]{Support\\Devices}} & 0.60 & 50 & 0.580 & 0.75 & 0.72$\pm$0.04 & 0.64 & -0.03 & -0.08 & -0.11 & 0.62 & 0.67$\pm$0.05 & 0.60 & 0.05 & -0.07 & -0.02 \\
\cline{2-16} \cline{3-16}
 & 0.80 & 105 & 0.705 & \mycolorbox{paleblue}{0.86} & \mycolorbox{paleblue}{0.93$\pm$0.02} & 0.85 & 0.07 & -0.08 & -0.01 & 0.77 & \mycolorbox{paleblue}{0.90$\pm$0.02} & 0.80 & 0.13 & -0.10 & 0.03 \\
\cline{2-16} \cline{3-16}
 & 1.00 & 345 & 0.458 & 0.76 & \mycolorbox{paleblue}{0.98$\pm$0.00} & 0.97 & 0.22 & -0.01 & 0.21 & 0.72 & \mycolorbox{paleblue}{0.98$\pm$0.00} & 0.97 & 0.26 & -0.01 & 0.25 \\
\cline{2-16} \cline{3-16}
 & All & 500 & 0.522 & 0.78 & \mycolorbox{paleblue}{0.93$\pm$0.01} & 0.90 & 0.15 & -0.03 & 0.12 & 0.72 & \mycolorbox{paleblue}{0.93$\pm$0.01} & 0.90 & 0.21 & -0.03 & 0.18 \\
\thickhline
\multicolumn{16}{c}{GPT-5.1}\\
\thickhline

\multirow[c]{4}{*}{\shortstack[l]{Cardio-\\megaly}} & 0.60 & 142 & 0.507 & 0.59 & 0.64$\pm$0.10 & 0.60 & 0.05 & -0.04 & 0.01 & 0.56 & 0.66$\pm$0.03 & 0.60 & 0.10 & -0.06 & 0.04 \\
\cline{2-16} \cline{3-16}
 & 0.80 & 196 & 0.372 & 0.61 & 0.71$\pm$0.04 & 0.75 & 0.10 & 0.04 & 0.14 & 0.71 & 0.79$\pm$0.03 & 0.80 & 0.08 & 0.01 & 0.09 \\
\cline{2-16} \cline{3-16}
 & 1.00 & 162 & 0.037 & 0.33 & 0.70$\pm$0.12 & 0.71 & 0.37 & 0.01 & 0.38 & \mycolorbox{paleblue}{0.90} & \mycolorbox{paleblue}{0.98$\pm$0.02} & 0.97 & 0.08 & -0.01 & 0.07 \\
\cline{2-16} \cline{3-16}
 & All & 500 & 0.302 & 0.58 & 0.68$\pm$0.07 & 0.68 & 0.10 & 0.00 & 0.10 & 0.73 & \mycolorbox{paleblue}{0.81$\pm$0.03} & 0.80 & 0.08 & -0.01 & 0.07 \\
\thickhline
\multirow[c]{4}{*}{\shortstack[l]{Lung\\Opacity}} & 0.60 & 106 & 0.519 & 0.66 & 0.63$\pm$0.07 & 0.61 & -0.03 & -0.02 & -0.05 & 0.58 & 0.62$\pm$0.03 & 0.60 & 0.04 & -0.02 & 0.02 \\
\cline{2-16} \cline{3-16}
 & 0.80 & 125 & 0.536 & \mycolorbox{paleblue}{0.82} & \mycolorbox{paleblue}{0.84$\pm$0.03} & 0.81 & 0.02 & -0.03 & -0.01 & 0.79 & \mycolorbox{paleblue}{0.83$\pm$0.03} & 0.80 & 0.04 & -0.03 & 0.01 \\
\cline{2-16} \cline{3-16}
 & 1.00 & 269 & 0.528 & \mycolorbox{paleblue}{0.90} & \mycolorbox{paleblue}{0.96$\pm$0.01} & 0.97 & 0.06 & 0.01 & 0.07 & \mycolorbox{paleblue}{0.89} & \mycolorbox{paleblue}{0.96$\pm$0.01} & 0.97 & 0.07 & 0.01 & 0.08 \\
\cline{2-16} \cline{3-16}
 & All & 500 & 0.528 & \mycolorbox{paleblue}{0.83} & \mycolorbox{paleblue}{0.86$\pm$0.02} & 0.86 & 0.03 & -0.00 & 0.03 & \mycolorbox{paleblue}{0.80} & \mycolorbox{paleblue}{0.86$\pm$0.01} & 0.85 & 0.06 & -0.01 & 0.05 \\
\thickhline
\multirow[c]{4}{*}{\shortstack[l]{Support\\Devices}} & 0.60 & 50 & 0.580 & 0.74 & 0.72$\pm$0.04 & 0.64 & -0.02 & -0.08 & -0.10 & 0.60 & 0.67$\pm$0.05 & 0.60 & 0.07 & -0.07 & 0.00 \\
\cline{2-16} \cline{3-16}
 & 0.80 & 105 & 0.705 & \mycolorbox{paleblue}{0.85} & \mycolorbox{paleblue}{0.93$\pm$0.02} & 0.85 & 0.08 & -0.08 & 0.00 & 0.76 & \mycolorbox{paleblue}{0.90$\pm$0.02} & 0.80 & 0.14 & -0.10 & 0.04 \\
\cline{2-16} \cline{3-16}
 & 1.00 & 345 & 0.458 & 0.76 & \mycolorbox{paleblue}{0.98$\pm$0.00} & 0.97 & 0.22 & -0.01 & 0.21 & 0.72 & \mycolorbox{paleblue}{0.98$\pm$0.00} & 0.97 & 0.26 & -0.01 & 0.25 \\
\cline{2-16} \cline{3-16}
 & All & 500 & 0.522 & 0.78 & \mycolorbox{paleblue}{0.93$\pm$0.01} & 0.90 & 0.15 & -0.03 & 0.12 & 0.72 & \mycolorbox{paleblue}{0.93$\pm$0.01} & 0.90 & 0.21 & -0.03 & 0.18 \\
\thickhline
 \end{tabular}
 }
 \caption{Examples of models (\textbf{M}) Gemini-3.0 and GPT-5.1 performance variation when  stratified by $p_d$ compared to average human (\textbf{H}) performance of 3 radiologists  and the corresponding standard deviation, sample size \textbf{S}.  For Lung Opacity, the overall ($p_d=All$) performance is over 80\%,  and even at stratified $p_d$, the models  achieves reasonable F1 and accuracy scores compared to humans. On the other hand, for Cardiomegaly and Support Devices the overall ($p_d=All$) performance falls below 80\% and and Gemini even achieves a  $\Delta=H-M$ of just 2 points for Cardiomegaly, however as $p_d \rightarrow  1$, the $\Delta$ is over 20 points.  Expected (\textbf{E}) performance given $\langle p_d, m \rangle$  per equations~\ref{equ:expectedacc} and \ref{equ:equationf1} (note: $p_d\rightarrow1.0$ is approximated as $p_d=0.97$).  Scores (Model or Human) over \mycolorbox{paleblue}{0.80} are highlighted and they all occur at $p_d \in {0.8, 1.0}$ but never at $0.60$. Full  results   in Appendix Tables~\ref{tab:appendix:gemini_profull},  and ~\ref{tab:gpt5.1}.}\label{tab:exampleperfvar}
 \end{table}

\begin{figure}[h!]
    \centering
    \includegraphics[width=0.32\linewidth]{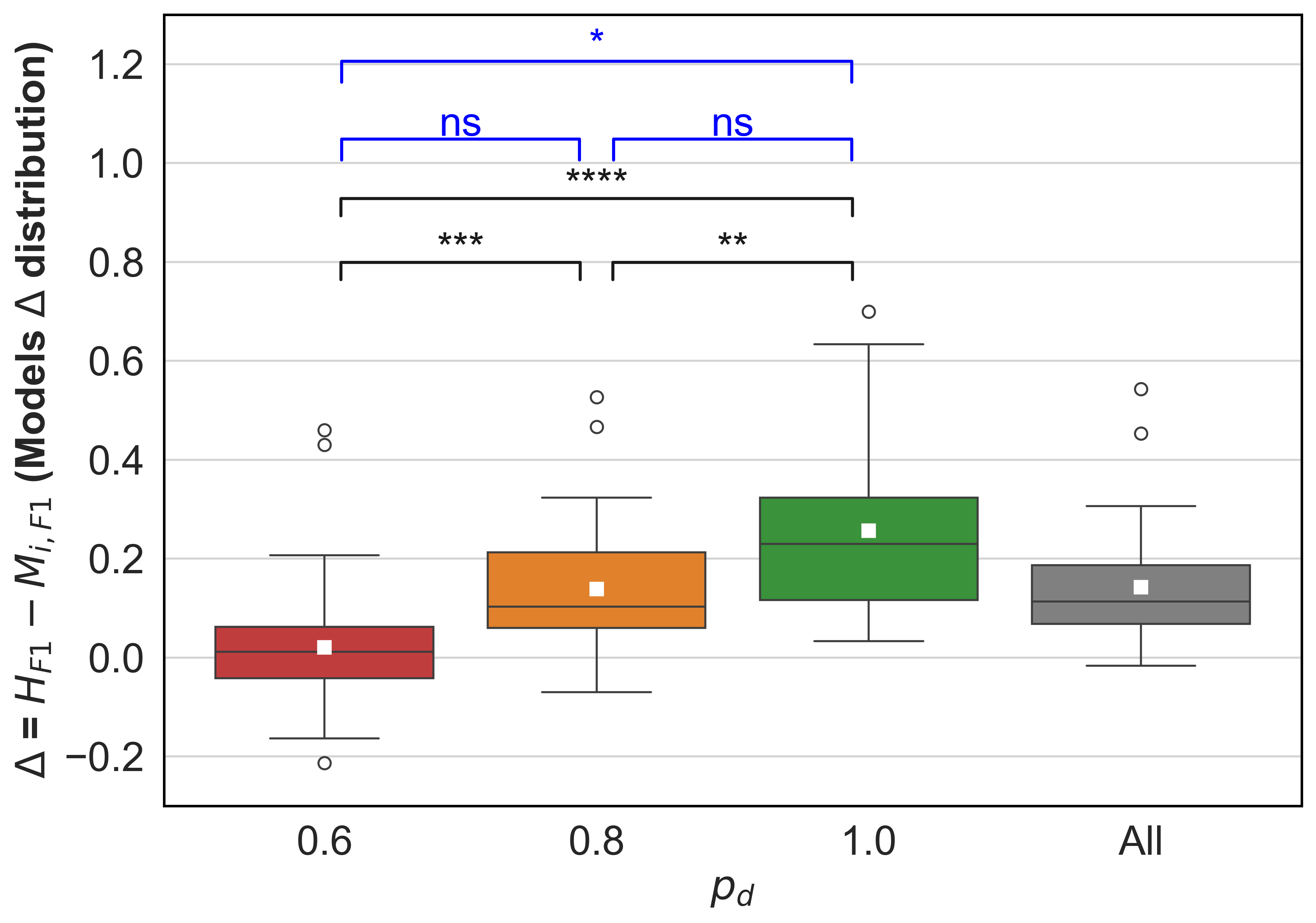}
\includegraphics[width=0.32\linewidth]
{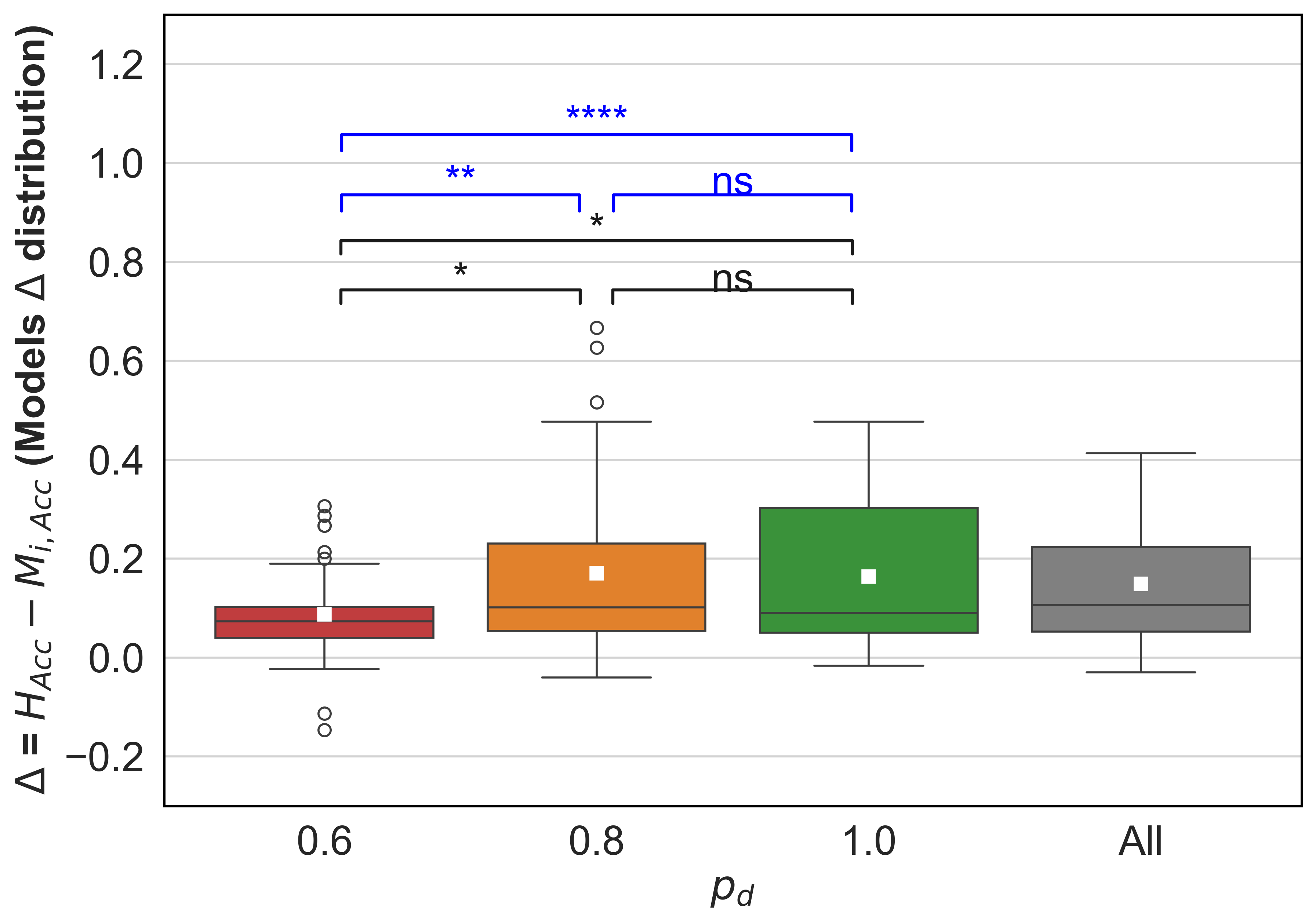}
\includegraphics[width=0.32\linewidth]{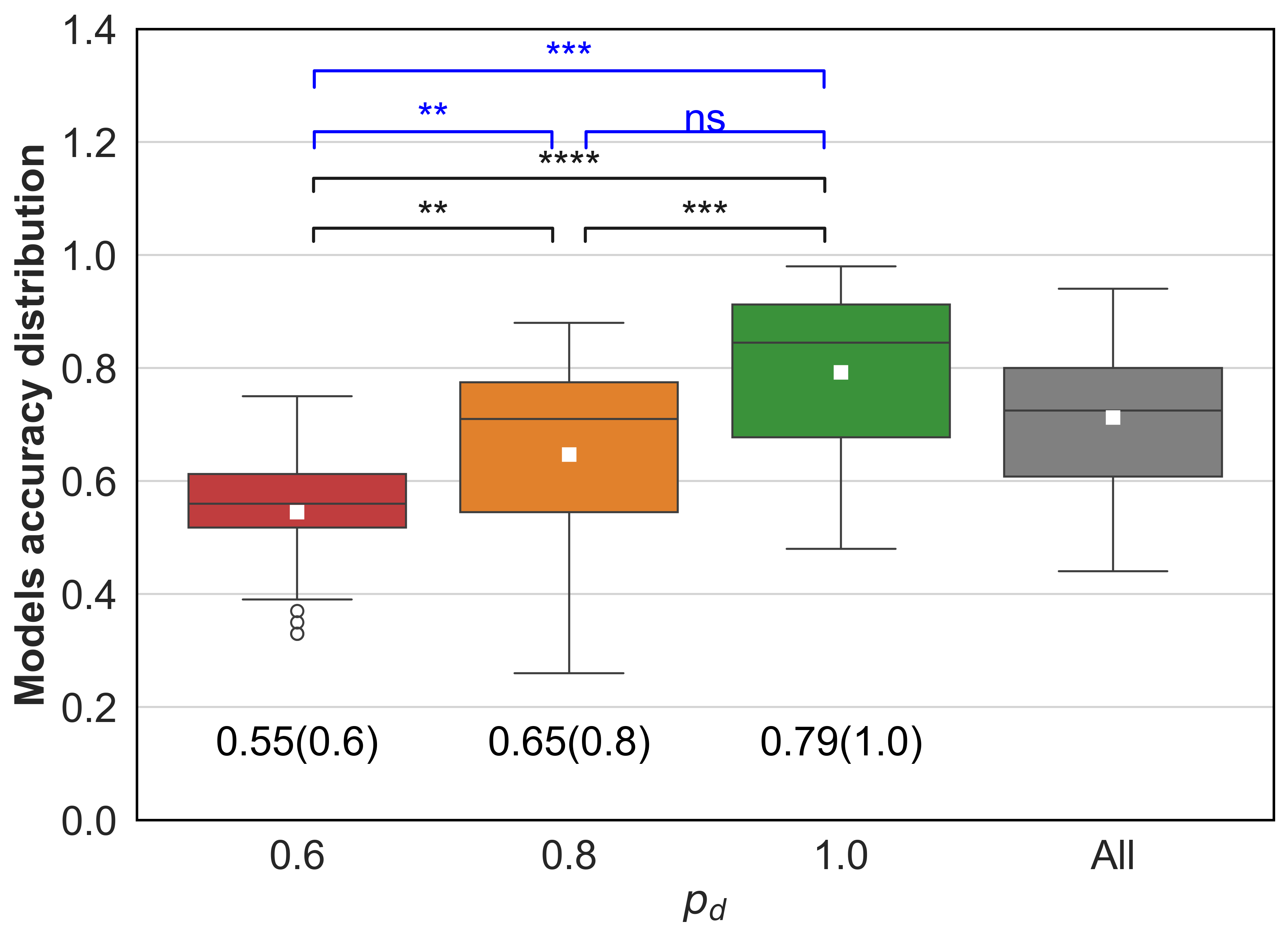}
    \caption{\textbf{(Left and Mid) Distribution of difference in Human vs.\  Model $\Delta=H-M$:} Each point $\Delta_{i, o, p_d}=H_{o, p_d}-M_{i, o, p_d}$ where $i\in\{\text{Gemini-3.1}, \text{Gemini-3.0}, \text{Gemini-2.5}, \text{GPT-5.1} \}$, $p_d\in\{0.6,0.8,1.0, \text{All}\}$,  $o  \in \text{Pathologies}$. $H_{o, p_d}$ is the human score (F1 or accuracy) against the pathology $o$ at $p_d$. Similarly, $M_{i, o, p_d}$ is the performance of the model $i$ against the pathology $o$ at $p_d$. Average $\Delta$ is generally higher as $p_d\rightarrow1.0$ compared to $p_d\rightarrow0.6$ where $\Delta$ F1 is close to zero.  \textbf{(Right) Distribution of absolute model accuracy:} All 4 model accuracies (agnostic of $m$) across different $p_d$. Empirical vs.\ (expected) mean annotated in text. \textbf{Note:} Significance Welch t-test for means ($H_0$ Null hypothesis: there is no difference between the means) and Levene's test (trimmed, proportion to cut=0.1) for variance ($H_0$: there is no difference in variance):  **** ($p<=0.0001$), *** ($p<=0.001$), ** ($p<=0.01$), * ($p<=0.05$), ns (not significant). \textcolor{blue}{Blue}: Indicates significance of variance.}
    \label{fig:delte}
\end{figure}

The overall distribution of   $\Delta=H-M$, where $\Delta$ tends to be relatively  higher $p_d\rightarrow1.0$ compared to  $p_d\rightarrow0.6$  where  average $\Delta=H-M$ for F1 drops to near zero, is  illustrated  in Figure~\ref{fig:delte}.
Without stratification by uncertainty, based on aggregate performance, we would draw a  misleading conclusion that  there is no statistically significant difference between a weak algorithm (including models) and expert doctors. For instance, previous studies have reported \cite{Tiu2022, Lee2025-yl} that the models achieve expert level performance in CheXpert dataset,  where the model scores F1 0.60 while the radiologists score F1 0.62 across 5 pathologies.  Drawing parallels to our study, for instance for Cardiomegaly, the difference between human and Gemini-3  $\Delta = 0.68 - 0.66  = 0.02$, but as $p_d\rightarrow1.0$ the performance of the Gemini-3 model compared to humans is much lower with $\Delta=0.35$ as shown in Table~\ref{tab:exampleperfvar}.

\begin{table}[h!]
     \setlength{\tabcolsep}{4pt}
    \centering
    \resizebox{1.0\columnwidth}{!}
    {
  \begin{tabular}{rrr||rrrrrr||rrrrrr}
\thickhline
  \multicolumn{3}{c||}{$ |GT| \in \langle 2, 3 \rangle$, $|L|$=2} & \multicolumn{6}{c||}{F1}        & \multicolumn{6}{c}{Accuracy (A)}\\
\thickhline
 $p_d$ & S & $m$  & AI & $\text{H}'$\ \   & $E$  & $\Delta_{\text{H}'-AI}$ & $\Delta_{E-\text{H}'}$ & $\Delta_{E-M}$ & AI & $\text{H}'$\ \ \  &   $E$  & $\Delta_{\text{H}'-AI}$ & $\Delta_{E-\text{H}'}$ & $\Delta_{E-M}$\\

\thickhline
% 0.67 & 15834 & 0.229 & 0.39 & 0.48$\pm$0.0045 & 0.09 & 0.26 & 0.37$\pm$0.0055 & 0.11 & 0.79 & 0.66$\pm$0.0055 & -0.13 & 0.42 & 0.67$\pm$0.0000 & 0.25 \\
% 1.00 & 185245 & 0.029 & 0.11 & 1.00$\pm$0.0000 & 0.89 & 0.06 & 1.00$\pm$0.0000 & 0.94 & 0.89 & 1.00$\pm$0.0000 & 0.11 & 0.58 & 1.00$\pm$0.0000 & 0.42 \\
% ALL & 201079 & 0.045 & 0.15 & 0.75$\pm$0.0000 & 0.60 & 0.08 & 0.66$\pm$0.0045 & 0.58 & 0.85 & 0.87$\pm$0.0000 & 0.02 & 0.56 & 0.97$\pm$0.0000 & 0.41 \\

0.67 & 15834 & 0.229 & 0.39 & 0.48$\pm$0.0045 & 0.48 & 0.09 & 0.00 & 0.09 & 0.42 & 0.67$\pm$0.0045 & 0.67 & 0.25 & 0.00 & 0.25 \\
1.00 & 185245 & 0.029 & 0.11 & 1.00$\pm$0.0000 & 1.00 & 0.89 & -0.00 & 0.89 & 0.58 & 1.00$\pm$0.0000 & 1.00 & 0.42 & 0.00 & 0.42 \\
ALL & 201079 & 0.045 & 0.15 & 0.75$\pm$0.0000 & 0.75 & 0.60 & -0.00 & 0.60 & 0.56 & 0.97$\pm$0.0000 & 0.97 & 0.41 & 0.00 & 0.41 \\
\bottomrule
\end{tabular}
}
    \caption{Stratified performance on existing public results  \cite{Eisemann2025-gq} on Mammogram. \textbf{S} is the sample size, $\text{H}'$ is simulated human performance based on GT distribution and hence matches expected performance. $\Delta_{E-M}$  narrows as certainty lowers as shown when $p_d \rightarrow 0.67$ compared to $p_d \rightarrow 1.0$.}
    %At $p_d=0.67, m=0.229$:  E(F1) = 0.48 ($\text{H}'=0.48$), E(P) = 0.37 ($\text{H}'=0.37$) and E(R)  = 0.67 ($\text{H}'=0.66$), Expected(A) = 0.67 ($\text{H}'=0.67$).}
    \label{tab:momogram}
\end{table}

In the population-wide  mammogram screening dataset \cite{Eisemann2025-gq}, which has public results for their proprietary AI model, we reuse existing results and stratify them by $p_d$. In this dataset, 2 doctors independently provide their finding and when there is disagreement, a third consensus label   is used as the ground truth. We select the set of around 200000 samples where doctors provided their finding (normal, suspicious) without AI assistance, and compare the doctors' majority finding with the model's prediction.  In this dataset another independent doctor's finding label is not available to compare with the ground truth, so we randomly select one of the labels 5 times  from the ground truth labels and average the score to mimic the average performance of any other human  $\text{H}'$.  As shown in Table~\ref{tab:momogram}, the difference in expected performance and models $\Delta_{E-M}$  narrows as certainty lowers as shown when $p_d \rightarrow 0.67$ compared to $p_d \rightarrow 1.0$.
\subsection{Uncertainty and its impact on model comparisons}
\begin{figure}
    \centering

    \includegraphics[width=0.32\linewidth]{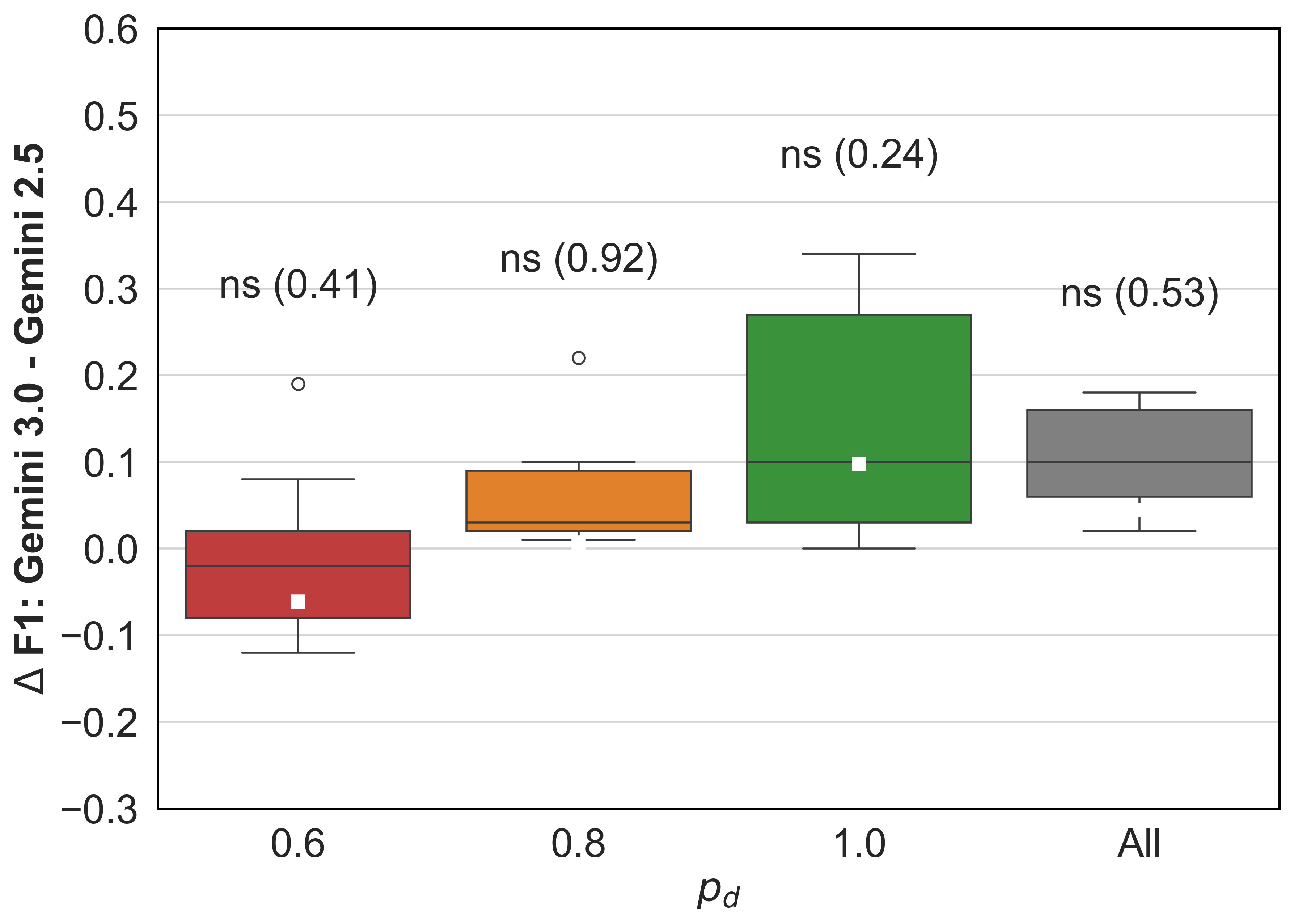}
\includegraphics[width=0.32\linewidth]{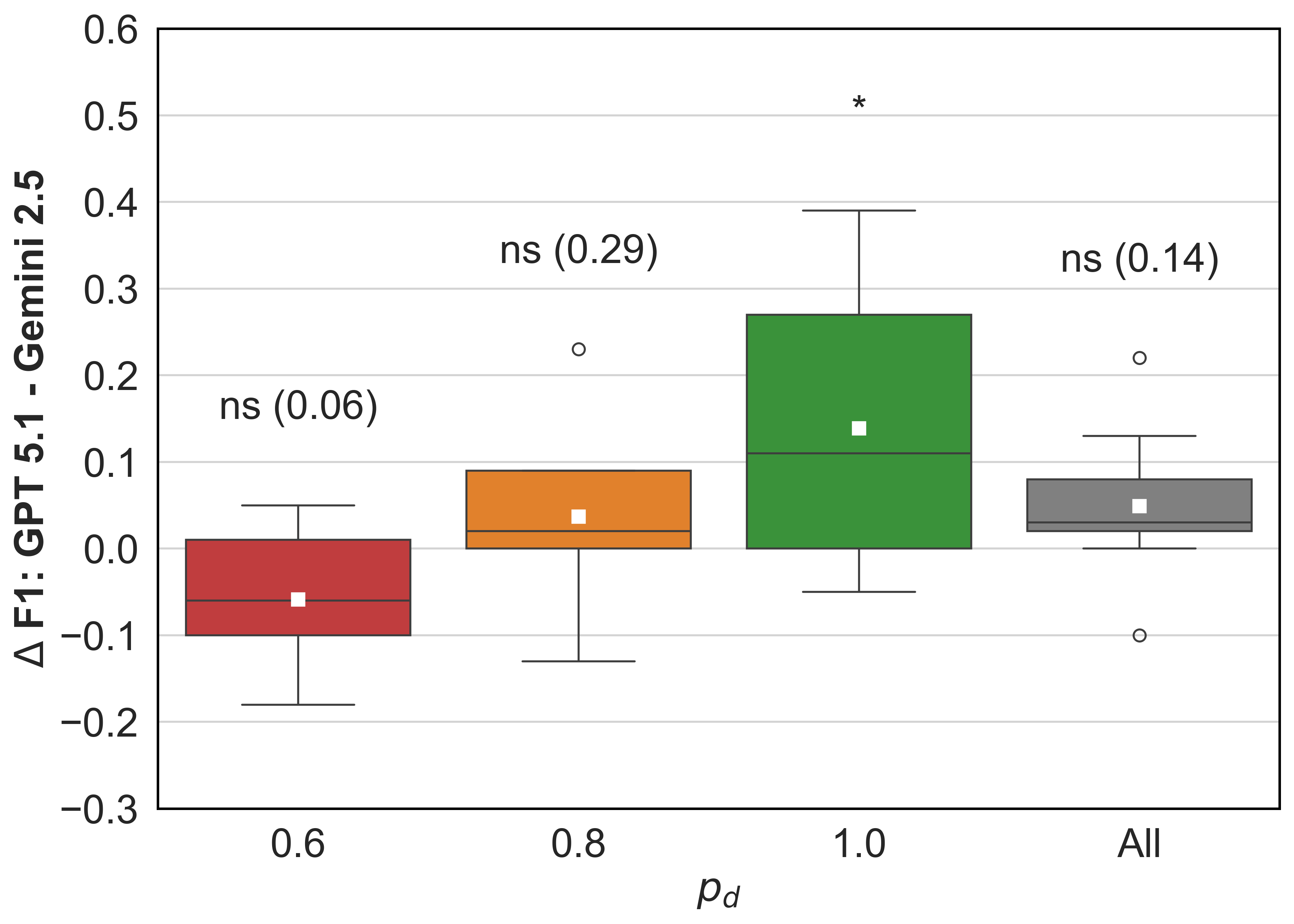}
\includegraphics[width=0.32\linewidth]{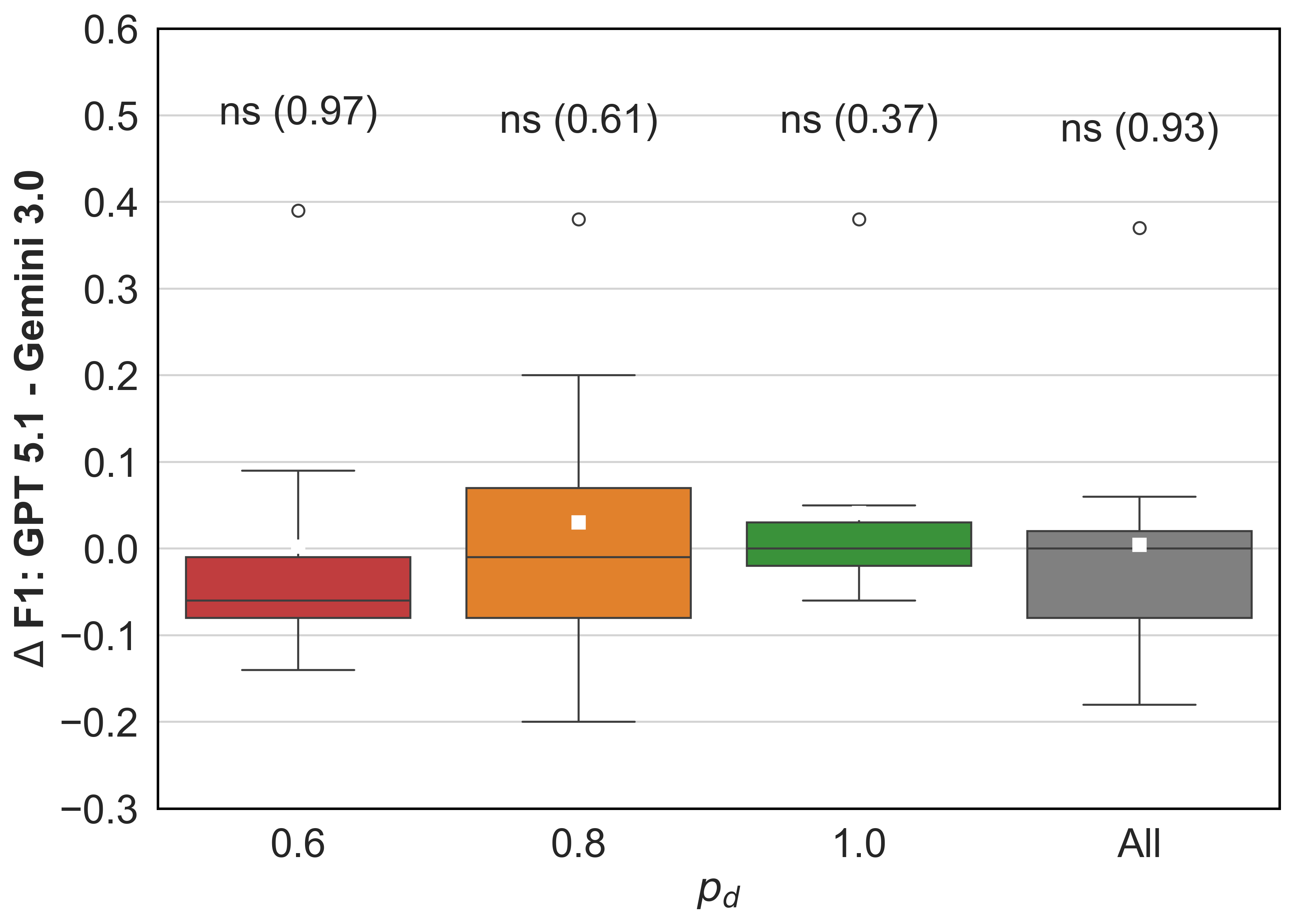}
    \caption{Model-2-Model comparison: \textbf{Left and Mid: Strong (SM) vs.\  weak model (WM)}:   The mean of $\Delta_{SM-WM}$ is less than zero at $p_d \rightarrow 0.6$, indicating that the weaker model is better than the stronger model, demonstrating the unreliability of comparison when ground truth is not certain. At $p_d \rightarrow 1.0$, $\Delta_{SM-WM}$ of stronger  model is higher than zero. \textbf{Right: Comparison of 2 SMs:}  $p_d \rightarrow 1.0$  shows that there is only marginal difference between two models but $p_d \rightarrow 0.6$  is not a clear indicator as it is not any different to weak vs.\ strong model comparison. Paired Student's T-test for statistical significance ($H_0$: No difference in performance between a pair of models).   \textbf{Note:} Improvements may  be marginal between models, hence   not statistically significant. %Previous works \cite{singh2025openaigpt5card, googlegeminieval}, rarely report   significance of improvement. }
    }
    \label{fig:strongvdweakmodel}
\end{figure}

%\vspace{-.5em}
A typical need is to compare the performance of models, including model versions.  Firstly, as shown in Figure~\ref{fig:delte}-(\textit{Right}), models' absolute performance  demonstrate a similar behavior to humans where as $p_d$ drops the absolute performance also drops. Secondly, when uncertainty is high ($p_d \rightarrow 0.6$)  the scores cannot differentiate between the capabilities of  models (as shown by low variance in accuracy across all 4 models) compared to when $p_d \rightarrow 1.0$ as shown in Figure~\ref{fig:delte}.    This also manifests when we compare Gemini-2.5 (relatively poor model compared to Gemini-3.0 and GPT-5.1),  where the paired  difference in performance between two models is less than or equal to 0 as $p_d \rightarrow 0.6$, and true performance gains with a superior model become more apparent as $p_d \rightarrow 1.0$   as shown in Figure~\ref{fig:strongvdweakmodel}.  In fact, as $p_d \rightarrow 0.6$   misleadingly indicates that Gemini-2.5 is better than Gemini-3.0 where at $p_d \rightarrow 1.0$, when Gemini-3.0 is a superior model.  In the case of comparing two strong models $\langle$GPT-5.1 and Gemini-3.0$\rangle$, while $p_d \rightarrow 0.6$ remains a non-differentiator and it is $p_d \rightarrow 1.0$ that indicates there is no substantial difference as shown in  Figure~\ref{fig:strongvdweakmodel}. We further show similar results in 3 multi-class NLI datasets across 4 models and G-Eval \cite{liu-etal-2023-g} results in Appendix~\ref{app:sec:performanceonotherdatasets}.
%\vspace{-1em}
\section{Discussion}

\subsection{Are inter rater agreement or soft label based metrics  free of this problem? }
IRA metrics such as Fleiss-$\kappa$ \cite{CTX:C6682}, Krippendorff's-$\alpha$ \cite{krippendorff2004reliability} when used to compare LLM and expert performance have also been \textit{empirically}  shown to have the same problem, where under high uncertainty a poor performer  can have similar or higher IRA with ground truth majority label compared to that of an  expert \cite{ICLR2025_87983214}. However, theoretical proof on why this happens even in IRA metrics needs further study.

We have theoretically demonstrated that when a dataset has  high ground truth variation (aka low $p_d$ across the samples) it is near impossible for an independent expert to achieve high scores when compared with majority label, akin to a fair coin toss resulting in over 90\% heads. Our work thus shows that this behavior is applicable independent of  datasets and models. Therefore when the underlying evaluation dataset contains high ground truth variation, regardless of the metric including attempting to match label variation (soft-label) instead of one hard label, an expert performance can drop to that of a random chance coin-toss resulting in heads and therefore such a dataset cannot effectively disambiguate an expert and a random labeller.
%The role of ground truth uncertainty and the performance of automated systems (such as LLMs or AI any system)  is  a subject of recent  investigations uncovering a plethora of challenges \cite{elangovan2025beyond, 10.1007/978-3-319-77249-3_6, kim2023interannotator, amidei-etal-2019-agreement}. These include  the impact of ignoring ground truth uncertainty \cite{elangovan2025beyond, plank-2022-problem, 10.1162/tacl_a_00584, STUTZ2025103556} to the applicability (\textit{or inapplicability}) and interpretation (\textit{or misinterpretations}) of typical IRA  metrics such as Cohen's-$\kappa$ \cite{Cohen1960-zn}, Krippendorff's-$\alpha$ \cite{krippendorff2004reliability} and Spearman's $\rho$ correlation.
%Fundamentally,  \textul{\textit{IRA metrics are designed for when no ground truth is available}} \cite{krippendorff2004reliability} and make assumptions about  human behavior under cognitive uncertainty to account for chance agreement \cite{doi:10.1080/23808985.2013.11679142, randolph2005free, krippendorff2004reliability}.
At the core, any aggregate metric is only  as reliable as its ability to  effectively summarize the underlying data \cite{eubanks2017re} and any gaps in the metric may be difficult to identify without detailed analysis. In summary, there are clear limitations in existing metrics as shown in Table~\ref{tab:momogram} and \ref{tab:exampleperfvar}, with respect to their ability to deal with ground truth uncertainty. Detailed analysis presents very different conclusions about relative performance as compared to what is communicated by the aggregate metric.

\subsection{Does low ground truth agreement limit  machine learning performance?}
Another question is whether low agreement  limits machine learning performance.
%KMV-20260425 Please check my rewording -- is the "in line with" bit correct? Or is it stronger, essentially label certainty and expected accuracy are correlated? (tyep looks good)
%In this paper we probabilistically show how when certainty is low quantified by $p_d$, say when $p_d\rightarrow0.6$  the expected accuracy is 60\% regardless of humans or machines especially when the sample size is large enough.
We have shown probabilistically that when label certainty is low (quantified by $p_d$), expected accuracy of any classifier is in line with that uncertainty. For $p_d\rightarrow0.6$, expected accuracy is 60\% regardless of whether humans or machines are classifying data, especially when the sample size is large enough.
Richie et.\ al.\ \cite{richie-etal-2022-inter} show that low IRA does not determine the ceiling of machine learning performance, demonstrating that models can score 67\% F1 although humans score 57\% under high uncertainty, also acknowledging that  high performance is positively correlated with high agreement. For $p_d=0.6$ and sample size 100, the probability of obtaining an accuracy  between 50--60 is $~$0.52, while obtaining between 50--70 is 97\%, so the fact that models outperform humans --- \textit{even if due to chance variation in ground truth} --- within this range is not surprising. This is also demonstrated in our empirical results, where models outperform humans under low certainty (despite the models being, in fact, quite weak compared to humans under high certainty), whilst the absolute performance for both models and humans is low under low certainty, as shown in Figure~\ref{fig:modelvshumansunderuncertainity}. Therefore our argument that under low certainty we might not be able to differentiate between an expert and non-expert holds.
%%KMV-20260425 Does this mean that performance at this level is meaningful and can demonstrate a significant difference?
With regard to absolute performance, the probability of obtaining a score over 80\%  with at least 100 samples is near zero  when $p_d\rightarrow0.6$, as shown in Table~\ref{tab:probablitypfgettinghighscore} and empirically in Figure~\ref{fig:modelvshumansunderuncertainity}. %This demonstrates how our

\section{Conclusion and Recommendations}
%\vspace{-1em}
\vspace{-.5em}
Large models are seemingly capable of achieving expert level performance \cite{Singhal2025-wx, Lee2025-yl, Tiu2022}. While typical metrics superficially might support such findings, a deeper analysis including establishing causation is often required for experiments and corresponding conclusions to be intrinsically valid and minimize the effect of confounders \cite{elangovan-etal-2024-principles, elangovan-etal-2021-memorization,Wortman1983-qg, Degtiar2023-rz}. In this paper, we demonstrated through  \textbf{(a)} the probabilistic paradigm of expected accuracy and F1, \textbf{(b)} simulations, and \textbf{(c)} empirically on 6 real world datasets and 9 models on how uncertainty associated with majority ground truth  label impacts evaluation.  In summary, we demonstrate 2 characteristics \textbf{(1)} that absolute performance numbers can be deceptively low for any type of expert (human or AI models) as a consequence of high variation in ground truth answers. As a consequence, \textbf{(2)} at high \textbf{\textit{un}certainty},  comparisons between systems tend to be unreliable and may indicate that a weak system has similar or better than  expert performance than stronger systems.
%KMV-20260425 I have rephrased here for clarity; please check that I have preserved meaning (looks good)
Such conclusions  of superior performance may be clouded by many confounders, including ground truth uncertainty.  In areas such as  radiology \cite{Tiu2022, Lee2025-yl}, where famously scientists predicted in 2016 that radiologists would  be replaced by deep neural networks  \cite{Torres2020-xx}, relative superior performance may be a mirage especially when absolute expert scores are not very high. For instance, we demonstrated probabilistically and empirically that for binary classification tasks, any performance below 80\% should not be considered a reliable basis for drawing conclusions about model quality solely based on aggregate metric and needs to account for positive class ratio $m$, sample size $S$ and stratification by ground truth uncertainty.  %For ins, taking the example of binary classification, a random labeller has a $p_r =0.5$ of selecting the right label, and given a sample size of $N=30$, the probability of achieving an 80\% accuracy or more  through  chance alone is   2\%, suggesting that scores below this level are increasingly susceptible to capability misinterpretation.

 To that effect, we make the following  recommendations  taking into account the fact that collecting multiple ground truth opinions can be expensive and time consuming.

 %for assessing the \textul{quality of evidence} presented to support the capability of any system,

\begin{enumerate}
    \item Evaluating the capability of a system requires  the  results  stratified by the probability of the  ground truth consensus when \textbf{absolute performance metrics is not  high even for experts}. The high threshold  depends on the probability of a random labeller achieving a similar score given  number of labels,  sample size  and positive class ratio   as shown in Tables~\ref{tab:probablitypfgettinghighscore}, ~\ref{tab:expectedprecisionwithclassimbalance}. This, therefore, requires that \textbf{multiple annotations become almost mandatory when overall expert performance is  not  high}.

    \item  The quality of evidence increases with 1) higher  performance of the system in high certainty bins 2) sample size of high certainty bins 3) number of ground truth answers collected per item.   The report thus must also include the probability of the ground truth   answer, ground truth annotation size, sample sizes and positive class ratio of each bin as shown in Table~\ref{tab:momogram}.
\end{enumerate}

Our work fills an important gap in interpreting variability in ground truth labels. We provide researchers with a much needed probabilistic paradigm to deal with ground truth uncertainty when measuring and comparing the performance of automated classifiers.

\bibliographystyle{plainnat}
\bibliography{references}

@ARTICLE{Torres2020-xx,
  title     = "When discussing the impact of {AI} on radiology, just remember:
               radiology is an infinite game",
  author    = "Torres, Felipe Soares",
  journal   = "Radiol. Bras.",
  publisher = "FapUNIFESP (SciELO)",
  volume    =  53,
  number    =  6,
  pages     = "VII--VIII",
  month     =  nov,
  year      =  2020,
  copyright = "http://creativecommons.org/licenses/by/4.0/",
  language  = "en"
}

@article {CTX:C6682,
	Title = {Measuring nominal scale agreement among many raters},
	Author = {Fleiss, JL},
	DOI = {10.1037/h0031619},
	Number = {5},
	Volume = {76},
	Month = {November},
	Year = {1971},
	Journal = {Psychological bulletin},
	ISSN = {0033-2909},
	Pages = {378—382},
	URL = {https://doi.org/10.1037/h0031619},
}

@article{10.2307/2346806,
    author = {Dawid, A. P. and Skene, A. M.},
    title = {Maximum Likelihood Estimation of Observer Error-Rates Using the EM Algorithm},
    journal = {Journal of the Royal Statistical Society Series C: Applied Statistics},
    volume = {28},
    number = {1},
    pages = {20-28},
    year = {1979},
    month = {03},
    issn = {0035-9254},
    doi = {10.2307/2346806},
    url = {https://doi.org/10.2307/2346806},
    eprint = {https://academic.oup.com/jrsssc/article-pdf/28/1/20/48619239/jrsssc_28_1_20.pdf},
}

@inproceedings{liu-etal-2023-g,
    title = "{G}-Eval: {NLG} Evaluation using Gpt-4 with Better Human Alignment",
    author = "Liu, Yang  and
      Iter, Dan  and
      Xu, Yichong  and
      Wang, Shuohang  and
      Xu, Ruochen  and
      Zhu, Chenguang",
    editor = "Bouamor, Houda  and
      Pino, Juan  and
      Bali, Kalika",
    booktitle = "Proceedings of the 2023 Conference on Empirical Methods in Natural Language Processing",
    month = dec,
    year = "2023",
    address = "Singapore",
    publisher = "Association for Computational Linguistics",
    url = "https://aclanthology.org/2023.emnlp-main.153/",
    doi = "10.18653/v1/2023.emnlp-main.153",
    pages = "2511--2522"
}

@inproceedings{williams-etal-2018-broad,
    title = "A Broad-Coverage Challenge Corpus for Sentence Understanding through Inference",
    author = "Williams, Adina  and
      Nangia, Nikita  and
      Bowman, Samuel",
    editor = "Walker, Marilyn  and
      Ji, Heng  and
      Stent, Amanda",
    booktitle = "Proceedings of the 2018 Conference of the North {A}merican Chapter of the Association for Computational Linguistics: Human Language Technologies, Volume 1 (Long Papers)",
    month = jun,
    year = "2018",
    address = "New Orleans, Louisiana",
    publisher = "Association for Computational Linguistics",
    url = "https://aclanthology.org/N18-1101/",
    doi = "10.18653/v1/N18-1101",
    pages = "1112--1122"
}

@book{kahneman_noise_2021,
	address = {New York},
	edition = {First edition},
	title = {Noise: a flaw in human judgment},
	isbn = {978-0-316-45140-6 978-0-316-26665-9},
	shorttitle = {Noise},
	publisher = {Little, Brown Spark},
	author = {Kahneman, Daniel and Sibony, Olivier and Sunstein, Cass R.},
	year = {2021},
	note = {OCLC: on1249942231},
}

@article{kahneman1982variants,
  title={Variants of uncertainty},
  author={Kahneman, Daniel and Tversky, Amos},
  journal={Cognition},
  volume={11},
  number={2},
  pages={143--157},
  year={1982},
  publisher={Elsevier}
}

@inproceedings{richie-etal-2022-inter,
    title = "Inter-annotator agreement is not the ceiling of machine learning performance: Evidence from a comprehensive set of simulations",
    author = "Richie, Russell  and
      Grover, Sachin  and
      Tsui, Fuchiang (Rich)",
    editor = "Demner-Fushman, Dina  and
      Cohen, Kevin Bretonnel  and
      Ananiadou, Sophia  and
      Tsujii, Junichi",
    booktitle = "Proceedings of the 21st Workshop on Biomedical Language Processing",
    month = may,
    year = "2022",
    address = "Dublin, Ireland",
    publisher = "Association for Computational Linguistics",
    url = "https://aclanthology.org/2022.bionlp-1.26/",
    doi = "10.18653/v1/2022.bionlp-1.26",
    pages = "275--284"
}

@article{eubanks2017re,
  title={(Re) visualizing rater agreement: Beyond single-parameter measures},
  author={Eubanks, David},
  journal={Journal of Writing Analytics},
  volume={1},
  year={2017}
}

@article{krippendorff2004reliability,
  title={Reliability in content analysis: Some common misconceptions and recommendations},
  author={Krippendorff, Klaus},
  journal={Human communication research},
  volume={30},
  number={3},
  pages={411--433},
  year={2004},
  publisher={Wiley Online Library}
}

@ARTICLE{Degtiar2023-rz,
  title     = "A review of generalizability and transportability",
  author    = "Degtiar, Irina and Rose, Sherri",
  journal   = "Annu. Rev. Stat. Appl.",
  publisher = "Annual Reviews",
  volume    =  10,
  number    =  1,
  pages     = "501--524",
  month     =  mar,
  year      =  2023,
  language  = "en"
}

@ARTICLE{Wortman1983-qg,
  title     = "Evaluation research: A methodological perspective",
  author    = "Wortman, P M",
  journal   = "Annu. Rev. Psychol.",
  publisher = "Annual Reviews",
  volume    =  34,
  number    =  1,
  pages     = "223--260",
  month     =  jan,
  year      =  1983,
  language  = "en"
}

@misc{Esty_2024, title={Navigating the uncertainties of medicine}, url={https://magazine.hms.harvard.edu/articles/navigating-uncertainties-medicine}, journal={Harvard Medicine Magazine}, author={Esty, Amos}, year={2024}, month={Jul}}

@Article{Tiu2022,
author={Tiu, Ekin
and Talius, Ellie
and Patel, Pujan
and Langlotz, Curtis P.
and Ng, Andrew Y.
and Rajpurkar, Pranav},
title={Expert-level detection of pathologies from unannotated chest X-ray images via self-supervised learning},
journal={Nature Biomedical Engineering},
year={2022},
month={Dec},
day={01},
volume={6},
number={12},
pages={1399-1406},
issn={2157-846X},
doi={10.1038/s41551-022-00936-9},
url={https://doi.org/10.1038/s41551-022-00936-9}
}

@article{STUTZ2025103556,
title = {Evaluating medical AI systems in dermatology under uncertain ground truth},
journal = {Medical Image Analysis},
volume = {103},
pages = {103556},
year = {2025},
issn = {1361-8415},
doi = {https://doi.org/10.1016/j.media.2025.103556},
url = {https://www.sciencedirect.com/science/article/pii/S1361841525001033},
author = {David Stutz and Ali Taylan Cemgil and Abhijit Guha Roy and Tatiana Matejovicova and Melih Barsbey and Patricia Strachan and Mike Schaekermann and Jan Freyberg and Rajeev Rikhye and Beverly Freeman and Javier Perez Matos and Umesh Telang and Dale R. Webster and Yuan Liu and Greg S. Corrado and Yossi Matias and Pushmeet Kohli and Yun Liu and Arnaud Doucet and Alan Karthikesalingam}
}

@ARTICLE{Bhise2018-ku,
  title     = "Defining and measuring diagnostic uncertainty in medicine: A
               systematic review",
  author    = "Bhise, Viraj and Rajan, Suja S and Sittig, Dean F and Morgan,
               Robert O and Chaudhary, Pooja and Singh, Hardeep",
  journal   = "J. Gen. Intern. Med.",
  publisher = "Springer Science and Business Media LLC",
  volume    =  33,
  number    =  1,
  pages     = "103--115",
  month     =  jan,
  year      =  2018,
  language  = "en"
}

@ARTICLE{Han2021-la,
  title     = "How physicians manage medical uncertainty: A qualitative study
               and conceptual taxonomy",
  author    = "Han, Paul K J and Strout, Tania D and Gutheil, Caitlin and
               Germann, Carl and King, Brian and Ofstad, Eirik and Gulbrandsen,
               P{\aa}l and Trowbridge, Robert",
  journal   = "Med. Decis. Making",
  publisher = "SAGE Publications",
  volume    =  41,
  number    =  3,
  pages     = "275--291",
  month     =  apr,
  year      =  2021,
  language  = "en"
}

@book{gawande2003complications,
  title={Complications: A Surgeon's Notes on an Imperfect Science},
  author={Gawande, A.},
  isbn={9781429972109},
  url={https://books.google.com/books?id=8dkuoAz9-WMC},
  year={2003},
  publisher={Henry Holt and Company}
}

@article{10.1001/archinte.168.7.741,
    author = {Green, Sandy M. and Martinez-Rumayor, Abelardo and Gregory, Shawn A. and Baggish, Aaron L. and O’Donoghue, Michelle L. and Green, Jamie A. and Lewandrowski, Kent B. and Januzzi, James L., Jr},
    title = {Clinical Uncertainty, Diagnostic Accuracy, and Outcomes in Emergency Department Patients Presenting With Dyspnea},
    journal = {Archives of Internal Medicine},
    volume = {168},
    number = {7},
    pages = {741-748},
    year = {2008},
    month = {04},
    issn = {0003-9926},
    doi = {10.1001/archinte.168.7.741},
    url = {https://doi.org/10.1001/archinte.168.7.741},
    eprint = {https://jamanetwork.com/journals/jamainternalmedicine/articlepdf/414143/ioi70246\_741\_748.pdf},
}

@Article{Lichtstein2023,
author={Lichtstein, Daniel M.},
title={Strategies to Deal with Uncertainty in Medicine},
journal={The American Journal of Medicine},
year={2023},
month={Apr},
day={01},
publisher={Elsevier},
volume={136},
number={4},
pages={339-340},
issn={0002-9343},
doi={10.1016/j.amjmed.2022.12.018},
url={https://doi.org/10.1016/j.amjmed.2022.12.018}
}

@ARTICLE{Kim2018-wz,
  title     = "Understanding uncertainty in medicine: concepts and implications
               in medical education",
  author    = "Kim, Kangmoon and Lee, Young-Mee",
  journal   = "Korean J. Med. Educ.",
  publisher = "Korean Society of Medical Education",
  volume    =  30,
  number    =  3,
  pages     = "181--188",
  month     =  sep,
  year      =  2018,
  language  = "en"
}

@inproceedings{plank-2022-problem,
    title = "The ``Problem'' of Human Label Variation: On Ground Truth in Data, Modeling and Evaluation",
    author = "Plank, Barbara",
    editor = "Goldberg, Yoav  and
      Kozareva, Zornitsa  and
      Zhang, Yue",
    booktitle = "Proceedings of the 2022 Conference on Empirical Methods in Natural Language Processing",
    month = dec,
    year = "2022",
    address = "Abu Dhabi, United Arab Emirates",
    publisher = "Association for Computational Linguistics",
    url = "https://aclanthology.org/2022.emnlp-main.731/",
    doi = "10.18653/v1/2022.emnlp-main.731",
    pages = "10671--10682"
}

@inproceedings{ICLR2025_87983214,
 author = {Elangovan, Aparna and Xu, Lei and Ko, Jongwoo and Elyasi, Mahsa and Liu, Ling and Bodapati, Sravan Babu and Roth, Dan},
 booktitle = {International Conference on Learning Representations},
 editor = {Y. Yue and A. Garg and N. Peng and F. Sha and R. Yu},
 pages = {54602--54631},
 title = {Beyond correlation: The impact of human uncertainty in measuring the effectiveness of automatic evaluation and LLM-as-a-judge},
 url = {https://proceedings.iclr.cc/paper_files/paper/2025/file/8798321486948322be2b4d658744ba72-Paper-Conference.pdf},
 volume = {2025},
 year = {2025}
}

@inproceedings{fc4b0d04b6c84ca18949b746f604eb0b,
title = "Human uncertainty makes classification more robust",
author = "Joshua Peterson and Ruairidh Battleday and Thomas Griffiths and Olga Russakovsky",
note = "Publisher Copyright: {\textcopyright} 2019 IEEE.; 17th IEEE/CVF International Conference on Computer Vision, ICCV 2019 ; Conference date: 27-10-2019 Through 02-11-2019",
year = "2019",
month = oct,
doi = "10.1109/ICCV.2019.00971",
language = "English (US)",
series = "Proceedings of the IEEE International Conference on Computer Vision",
publisher = "Institute of Electrical and Electronics Engineers Inc.",
pages = "9616--9625",
booktitle = "Proceedings - 2019 International Conference on Computer Vision, ICCV 2019",
address = "United States",

}

@ARTICLE{Eisemann2025-gq,
  title     = "Nationwide real-world implementation of {AI} for cancer
               detection in population-based mammography screening",
  author    = "Eisemann, Nora and Bunk, Stefan and Mukama, Trasias and Baltus,
               Hannah and Elsner, Susanne A and Gomille, Timo and Hecht, Gerold
               and Heywang-K{\"o}brunner, Sylvia and Rathmann, Regine and
               Siegmann-Luz, Katja and T{\"o}llner, Thilo and Vomweg, Toni
               Werner and Leibig, Christian and Katalinic, Alexander",
  journal   = "Nat. Med.",
  publisher = "Springer Science and Business Media LLC",
  volume    =  31,
  number    =  3,
  pages     = "917--924",
  month     =  mar,
  year      =  2025,
  copyright = "https://creativecommons.org/licenses/by/4.0",
  language  = "en"
}

@ARTICLE{Irvin2019-im,
  title     = "{CheXpert}: Chest {X-Rays}",
  author    = "Irvin, Jeremy and Rajpurkar, Pranav and Ko, Michael and Yu,
               Yifan and Ciurea-Ilcus, Silviana and Chute, Chris and Marklund,
               Henrik and Haghgoo, Behzad and Ball, Robyn and Shpanskaya, Katie
               and Seekins, Jayne and Mong, David A and Halabi, Safwan S and
               Sandberg, Jesse K and Jones, Ricky and Larson, David B and
               Langlotz, Curtis P and Patel, Bhavik N and Lungren, Matthew P
               and Ng, Andrew Y",
  publisher = "Center for Artificial Intelligence in Medicine and Imaging",
  year      =  2019
}

@ARTICLE{Singhal2025-wx,
  title     = "Toward expert-level medical question answering with large
               language models",
  author    = "Singhal, Karan and Tu, Tao and Gottweis, Juraj and Sayres, Rory
               and Wulczyn, Ellery and Amin, Mohamed and Hou, Le and Clark,
               Kevin and Pfohl, Stephen R and Cole-Lewis, Heather and Neal,
               Darlene and Rashid, Qazi Mamunur and Schaekermann, Mike and
               Wang, Amy and Dash, Dev and Chen, Jonathan H and Shah, Nigam H
               and Lachgar, Sami and Mansfield, Philip Andrew and Prakash,
               Sushant and Green, Bradley and Dominowska, Ewa and Ag{\"u}era Y
               Arcas, Blaise and Toma{\v s}ev, Nenad and Liu, Yun and Wong,
               Renee and Semturs, Christopher and Mahdavi, S Sara and Barral,
               Joelle K and Webster, Dale R and Corrado, Greg S and Matias,
               Yossi and Azizi, Shekoofeh and Karthikesalingam, Alan and
               Natarajan, Vivek",
  journal   = "Nat. Med.",
  publisher = "Springer Science and Business Media LLC",
  volume    =  31,
  number    =  3,
  pages     = "943--950",
  month     =  mar,
  year      =  2025,
  copyright = "https://creativecommons.org/licenses/by-nc-nd/4.0",
  language  = "en"
}

@inproceedings{snli:emnlp2015,
	Author = {Bowman, Samuel R. and Angeli, Gabor and Potts, Christopher and Manning, Christopher D.},
	Booktitle = {Proceedings of the 2015 Conference on Empirical Methods in Natural Language Processing (EMNLP)},
	Publisher = {Association for Computational Linguistics},
	Title = {A large annotated corpus for learning natural language inference},
	Year = {2015}
}

@inproceedings{elangovan-etal-2021-memorization,
    title = "Memorization vs. Generalization : Quantifying Data Leakage in {NLP} Performance Evaluation",
    author = "Elangovan, Aparna  and
      He, Jiayuan  and
      Verspoor, Karin",
    editor = "Merlo, Paola  and
      Tiedemann, Jorg  and
      Tsarfaty, Reut",
    booktitle = "Proceedings of the 16th Conference of the European Chapter of the Association for Computational Linguistics: Main Volume",
    month = apr,
    year = "2021",
    address = "Online",
    publisher = "Association for Computational Linguistics",
    url = "https://aclanthology.org/2021.eacl-main.113/",
    doi = "10.18653/v1/2021.eacl-main.113",
    pages = "1325--1335"
}

@inproceedings{elangovan-etal-2024-principles,
    title = "Principles from Clinical Research for {NLP} Model Generalization",
    author = "Elangovan, Aparna  and
      He, Jiayuan  and
      Li, Yuan  and
      Verspoor, Karin",
    editor = "Duh, Kevin  and
      Gomez, Helena  and
      Bethard, Steven",
    booktitle = "Proceedings of the 2024 Conference of the North American Chapter of the Association for Computational Linguistics: Human Language Technologies (Volume 1: Long Papers)",
    month = jun,
    year = "2024",
    address = "Mexico City, Mexico",
    publisher = "Association for Computational Linguistics",
    url = "https://aclanthology.org/2024.naacl-long.127/",
    doi = "10.18653/v1/2024.naacl-long.127",
    pages = "2293--2309"
}

@ARTICLE{Lee2025-yl,
  title     = "{CXR-LLaVA}: a multimodal large language model for interpreting
               chest X-ray images",
  author    = "Lee, Seowoo and Youn, Jiwon and Kim, Hyungjin and Kim, Mansu and
               Yoon, Soon Ho",
  journal   = "Eur. Radiol.",
  publisher = "Springer Science and Business Media LLC",
  volume    =  35,
  number    =  7,
  pages     = "4374--4386",
  month     =  jul,
  year      =  2025,
  copyright = "https://creativecommons.org/licenses/by/4.0",
  language  = "en"
}

\newpage
\begin{appendices}

\section{Limitations}\label{sec:limitations}
Our work focuses on the  ground truth uncertainty and its associated probability for subjective use cases. \textit{Note:} In the case  of objective use case the ground truth probability can be treated as 1.0.  The primary assumption we make is that the probability of ground truth is computed based on the percentage of vote it receives, using the  majority label as the ground truth which is a common practice. However, there maybe other ways of deriving ground truth, such as EM-based estimation \cite{10.2307/2346806} and hence the associated computation of ground truth probability changes. We treat this as future work, to compute the probability of ground truth when using  methods  alternative to majority label. We have also stated this majority label assumption in Section~\ref{sec:statisticalexplanation}.

\section{Prompts}
The models were prompted as follows ``\textit{Given the patient's chest x-ray, answer the following questions: Does this patient have \{\{condition \}\}?}'' with default temperature settings.

\section{Simulation details}
The simulation was conducted by creating 10 annotations  per item with varying $p_d$, where each annotation was 1 or 0 mimicking binary classification problems.  The sample size of the simulation was sampling 500 items per run  and controlled for positive class ratio $m$.

\section{Label distribution in CheXpert dataset}\label{labeldistribution}

\begin{figure}[h!]
    \centering
    \includegraphics[width=0.5\linewidth]{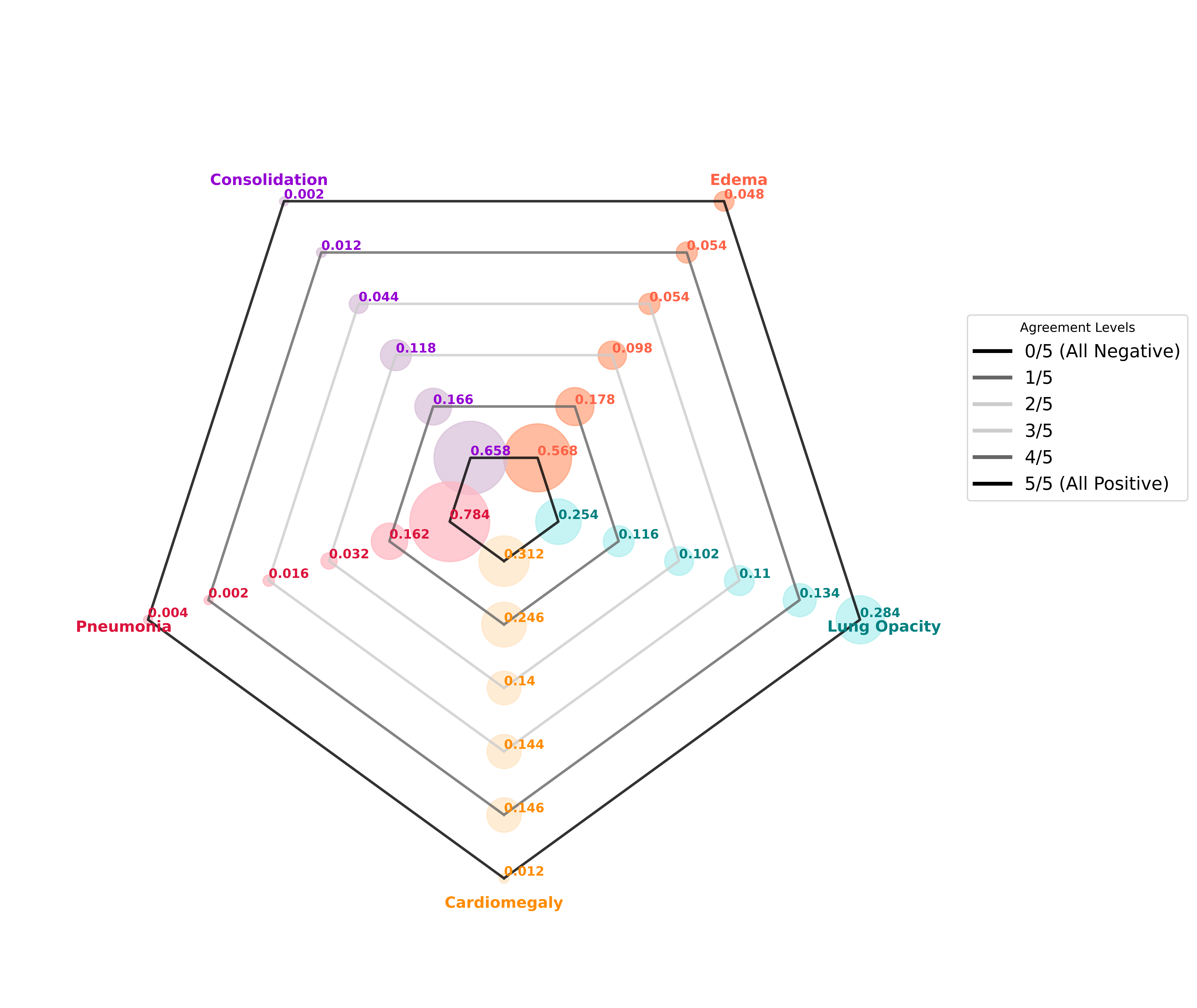}
    \caption{Proportion of samples (CheXpert test set with 500 samples) by   agreement level as annotated by 5 radiologists independently. The number of samples with 100\% agreement ($p_d=1.0$) on positive findings drops dramatically to less than 0.01 for 4 of 5 the pathologies. The exception is lung opacity, where the proportion of samples approximately remains the same regardless  of agreement levels. Agreement level for all  pathologies are in Appendix~\ref{fig:chextpertagreement}.}
    \label{fig:uncertaintyhexagon}
\end{figure}

\begin{figure}[h!]
    \centering
    \includegraphics[width=0.9\linewidth]{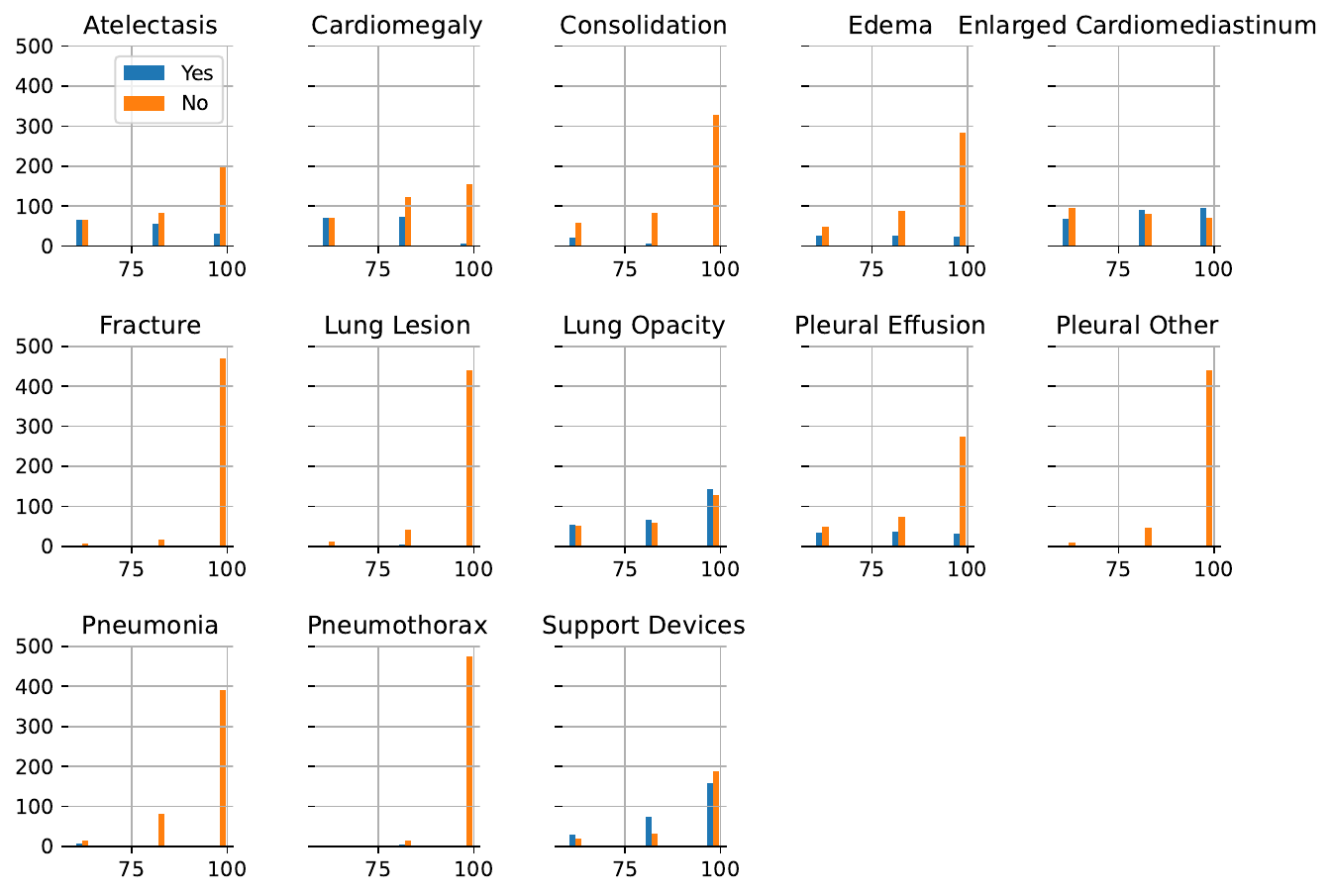}

        \caption{Chest x ray agreement for binary classification across 13 labels. The agreement is across 5 radiologists. The x-axis shows the percentage agreement for the majority label ( yes/ no)  and the corresponding y-axis indicate the total number of records with that agreement. For instance, for Atelectasis, 100\% agreement ( 5 out of 5 radiologists) is observed for 32 records for a positive finding (yes), while 100\% agreement for negative finding for 200 records. }
    \label{fig:chextpertagreement}
\end{figure}

\newpage
\section{Performance on other datasets}\label{app:sec:performanceonotherdatasets}

\subsection{Stratified Performance on NLI}
\begin{figure}[h!]
    \centering
    \includegraphics[width=0.5\linewidth]{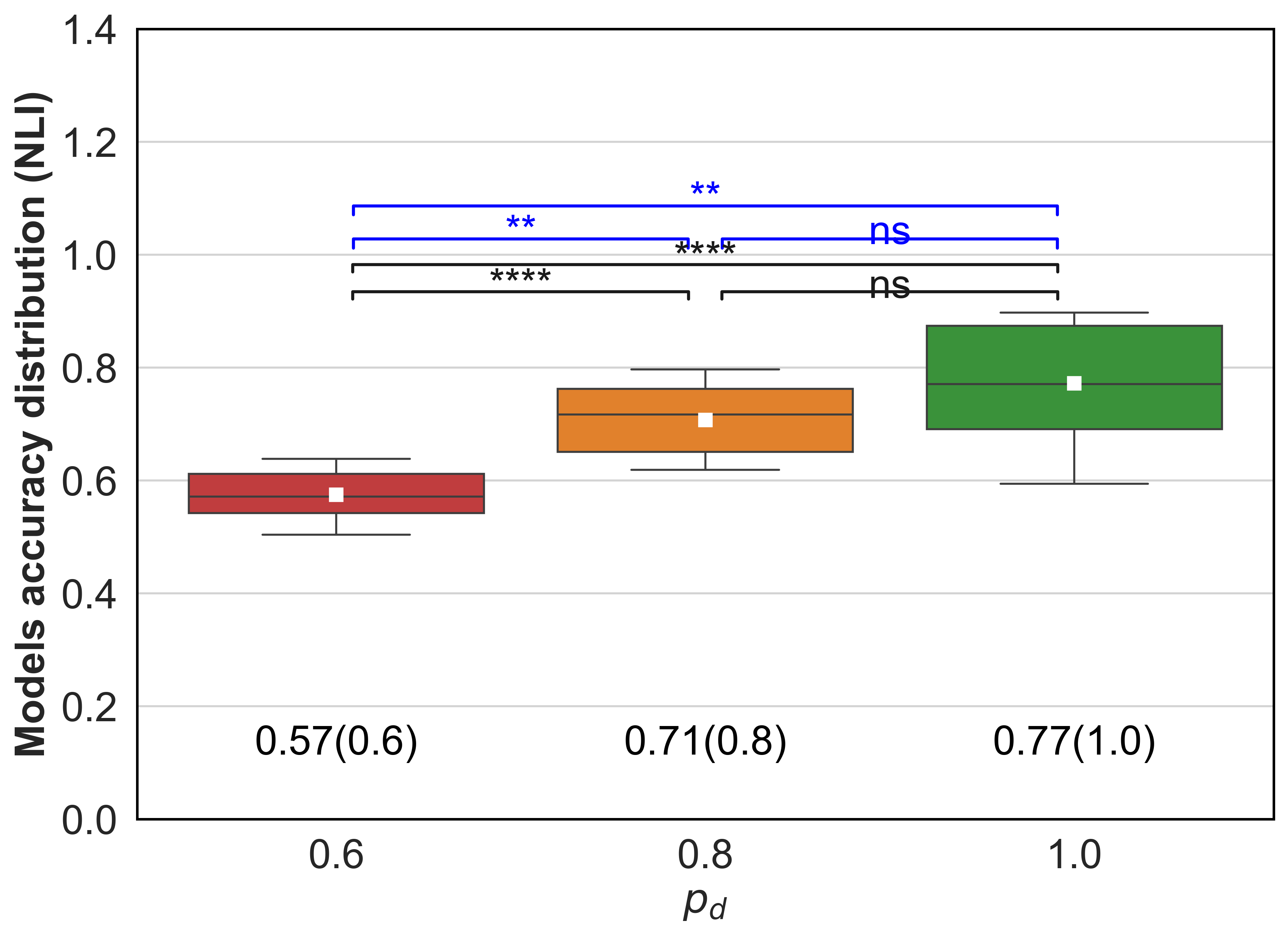}
    \caption{MNLI Matched, MNLI Mismatched and SNLI performance (Accuracy) distribution by $p_d$ on 4 models - Mistral, Mistral Large, Sonnet and Llama 3.1. These are existing results from Elangovan et. al.  \cite{ICLR2025_87983214}. The variation in scores is low when $p_d \rightarrow 0.6$   compared to $p_d \rightarrow 1.0$, as demonstrated by the statistical significance using \textcolor{blue}{Levene's test} for variance. This shows the lack of differentiating capability between strong vs. weak models when $p_d \rightarrow 0.6$. The mean performance at $p_d \rightarrow 1.0$ is also higher (statistically significant using Welch's t-test) compared to $p_d \rightarrow 0.6$ Detailed tabular results are available in Table~\ref{app:tab:nli}.}
    \label{app:fig:mulitmodelNLI}
\end{figure}

\begin{table}[h!]
    \centering
    % \begin{tabular}{lrrrlrrr}
    % \hline
    % Dataset & $p_d$ & S & $E(A)$  & $\text{H}'$ & $\Delta_{E\text{H}'}$ & $M$ & $\Delta_{EM}$ \\
    % \hline
\begin{tabular}{lllllllrl}
\toprule
Dataset & $p_d$ & S &  $E(A)$ & $\text{H}'$ & $\Delta_{E\text{H}'}$  & Model & A &  $\Delta_{EM}$ \\
\midrule
\multirow[t]{12}{*}{MNLI M  $\langle$GT = 5, L=3$\rangle$} & \multirow[t]{4}{*}{0.60} & \multirow[t]{4}{*}{1599} & \multirow[t]{4}{*}{0.60} & \multirow[t]{4}{*}{0.61$\pm$0.02} & \multirow[t]{4}{*}{-0.01} & Llama3.1 & 0.55 & 0.05 \\
 &  &  &  &  &  & Mistral & 0.50 & 0.10 \\
 &  &  &  &  &  & Mistral-L & 0.59 & 0.01 \\
 &  &  &  &  &  & Sonnet & 0.61 & -0.01 \\
\cline{2-9} \cline{3-9} \cline{4-9} \cline{5-9} \cline{6-9}
 & \multirow[t]{4}{*}{0.80} & \multirow[t]{4}{*}{2457} & \multirow[t]{4}{*}{0.80} & \multirow[t]{4}{*}{0.81$\pm$0.01} & \multirow[t]{4}{*}{-0.01} & Llama3.1 & 0.68 & 0.12 \\
 &  &  &  &  &  & Mistral & 0.63 & 0.17 \\
 &  &  &  &  &  & Mistral-L & 0.75 & 0.05 \\
 &  &  &  &  &  & Sonnet & 0.78 & 0.02 \\
\cline{2-9} \cline{3-9} \cline{4-9} \cline{5-9} \cline{6-9}
 & \multirow[t]{4}{*}{1.00} & \multirow[t]{4}{*}{5759} & \multirow[t]{4}{*}{1.00} & \multirow[t]{4}{*}{1.00$\pm$0.00} & \multirow[t]{4}{*}{0.00} & Llama3.1 & 0.75 & 0.25 \\
 &  &  &  &  &  & Mistral & 0.69 & 0.31 \\
 &  &  &  &  &  & Mistral-L & 0.87 & 0.13 \\
 &  &  &  &  &  & Sonnet & 0.89 & 0.11 \\
\cline{1-9} \cline{2-9} \cline{3-9} \cline{4-9} \cline{5-9} \cline{6-9}
\multirow[t]{12}{*}{MNLI MM $\langle$GT = 5, L=3$\rangle$} & \multirow[t]{4}{*}{0.60} & \multirow[t]{4}{*}{1446} & \multirow[t]{4}{*}{0.60} & \multirow[t]{4}{*}{0.60$\pm$0.01} & \multirow[t]{4}{*}{0.00} & Llama3.1 & 0.55 & 0.05 \\
 &  &  &  &  &  & Mistral & 0.52 & 0.08 \\
 &  &  &  &  &  & Mistral-L & 0.61 & -0.01 \\
 &  &  &  &  &  & Sonnet & 0.62 & -0.02 \\
\cline{2-9} \cline{3-9} \cline{4-9} \cline{5-9} \cline{6-9}
 & \multirow[t]{4}{*}{0.80} & \multirow[t]{4}{*}{2402} & \multirow[t]{4}{*}{0.80} & \multirow[t]{4}{*}{0.80$\pm$0.00} & \multirow[t]{4}{*}{-0.00} & Llama3.1 & 0.66 & 0.14 \\
 &  &  &  &  &  & Mistral & 0.62 & 0.18 \\
 &  &  &  &  &  & Mistral-L & 0.75 & 0.05 \\
 &  &  &  &  &  & Sonnet & 0.77 & 0.03 \\
\cline{2-9} \cline{3-9} \cline{4-9} \cline{5-9} \cline{6-9}
 & \multirow[t]{4}{*}{1.00} & \multirow[t]{4}{*}{5984} & \multirow[t]{4}{*}{1.00} & \multirow[t]{4}{*}{1.00$\pm$0.00} & \multirow[t]{4}{*}{0.00} & Llama3.1 & 0.75 & 0.25 \\
 &  &  &  &  &  & Mistral & 0.69 & 0.31 \\
 &  &  &  &  &  & Mistral-L & 0.88 & 0.12 \\
 &  &  &  &  &  & Sonnet & 0.90 & 0.10 \\
\cline{1-9} \cline{2-9} \cline{3-9} \cline{4-9} \cline{5-9} \cline{6-9}
\multirow[t]{16}{*}{SNLI MM $\langle GT$ $\in \{5, 4^\dagger\}$, L=3$\rangle$} & \multirow[t]{4}{*}{0.60} & \multirow[t]{4}{*}{1507} & \multirow[t]{4}{*}{0.60} & \multirow[t]{4}{*}{0.61$\pm$0.00} & \multirow[t]{4}{*}{-0.01} & Llama3.1 & 0.54 & 0.06 \\
 &  &  &  &  &  & Mistral & 0.54 & 0.06 \\
 &  &  &  &  &  & Mistral-L & 0.64 & -0.04 \\
 &  &  &  &  &  & Sonnet & 0.61 & -0.01 \\
\cline{2-9} \cline{3-9} \cline{4-9} \cline{5-9} \cline{6-9}
 & \multirow[t]{4}{*}{0.75$^\dagger$} & \multirow[t]{4}{*}{7} & \multirow[t]{4}{*}{0.75} & \multirow[t]{4}{*}{0.71$\pm$0.14} & \multirow[t]{4}{*}{0.04} & Llama3.1 & 0.71 & 0.04 \\
 &  &  &  &  &  & Mistral & 0.57 & 0.18 \\
 &  &  &  &  &  & Mistral-L & 0.86 & -0.11 \\
 &  &  &  &  &  & Sonnet & 0.71 & 0.04 \\
\cline{2-9} \cline{3-9} \cline{4-9} \cline{5-9} \cline{6-9}
 & \multirow[t]{4}{*}{0.80} & \multirow[t]{4}{*}{2845} & \multirow[t]{4}{*}{0.80} & \multirow[t]{4}{*}{0.80$\pm$0.00} & \multirow[t]{4}{*}{0.00} & Llama3.1 & 0.66 & 0.14 \\
 &  &  &  &  &  & Mistral & 0.62 & 0.18 \\
 &  &  &  &  &  & Mistral-L & 0.80 & 0.00 \\
 &  &  &  &  &  & Sonnet & 0.76 & 0.04 \\
\cline{2-9} \cline{3-9} \cline{4-9} \cline{5-9} \cline{6-9}
 & \multirow[t]{4}{*}{1.00} & \multirow[t]{4}{*}{5483} & \multirow[t]{4}{*}{1.00} & \multirow[t]{4}{*}{1.00$\pm$0.00} & \multirow[t]{4}{*}{0.00} & Llama3.1 & 0.65 & 0.35 \\
 &  &  &  &  &  & Mistral & 0.59 & 0.41 \\
 &  &  &  &  &  & Mistral-L & 0.81 & 0.19 \\
 &  &  &  &  &  & Sonnet & 0.80 & 0.20 \\
\cline{1-9} \cline{2-9} \cline{3-9} \cline{4-9} \cline{5-9} \cline{6-9}

\end{tabular}
    \caption{Expected accuracy (\textbf{A}) vs empirical accuracy. Here M is the existing result on Models  by Elangovan et al \cite{ICLR2025_87983214}. \textbf{GT} is the number of ground truth annotations and $L$ is the number of labels. \text{H}' is the simulated performance of another Human.  $\Delta_{EM} = E(A) - M$, $\Delta_{E\text{H}'} = E(A) - \text{H}'$. $^\dagger$Only 7 samples (in $p_d \rightarrow 0.75$) have 4 annotations in the SNLI dataset, the rest have 5.}
    \label{app:tab:nli}
\end{table}

\newpage
\subsection{Stratified performance on GEval}

\begin{figure}
    \centering
    \includegraphics[width=0.5\linewidth]{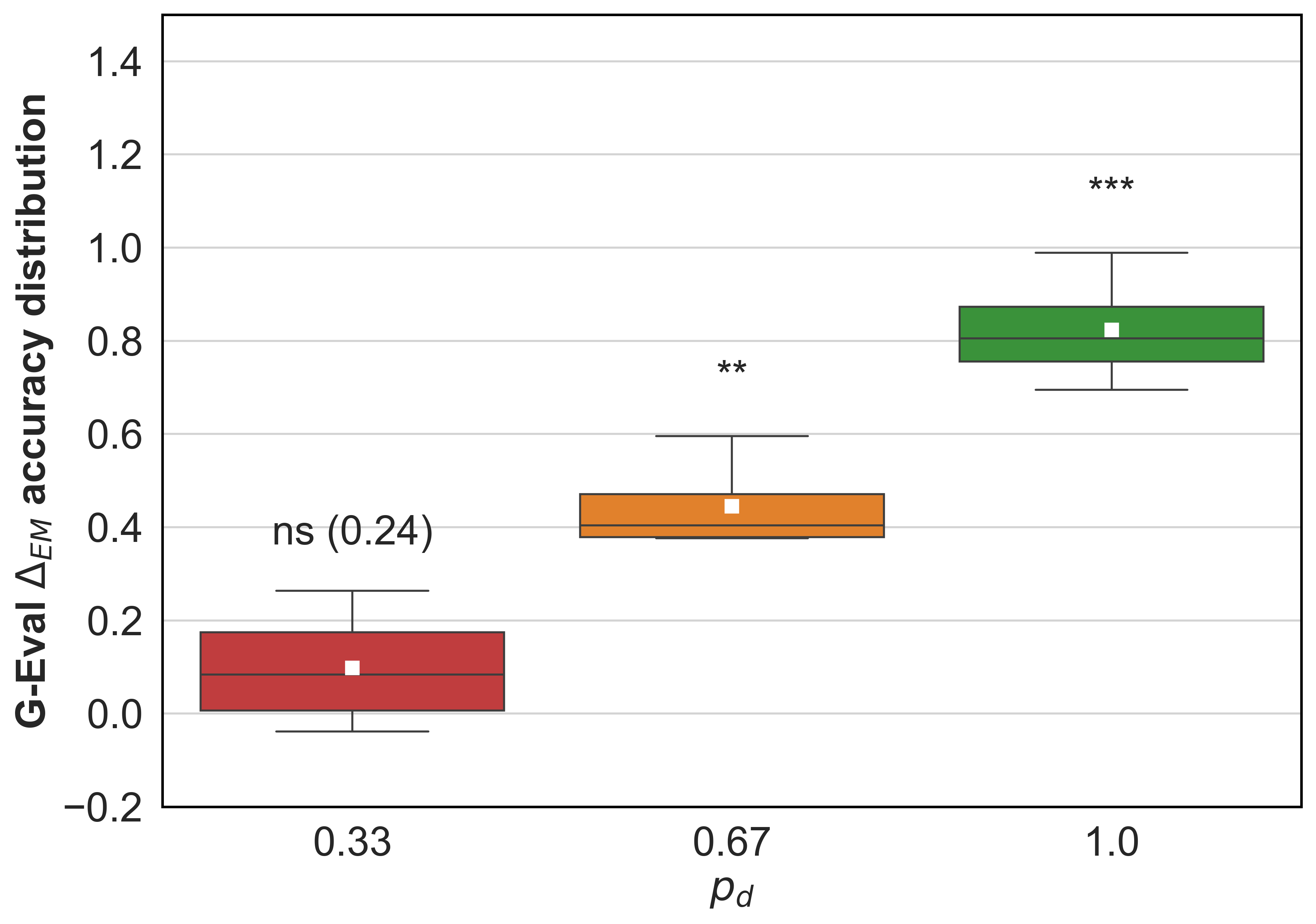}
    \caption{Difference in expected vs.\ actual model performance of G-Eval original results \cite{liu-etal-2023-g} for 4 criteria - Coherence, relevance, consistency and fluency. Ground truth  is the majority label. Model performance is the most frequent label from the model.  $\Delta_{EM}$ is close to zero $p_d \rightarrow 0.33$, the gap between expected and the model's substantially lower performance  becomes apparent as $p_d \rightarrow 1.0$. Detailed tabular performance in Appendix Table~\ref{app:tab:GEval}. Examples of differences between G-Eval and human experts are shown in Appendix Table~\ref{app:tab:geval_incosistency_Example}. Paired (expected vs. actual) T-test for significance. }
    \label{fig:placeholder}
\end{figure}

\begin{table}[h!]
    \begin{tabular}{llrrlr}
\toprule

Criteria & $p_d$ & $S$ & E(A) & Score & $\Delta_{EM}$  \\
\midrule
\multirow[t]{3}{*}{Coherence $\langle GT$ =3, L=5$\rangle$} & 0.33 & 341 & 0.33 & 0.185 & 0.145 \\
\cline{2-6}
 & 0.67 & 1059 & 0.67 & 0.291 & 0.379 \\
\cline{2-6}
 & 1.00 & 200 & 1.00 & 0.165 & 0.835 \\
\cline{1-6} \cline{2-6}
\multirow[t]{3}{*}{Consistency $\langle GT$ =3, L=5$\rangle$} & 0.33 & 19 & 0.33 & 0.368 & -0.038 \\
\cline{2-6}
 & 0.67 & 241 & 0.67 & 0.241 & 0.429 \\
\cline{2-6}
 & 1.00 & 1340 & 1.00 & 0.305 & 0.695 \\
\cline{1-6} \cline{2-6}
\multirow[t]{3}{*}{Fluency $\langle GT$ =3, L=5$\rangle$} & 0.33 & 45 & 0.33 & 0.067 & 0.263 \\
\cline{2-6}
 & 0.67 & 364 & 0.67 & 0.074 & 0.596 \\
\cline{2-6}
 & 1.00 & 1191 & 1.00 & 0.011 & 0.989 \\
\cline{1-6} \cline{2-6}
\multirow[t]{3}{*}{Relevance $\langle GT$ =3, L=5$\rangle$} & 0.33 & 315 & 0.33 & 0.308 & 0.022 \\
\cline{2-6}
 & 0.67 & 1057 & 0.67 & 0.293 & 0.377 \\
\cline{2-6}
 & 1.00 & 228 & 1.00 & 0.224 & 0.776 \\
\cline{1-6} \cline{2-6}
\end{tabular}
    \caption{Accuracy of GEval: Model performance is the most frequent label from the model. Human ground truth is also the majority label. $\Delta_{EM}$ narrows down the most as $p_d \rightarrow 0.33$}
    \label{app:tab:GEval}
\end{table}

\begin{table}[h!]
    \centering
    {\tiny
   \begin{tabular}{llrll}
\toprule
  Id & ModelId & $p_d$ & Experts & Model \\
\midrule
dm-test-9b5a04b9879fe716a2dcd66dd331f99534603c16 & M22 & 1.00 & [5, 5, 5] & [' 4', ' 4', ' 3', ' 4', ' 3', ' 4', ' 3', ' 4', ' 3', ' 4', ' 4', ' 3', ' 4', ' 4', ' 4', ' 4', ' 3', ' 3', ' 3', ' 4'] \\
dm-test-9beaaf73e4517d5bf1bc284f386485d571da96cf & M15 & 1.00 & [5, 5, 5] & [' 3', ' 3', ' 2', ' 3', ' 4', ' 3', ' 3', ' 3', ' 2', ' 3', ' 3', ' 3', ' 3', ' 3', ' 3', ' 3', ' 3', ' 4', ' 3', ' 3'] \\
dm-test-b255397ee863bd8faf9def57c0f76769153d905e & M9 & 1.00 & [3, 3, 3] & [' 2', ' 2', ' 2', ' 2', ' 2', ' 2', ' 2', ' 2', ' 2', ' 2', ' 2', ' 2', ' 2', ' 2', ' 2', ' 2', ' 2', ' 2', ' 2', ' 2'] \\

\bottomrule
\end{tabular}
}
    \caption{Examples of differences at GEval vs Human experts for coherence.}
    \label{app:tab:geval_incosistency_Example}
\end{table}

\newpage
\section{Full performance of CheXpert dataset on various model}\label{tab:appendix:fullperformancechexpert}

\begin{figure}[h!]
    \centering

    \includegraphics[width=0.32\linewidth]{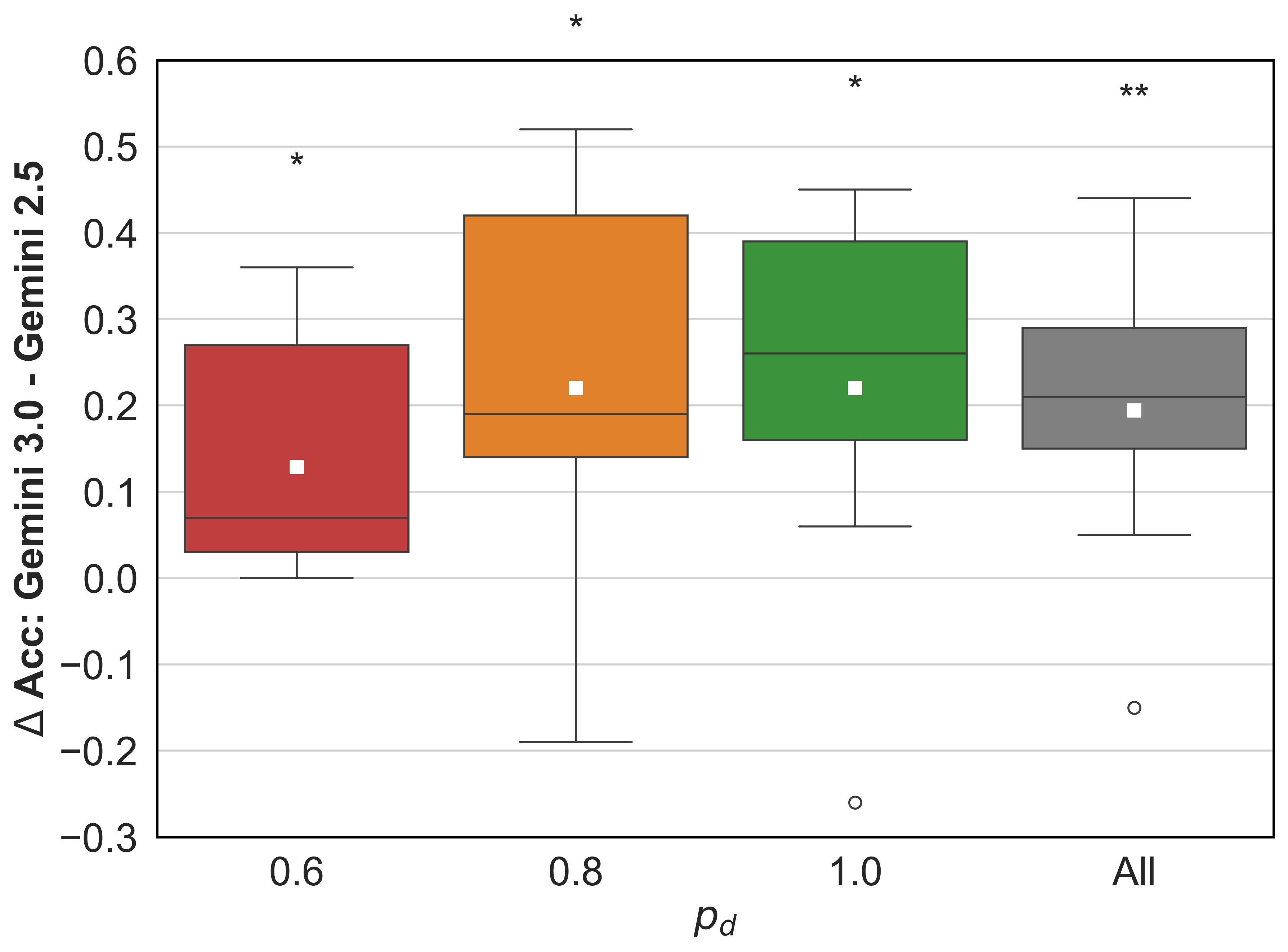}
\includegraphics[width=0.32\linewidth]{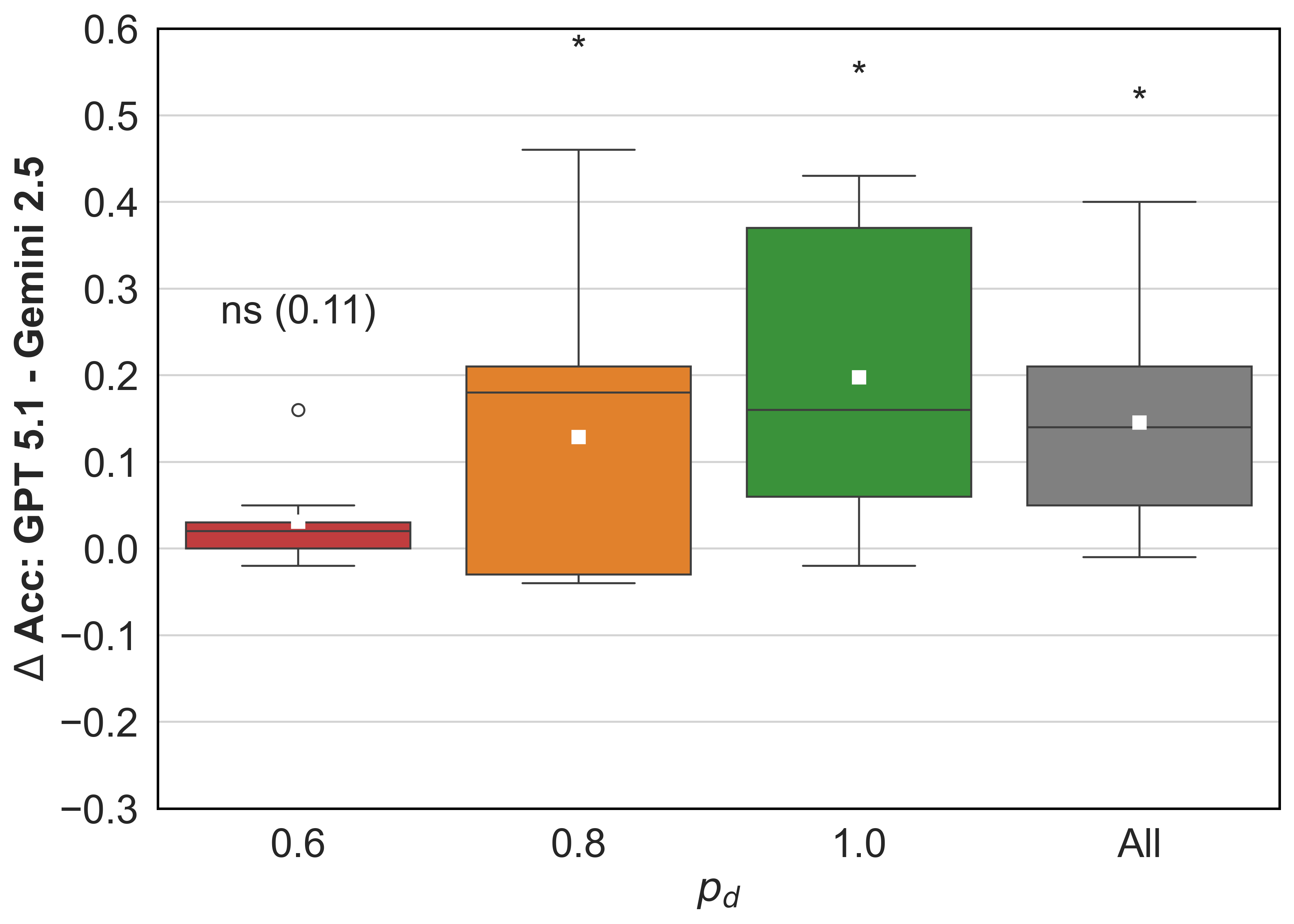}
\includegraphics[width=0.32\linewidth]{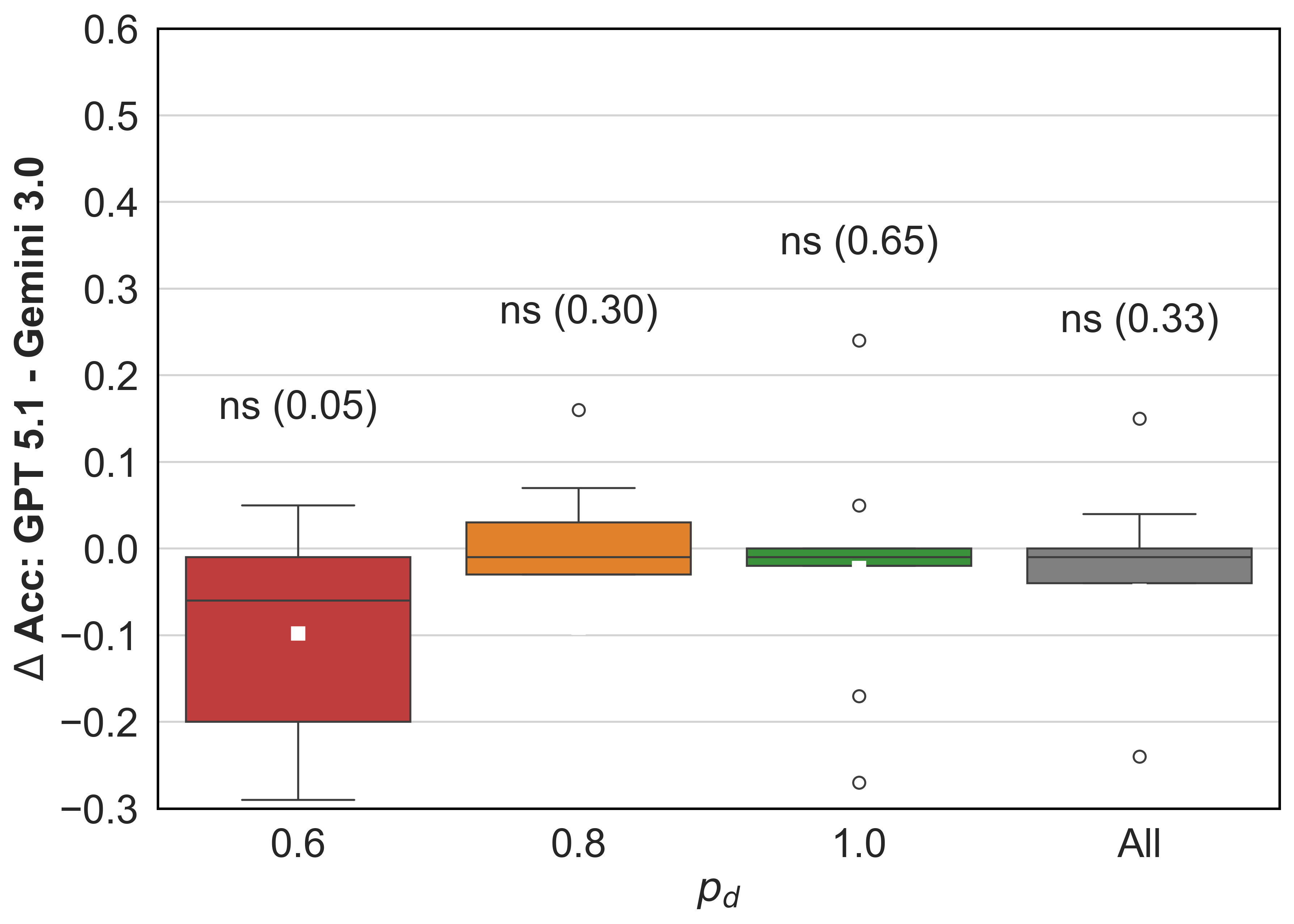}
    \caption{Model-2-Model accuracy comparison: Paired Student's T-test to measure the statistical significance where the null hypothesis ($H_o$)is there is no difference between the pair of models. Significance: * ($p<=0.05$), ns (not significant). \textbf{Left and Mid}: Comparison of strong and weak model:   \textbf{Right}: Comparison of 2 strong models.
    }\label{app:fig:strongvdweakmodelAccuracy}
\end{figure}

\begin{table}[h!]
    \setlength{\tabcolsep}{4pt}
    \centering
    \resizebox{1.0\columnwidth}{!}
    {
    \begin{tabular}{p{0.1\linewidth}rrr|rrrrrr|rrrrrr}
\toprule
  &  &  &  & \multicolumn{6}{c|}{F1} & \multicolumn{6}{c}{Accuracy}\\
 \hline
 Pathology & $p_d$ & S & $m$  & AI & $\text{H}$\ \   & $E$  & $\Delta_{\text{H}-AI}$ & $\Delta_{E-\text{H}}$ & $\Delta_{E-M}$ & AI & $\text{H}$\ \ \  &   $E$  & $\Delta_{\text{H}-AI}$ & $\Delta_{E-\text{H}}$ & $\Delta_{E-M}$\\

\thickhline
% \multicolumn{11}{c}{CXR-Llava Baselines}\\
% \multirow[c]{4}{*}{Atelectasis}& All &  & &  0.69 \\
% \multirow[c]{4}{*}{Cardiomegaly} & All & & &  0.62 \\
% \multirow[c]{4}{*}{Consolidation}  &  All& & &  0.24 \\
% \multirow[c]{4}{*}{Edema}  &  All & & &   0.50 \\
% \multirow[c]{4}{*}{Pleural effusion}  &  All & & &   0.63 \\
% \multirow[c]{4}{*}{Lung opacity}  &  All & & & 0.8\\
% \multirow[c]{4}{*}{Support devices}&  All & & &0.78\\

% \midrule
\midrule
\multirow[c]{4}{*}{Atelectasis} & 0.60 & 130 & 0.500 & 0.57 & 0.66$\pm$0.03 & 0.60 & 0.09 & -0.06 & 0.03 & 0.53 & 0.60$\pm$0.02 & 0.60 & 0.07 & -0.00 & 0.07 \\
\cline{2-16} \cline{3-16}
 & 0.80 & 140 & 0.400 & 0.63 & 0.70$\pm$0.03 & 0.76 & 0.07 & 0.06 & 0.13 & 0.67 & 0.73$\pm$0.03 & 0.80 & 0.06 & 0.07 & 0.13 \\
\cline{2-16} \cline{3-16}
 & 1.00 & 230 & 0.139 & 0.67 & 0.74$\pm$0.03 & 0.90 & 0.07 & 0.16 & 0.23 & \mycolorbox{paleblue}{0.88} & \mycolorbox{paleblue}{0.91$\pm$0.02} & 0.97 & 0.03 & 0.06 & 0.09 \\
\cline{2-16} \cline{3-16}
 & All & 500 & 0.306 & 0.62 & 0.69$\pm$0.02 & 0.73 & 0.07 & 0.04 & 0.11 & 0.73 & 0.78$\pm$0.02 & 0.83 & 0.05 & 0.05 & 0.10 \\
\thickhline
\multirow[c]{4}{*}{\shortstack[l]{Cardio-\\megaly}} & 0.60 & 142 & 0.507 & 0.67 & 0.64$\pm$0.10 & 0.60 & -0.03 & -0.04 & -0.07 & 0.63 & 0.66$\pm$0.03 & 0.60 & 0.03 & -0.06 & -0.03 \\
\cline{2-16} \cline{3-16}
 & 0.80 & 196 & 0.372 & 0.69 & 0.71$\pm$0.04 & 0.75 & 0.02 & 0.04 & 0.06 & 0.72 & 0.79$\pm$0.03 & 0.80 & 0.07 & 0.01 & 0.08 \\
\cline{2-16} \cline{3-16}
 & 1.00 & 162 & 0.037 & 0.35 & 0.70$\pm$0.12 & 0.71 & 0.35 & 0.01 & 0.36 & \mycolorbox{paleblue}{0.91} & \mycolorbox{paleblue}{0.98$\pm$0.02} & 0.97 & 0.07 & -0.01 & 0.06 \\
\cline{2-16} \cline{3-16}
 & All & 500 & 0.302 & 0.66 & 0.68$\pm$0.07 & 0.68 & 0.02 & 0.00 & 0.02 & 0.75 & \mycolorbox{paleblue}{0.81$\pm$0.03} & 0.80 & 0.06 & -0.01 & 0.05 \\
\thickhline
\multirow[c]{4}{*}{\shortstack[l]{Consoli- \\dation}} & 0.60 & 81 & 0.272 & 0.49 & 0.44$\pm$0.07 & 0.45 & -0.05 & 0.01 & -0.04 & 0.64 & 0.66$\pm$0.06 & 0.60 & 0.02 & -0.06 & -0.04 \\
\cline{2-16} \cline{3-16}
 & 0.80 & 89 & 0.067 & 0.36 & 0.36$\pm$0.17 & 0.35 & 0.00 & -0.01 & -0.01 & 0.76 & \mycolorbox{paleblue}{0.82$\pm$0.11} & 0.80 & 0.06 & -0.02 & 0.04 \\
\cline{2-16} \cline{3-16}
 & 1.00 & 330 & 0.003 & 0.00 & 0.23$\pm$0.21 & 0.16 & 0.23 & -0.07 & 0.16 & \mycolorbox{paleblue}{0.92} & \mycolorbox{paleblue}{0.98$\pm$0.01} & 0.97 & 0.06 & -0.01 & 0.05 \\
\cline{2-16} \cline{3-16}
 & All & 500 & 0.058 & 0.34 & 0.39$\pm$0.06 & 0.39 & 0.05 & 0.00 & 0.05 & \mycolorbox{paleblue}{0.85} & \mycolorbox{paleblue}{0.90$\pm$0.03} & 0.88 & 0.05 & -0.02 & 0.03 \\
\thickhline
\multirow[c]{4}{*}{Edema} & 0.60 & 76 & 0.355 & 0.56 & 0.50$\pm$0.03 & 0.52 & -0.06 & 0.02 & -0.04 & 0.75 & 0.60$\pm$0.09 & 0.60 & -0.15 & -0.00 & -0.15 \\
\cline{2-16} \cline{3-16}
 & 0.80 & 116 & 0.233 & 0.44 & 0.57$\pm$0.07 & 0.65 & 0.13 & 0.08 & 0.21 & \mycolorbox{paleblue}{0.80} & 0.76$\pm$0.07 & 0.80 & -0.04 & 0.04 & 0.00 \\
\cline{2-16} \cline{3-16}
 & 1.00 & 308 & 0.078 & 0.58 & 0.70$\pm$0.07 & 0.83 & 0.12 & 0.13 & 0.25 & \mycolorbox{paleblue}{0.95} & \mycolorbox{paleblue}{0.94$\pm$0.02} & 0.97 & -0.01 & 0.03 & 0.02 \\
\cline{2-16} \cline{3-16}
 & All & 500 & 0.156 & 0.52 & 0.58$\pm$0.02 & 0.66 & 0.06 & 0.08 & 0.14 & \mycolorbox{paleblue}{0.88} & \mycolorbox{paleblue}{0.85$\pm$0.03} & 0.87 & -0.03 & 0.02 & -0.01 \\
\thickhline
\multirow[c]{4}{*}{\shortstack[l]{Enlarged \\Cardiomed- \\iastinum}} & 0.60 & 163 & 0.417 & 0.05 & 0.51$\pm$0.06 & 0.56 & 0.46 & 0.05 & 0.51 & 0.56 & 0.66$\pm$0.02 & 0.60 & 0.10 & -0.06 & 0.04 \\
\cline{2-16} \cline{3-16}
 & 0.80 & 170 & 0.529 & 0.28 & 0.75$\pm$0.04 & 0.81 & 0.47 & 0.06 & 0.53 & 0.55 & 0.78$\pm$0.02 & 0.80 & 0.23 & 0.02 & 0.25 \\
\cline{2-16} \cline{3-16}
 & 1.00 & 167 & 0.569 & 0.46 & \mycolorbox{paleblue}{0.92$\pm$0.03} & 0.97 & 0.46 & 0.05 & 0.51 & 0.60 & \mycolorbox{paleblue}{0.91$\pm$0.03} & 0.97 & 0.31 & 0.06 & 0.37 \\
\cline{2-16} \cline{3-16}
 & All & 500 & 0.506 & 0.30 & 0.75$\pm$0.03 & 0.80 & 0.45 & 0.05 & 0.50 & 0.57 & 0.79$\pm$0.02 & 0.79 & 0.22 & 0.00 & 0.22 \\
\thickhline
\multirow[c]{4}{*}{\shortstack[l]{Lung\\Opacity}} & 0.60 & 106 & 0.519 & 0.57 & 0.63$\pm$0.07 & 0.61 & 0.06 & -0.02 & 0.04 & 0.59 & 0.62$\pm$0.03 & 0.60 & 0.03 & -0.02 & 0.01 \\
\cline{2-16} \cline{3-16}
 & 0.80 & 125 & 0.536 & \mycolorbox{paleblue}{0.83} & \mycolorbox{paleblue}{0.84$\pm$0.03} & 0.81 & 0.01 & -0.03 & -0.02 & \mycolorbox{paleblue}{0.82} & \mycolorbox{paleblue}{0.83$\pm$0.03} & 0.80 & 0.01 & -0.03 & -0.02 \\
\cline{2-16} \cline{3-16}
 & 1.00 & 269 & 0.528 & \mycolorbox{paleblue}{0.89} & \mycolorbox{paleblue}{0.96$\pm$0.01} & 0.97 & 0.07 & 0.01 & 0.08 & \mycolorbox{paleblue}{0.89} & \mycolorbox{paleblue}{0.96$\pm$0.01} & 0.97 & 0.07 & 0.01 & 0.08 \\
\cline{2-16} \cline{3-16}
 & All & 500 & 0.528 & \mycolorbox{paleblue}{0.81} & \mycolorbox{paleblue}{0.86$\pm$0.02} & 0.86 & 0.05 & -0.00 & 0.05 & \mycolorbox{paleblue}{0.81} & \mycolorbox{paleblue}{0.86$\pm$0.01} & 0.85 & 0.05 & -0.01 & 0.04 \\
\thickhline
\multirow[c]{4}{*}{\shortstack[l]{Pleural\\Effusion}} & 0.60 & 83 & 0.422 & 0.52 & 0.61$\pm$0.08 & 0.56 & 0.09 & -0.05 & 0.04 & 0.51 & 0.62$\pm$0.11 & 0.60 & 0.11 & -0.02 & 0.09 \\
\cline{2-16} \cline{3-16}
 & 0.80 & 110 & 0.336 & 0.56 & 0.78$\pm$0.07 & 0.73 & 0.22 & -0.05 & 0.17 & 0.62 & \mycolorbox{paleblue}{0.82$\pm$0.07} & 0.80 & 0.20 & -0.02 & 0.18 \\
\cline{2-16} \cline{3-16}
 & 1.00 & 307 & 0.104 & 0.57 & \mycolorbox{paleblue}{0.83$\pm$0.08} & 0.87 & 0.26 & 0.04 & 0.30 & \mycolorbox{paleblue}{0.87} & \mycolorbox{paleblue}{0.96$\pm$0.02} & 0.97 & 0.09 & 0.01 & 0.10 \\
\cline{2-16} \cline{3-16}
 & All & 500 & 0.208 & 0.55 & 0.74$\pm$0.08 & 0.72 & 0.19 & -0.02 & 0.17 & 0.75 & \mycolorbox{paleblue}{0.87$\pm$0.05} & 0.87 & 0.12 & -0.00 & 0.12 \\
\thickhline
\multirow[c]{4}{*}{Pneumonia} & 0.60 & 24 & 0.333 & 0.57 & 0.22$\pm$0.08 & 0.50 & -0.35 & 0.28 & -0.07 & 0.62 & 0.60$\pm$0.11 & 0.60 & -0.02 & 0.00 & -0.02 \\
\cline{2-16} \cline{3-16}
 & 0.80 & 82 & 0.012 & 0.12 & 0.30$\pm$0.17 & 0.09 & 0.18 & -0.21 & -0.03 & \mycolorbox{paleblue}{0.82} & \mycolorbox{paleblue}{0.93$\pm$0.05} & 0.80 & 0.11 & -0.13 & -0.02 \\
\cline{2-16} \cline{3-16}
 & 1.00 & 394 & 0.005 & 0.08 & 0.34$\pm$0.29 & 0.25 & 0.26 & -0.09 & 0.17 & \mycolorbox{paleblue}{0.94} & \mycolorbox{paleblue}{0.98$\pm$0.01} & 0.97 & 0.04 & -0.01 & 0.03 \\
\cline{2-16} \cline{3-16}
 & All & 500 & 0.022 & 0.25 & 0.27$\pm$0.13 & 0.29 & 0.02 & 0.02 & 0.04 & \mycolorbox{paleblue}{0.90} & \mycolorbox{paleblue}{0.95$\pm$0.02} & 0.92 & 0.05 & -0.03 & 0.02 \\
\thickhline
\multirow[c]{4}{*}{\shortstack[l]{Support\\Devices}} & 0.60 & 50 & 0.580 & 0.75 & 0.72$\pm$0.04 & 0.64 & -0.03 & -0.08 & -0.11 & 0.62 & 0.67$\pm$0.05 & 0.60 & 0.05 & -0.07 & -0.02 \\
\cline{2-16} \cline{3-16}
 & 0.80 & 105 & 0.705 & \mycolorbox{paleblue}{0.86} & \mycolorbox{paleblue}{0.93$\pm$0.02} & 0.85 & 0.07 & -0.08 & -0.01 & 0.77 & \mycolorbox{paleblue}{0.90$\pm$0.02} & 0.80 & 0.13 & -0.10 & 0.03 \\
\cline{2-16} \cline{3-16}
 & 1.00 & 345 & 0.458 & 0.76 & \mycolorbox{paleblue}{0.98$\pm$0.00} & 0.97 & 0.22 & -0.01 & 0.21 & 0.72 & \mycolorbox{paleblue}{0.98$\pm$0.00} & 0.97 & 0.26 & -0.01 & 0.25 \\
\cline{2-16} \cline{3-16}
 & All & 500 & 0.522 & 0.78 & \mycolorbox{paleblue}{0.93$\pm$0.01} & 0.90 & 0.15 & -0.03 & 0.12 & 0.72 & \mycolorbox{paleblue}{0.93$\pm$0.01} & 0.90 & 0.21 & -0.03 & 0.18 \\
\thickhline
\end{tabular}
   }
  \caption{Performance of Gemini Pro 3.0 Preview (\textbf{GP}) model vs.\ human (\textbf{H}) radiologists,  compared against ground truth (majority label across 5 human radiologists). The data is stratified by $p_d$ probability of observed agreement among the 5 radiologists for the majority label. Size (S) is the number of samples  corresponding to that $p_d$, positive class ratio is $m$. \textbf{H} is the average performance of additional 3 radiologists and the corresponding standard deviation.  Humans  achieve a peak F1-score of over \textul{\textbf{0.98}} @ $p_d=1.0$, while the best performance at $p_d=0.6$ is \textbf{0.72} across all pathologies. Similar trend is seen for recall, precision and accuracy. At low positive class ratio ($m<0.01$), even at $p_d=1.0$  (see Pneumonia and Consolidation), F1 is $< 0.35$. $\Delta = H - GP$ is the difference between the average of human performance and the model. All the scores (Model or Human) over \mycolorbox{paleblue}{0.80} are highlighted and they all occur at $p_d \in {0.8, 1.0}$ but never at $0.60$ given the sample size ($S$).}
    \label{tab:appendix:gemini_profull}
\end{table}

\clearpage

\begin{table}[h!]
    \setlength{\tabcolsep}{4pt}
    \centering
    \resizebox{1.0\columnwidth}{!}
    {
    \begin{tabular}{p{0.1\linewidth}rrr|rrrrrr|rrrrrr}
\toprule
  &  &  &  & \multicolumn{6}{c|}{F1} & \multicolumn{6}{c}{Accuracy}\\
 \hline
 Pathology & $p_d$ & S & $m$  & AI & $\text{H}$\ \   & $E$  & $\Delta_{\text{H}-AI}$ & $\Delta_{E-\text{H}}$ & $\Delta_{E-M}$ & AI & $\text{H}$\ \ \  &   $E$  & $\Delta_{\text{H}-AI}$ & $\Delta_{E-\text{H}}$ & $\Delta_{E-M}$\\
\midrule
\multirow[c]{4}{*}{Atelectasis} & 0.60 & 130 & 0.500 & 0.61 & 0.66$\pm$0.03 & 0.60 & 0.05 & -0.06 & -0.01 & 0.52 & 0.60$\pm$0.02 & 0.60 & 0.08 & -0.00 & 0.08 \\
\cline{2-16} \cline{3-16}
 & 0.80 & 140 & 0.400 & 0.63 & 0.70$\pm$0.03 & 0.76 & 0.07 & 0.06 & 0.13 & 0.63 & 0.73$\pm$0.03 & 0.80 & 0.10 & 0.07 & 0.17 \\
\cline{2-16} \cline{3-16}
 & 1.00 & 230 & 0.139 & 0.58 & 0.74$\pm$0.03 & 0.90 & 0.16 & 0.16 & 0.32 & \mycolorbox{paleblue}{0.82} & \mycolorbox{paleblue}{0.91$\pm$0.02} & 0.97 & 0.09 & 0.06 & 0.15 \\
\cline{2-16} \cline{3-16}
 & All & 500 & 0.306 & 0.61 & 0.69$\pm$0.02 & 0.73 & 0.08 & 0.04 & 0.12 & 0.69 & 0.78$\pm$0.02 & 0.83 & 0.09 & 0.05 & 0.14 \\
\thickhline
\multirow[c]{4}{*}{\shortstack[l]{Cardio-\\megaly}} & 0.60 & 142 & 0.507 & 0.43 & 0.64$\pm$0.10 & 0.60 & 0.21 & -0.04 & 0.17 & 0.62 & 0.66$\pm$0.03 & 0.60 & 0.04 & -0.06 & -0.02 \\
\cline{2-16} \cline{3-16}
 & 0.80 & 196 & 0.372 & 0.39 & 0.71$\pm$0.04 & 0.75 & 0.32 & 0.04 & 0.36 & 0.71 & 0.79$\pm$0.03 & 0.80 & 0.08 & 0.01 & 0.09 \\
\cline{2-16} \cline{3-16}
 & 1.00 & 162 & 0.037 & 0.00 & 0.70$\pm$0.12 & 0.71 & 0.70 & 0.01 & 0.71 & \mycolorbox{paleblue}{0.96} & \mycolorbox{paleblue}{0.98$\pm$0.02} & 0.97 & 0.02 & -0.01 & 0.01 \\
\cline{2-16} \cline{3-16}
 & All & 500 & 0.302 & 0.39 & 0.68$\pm$0.07 & 0.68 & 0.29 & 0.00 & 0.29 & 0.77 & \mycolorbox{paleblue}{0.81$\pm$0.03} & 0.80 & 0.04 & -0.01 & 0.03 \\
\thickhline
\multirow[c]{4}{*}{\shortstack[l]{Consoli- \\dation}} & 0.60 & 81 & 0.272 & 0.30 & 0.44$\pm$0.07 & 0.45 & 0.14 & 0.01 & 0.15 & 0.65 & 0.66$\pm$0.06 & 0.60 & 0.01 & -0.06 & -0.05 \\
\cline{2-16} \cline{3-16}
 & 0.80 & 89 & 0.067 & 0.35 & 0.36$\pm$0.17 & 0.35 & 0.01 & -0.01 & 0.00 & \mycolorbox{paleblue}{0.83} & \mycolorbox{paleblue}{0.82$\pm$0.11} & 0.80 & -0.01 & -0.02 & -0.03 \\
\cline{2-16} \cline{3-16}
 & 1.00 & 330 & 0.003 & 0.00 & 0.23$\pm$0.21 & 0.16 & 0.23 & -0.07 & 0.16 & \mycolorbox{paleblue}{0.98} & \mycolorbox{paleblue}{0.98$\pm$0.01} & 0.97 & 0.00 & -0.01 & -0.01 \\
\cline{2-16} \cline{3-16}
 & All & 500 & 0.058 & 0.29 & 0.39$\pm$0.06 & 0.39 & 0.10 & 0.00 & 0.10 & \mycolorbox{paleblue}{0.90} & \mycolorbox{paleblue}{0.90$\pm$0.03} & 0.88 & 0.00 & -0.02 & -0.02 \\
\thickhline
\multirow[c]{4}{*}{Edema} & 0.60 & 76 & 0.355 & 0.49 & 0.50$\pm$0.03 & 0.52 & 0.01 & 0.02 & 0.03 & 0.62 & 0.60$\pm$0.09 & 0.60 & -0.02 & -0.00 & -0.02 \\
\cline{2-16} \cline{3-16}
 & 0.80 & 116 & 0.233 & 0.40 & 0.57$\pm$0.07 & 0.65 & 0.17 & 0.08 & 0.25 & 0.69 & 0.76$\pm$0.07 & 0.80 & 0.07 & 0.04 & 0.11 \\
\cline{2-16} \cline{3-16}
 & 1.00 & 308 & 0.078 & 0.58 & 0.70$\pm$0.07 & 0.83 & 0.12 & 0.13 & 0.25 & \mycolorbox{paleblue}{0.93} & \mycolorbox{paleblue}{0.94$\pm$0.02} & 0.97 & 0.01 & 0.03 & 0.04 \\
\cline{2-16} \cline{3-16}
 & All & 500 & 0.156 & 0.49 & 0.58$\pm$0.02 & 0.66 & 0.09 & 0.08 & 0.17 & \mycolorbox{paleblue}{0.83} & \mycolorbox{paleblue}{0.85$\pm$0.03} & 0.87 & 0.02 & 0.02 & 0.04 \\
\thickhline
\multirow[c]{4}{*}{\shortstack[l]{Enlarged \\Cardiomed- \\iastinum}} & 0.60 & 163 & 0.417 & 0.08 & 0.51$\pm$0.06 & 0.56 & 0.43 & 0.05 & 0.48 & 0.59 & 0.66$\pm$0.02 & 0.60 & 0.07 & -0.06 & 0.01 \\
\cline{2-16} \cline{3-16}
 & 0.80 & 170 & 0.529 & 0.22 & 0.75$\pm$0.04 & 0.81 & 0.53 & 0.06 & 0.59 & 0.53 & 0.78$\pm$0.02 & 0.80 & 0.25 & 0.02 & 0.27 \\
\cline{2-16} \cline{3-16}
 & 1.00 & 167 & 0.569 & 0.29 & \mycolorbox{paleblue}{0.92$\pm$0.03} & 0.97 & 0.63 & 0.05 & 0.68 & 0.53 & \mycolorbox{paleblue}{0.91$\pm$0.03} & 0.97 & 0.38 & 0.06 & 0.44 \\
\cline{2-16} \cline{3-16}
 & All & 500 & 0.506 & 0.21 & 0.75$\pm$0.03 & 0.80 & 0.54 & 0.05 & 0.59 & 0.55 & 0.79$\pm$0.02 & 0.79 & 0.24 & 0.00 & 0.24 \\
\thickhline
\multirow[c]{4}{*}{\shortstack[l]{Lung\\Opacity}} & 0.60 & 106 & 0.519 & 0.57 & 0.63$\pm$0.07 & 0.61 & 0.06 & -0.02 & 0.04 & 0.57 & 0.62$\pm$0.03 & 0.60 & 0.05 & -0.02 & 0.03 \\
\cline{2-16} \cline{3-16}
 & 0.80 & 125 & 0.536 & \mycolorbox{paleblue}{0.82} & \mycolorbox{paleblue}{0.84$\pm$0.03} & 0.81 & 0.02 & -0.03 & -0.01 & \mycolorbox{paleblue}{0.82} & \mycolorbox{paleblue}{0.83$\pm$0.03} & 0.80 & 0.01 & -0.03 & -0.02 \\
\cline{2-16} \cline{3-16}
 & 1.00 & 269 & 0.528 & \mycolorbox{paleblue}{0.88} & \mycolorbox{paleblue}{0.96$\pm$0.01} & 0.97 & 0.08 & 0.01 & 0.09 & \mycolorbox{paleblue}{0.88} & \mycolorbox{paleblue}{0.96$\pm$0.01} & 0.97 & 0.08 & 0.01 & 0.09 \\
\cline{2-16} \cline{3-16}
 & All & 500 & 0.528 & \mycolorbox{paleblue}{0.80} & \mycolorbox{paleblue}{0.86$\pm$0.02} & 0.86 & 0.06 & -0.00 & 0.06 & \mycolorbox{paleblue}{0.80} & \mycolorbox{paleblue}{0.86$\pm$0.01} & 0.85 & 0.06 & -0.01 & 0.05 \\
\thickhline
\multirow[c]{4}{*}{\shortstack[l]{Pleural\\Effusion}} & 0.60 & 83 & 0.422 & 0.55 & 0.61$\pm$0.08 & 0.56 & 0.06 & -0.05 & 0.01 & 0.43 & 0.62$\pm$0.11 & 0.60 & 0.19 & -0.02 & 0.17 \\
\cline{2-16} \cline{3-16}
 & 0.80 & 110 & 0.336 & 0.57 & 0.78$\pm$0.07 & 0.73 & 0.21 & -0.05 & 0.16 & 0.59 & \mycolorbox{paleblue}{0.82$\pm$0.07} & 0.80 & 0.23 & -0.02 & 0.21 \\
\cline{2-16} \cline{3-16}
 & 1.00 & 307 & 0.104 & 0.53 & \mycolorbox{paleblue}{0.83$\pm$0.08} & 0.87 & 0.30 & 0.04 & 0.34 & \mycolorbox{paleblue}{0.82} & \mycolorbox{paleblue}{0.96$\pm$0.02} & 0.97 & 0.14 & 0.01 & 0.15 \\
\cline{2-16} \cline{3-16}
 & All & 500 & 0.208 & 0.55 & 0.74$\pm$0.08 & 0.72 & 0.19 & -0.02 & 0.17 & 0.70 & \mycolorbox{paleblue}{0.87$\pm$0.05} & 0.87 & 0.17 & -0.00 & 0.17 \\
\thickhline
\multirow[c]{4}{*}{Pneumonia} & 0.60 & 24 & 0.333 & 0.36 & 0.22$\pm$0.08 & 0.50 & -0.14 & 0.28 & 0.14 & 0.71 & 0.60$\pm$0.11 & 0.60 & -0.11 & 0.00 & -0.11 \\
\cline{2-16} \cline{3-16}
 & 0.80 & 82 & 0.012 & 0.00 & 0.30$\pm$0.17 & 0.09 & 0.30 & -0.21 & 0.09 & \mycolorbox{paleblue}{0.88} & \mycolorbox{paleblue}{0.93$\pm$0.05} & 0.80 & 0.05 & -0.13 & -0.08 \\
\cline{2-16} \cline{3-16}
 & 1.00 & 394 & 0.005 & 0.12 & 0.34$\pm$0.29 & 0.25 & 0.22 & -0.09 & 0.13 & \mycolorbox{paleblue}{0.96} & \mycolorbox{paleblue}{0.98$\pm$0.01} & 0.97 & 0.02 & -0.01 & 0.01 \\
\cline{2-16} \cline{3-16}
 & All & 500 & 0.022 & 0.16 & 0.27$\pm$0.13 & 0.29 & 0.11 & 0.02 & 0.13 & \mycolorbox{paleblue}{0.94} & \mycolorbox{paleblue}{0.95$\pm$0.02} & 0.92 & 0.01 & -0.03 & -0.02 \\
\thickhline
\multirow[c]{4}{*}{\shortstack[l]{Support\\Devices}} & 0.60 & 50 & 0.580 & 0.74 & 0.72$\pm$0.04 & 0.64 & -0.02 & -0.08 & -0.10 & 0.60 & 0.67$\pm$0.05 & 0.60 & 0.07 & -0.07 & 0.00 \\
\cline{2-16} \cline{3-16}
 & 0.80 & 105 & 0.705 & \mycolorbox{paleblue}{0.88} & \mycolorbox{paleblue}{0.93$\pm$0.02} & 0.85 & 0.05 & -0.08 & -0.03 & \mycolorbox{paleblue}{0.82} & \mycolorbox{paleblue}{0.90$\pm$0.02} & 0.80 & 0.08 & -0.10 & -0.02 \\
\cline{2-16} \cline{3-16}
 & 1.00 & 345 & 0.458 & \mycolorbox{paleblue}{0.80} & \mycolorbox{paleblue}{0.98$\pm$0.00} & 0.97 & 0.18 & -0.01 & 0.17 & 0.79 & \mycolorbox{paleblue}{0.98$\pm$0.00} & 0.97 & 0.19 & -0.01 & 0.18 \\
\cline{2-16} \cline{3-16}
 & All & 500 & 0.522 & \mycolorbox{paleblue}{0.82} & \mycolorbox{paleblue}{0.93$\pm$0.01} & 0.90 & 0.11 & -0.03 & 0.08 & 0.78 & \mycolorbox{paleblue}{0.93$\pm$0.01} & 0.90 & 0.15 & -0.03 & 0.12 \\

\thickhline
\end{tabular}
}
\caption{Performance on CheXpert using Gemini-3.1 pro}
    \label{tab:appendix:gemini31pro}
\end{table}

\begin{figure}[h!]
        \centering

          \includegraphics[width=0.48\linewidth]{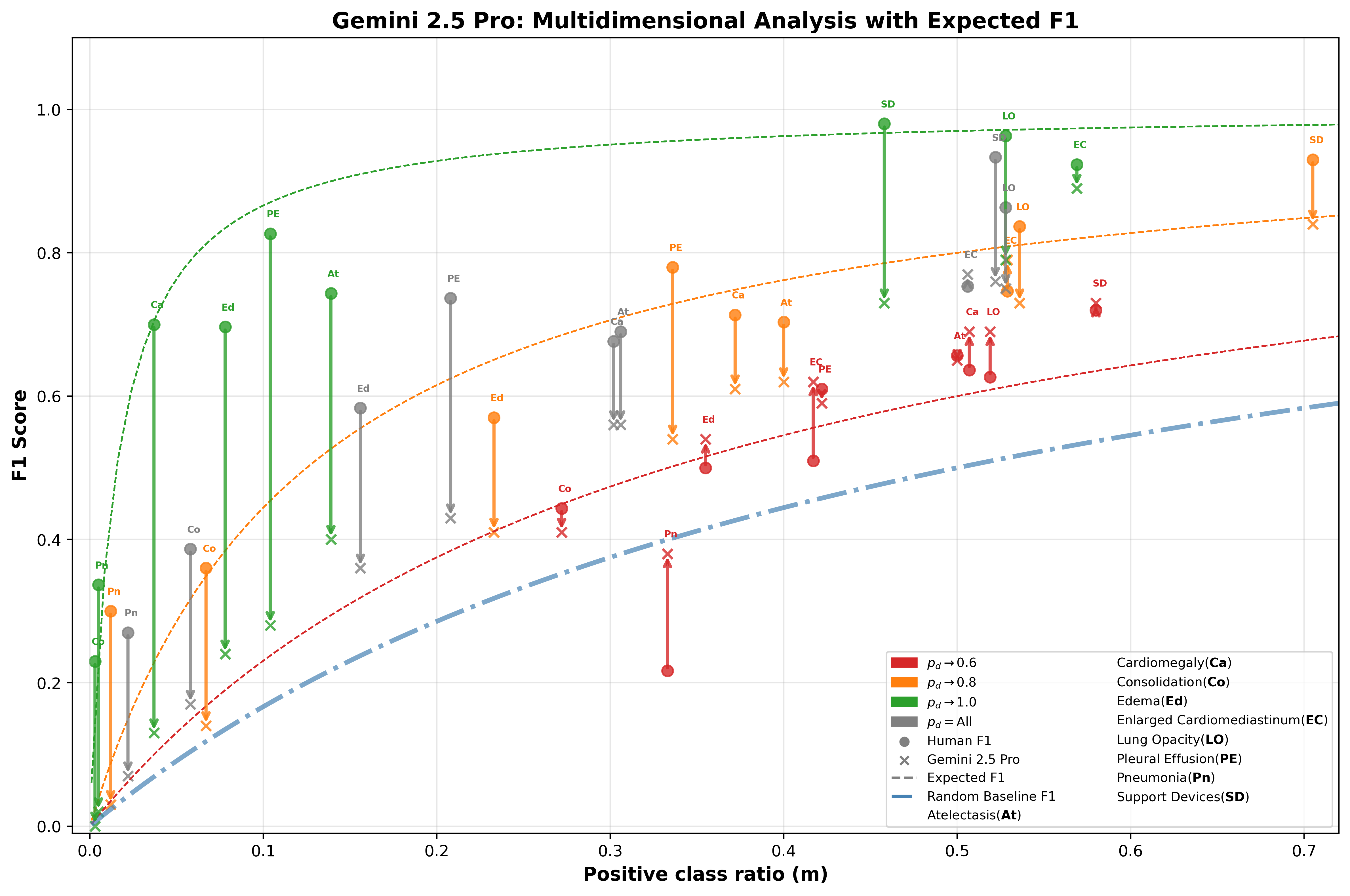}
         \includegraphics[width=0.48\linewidth]{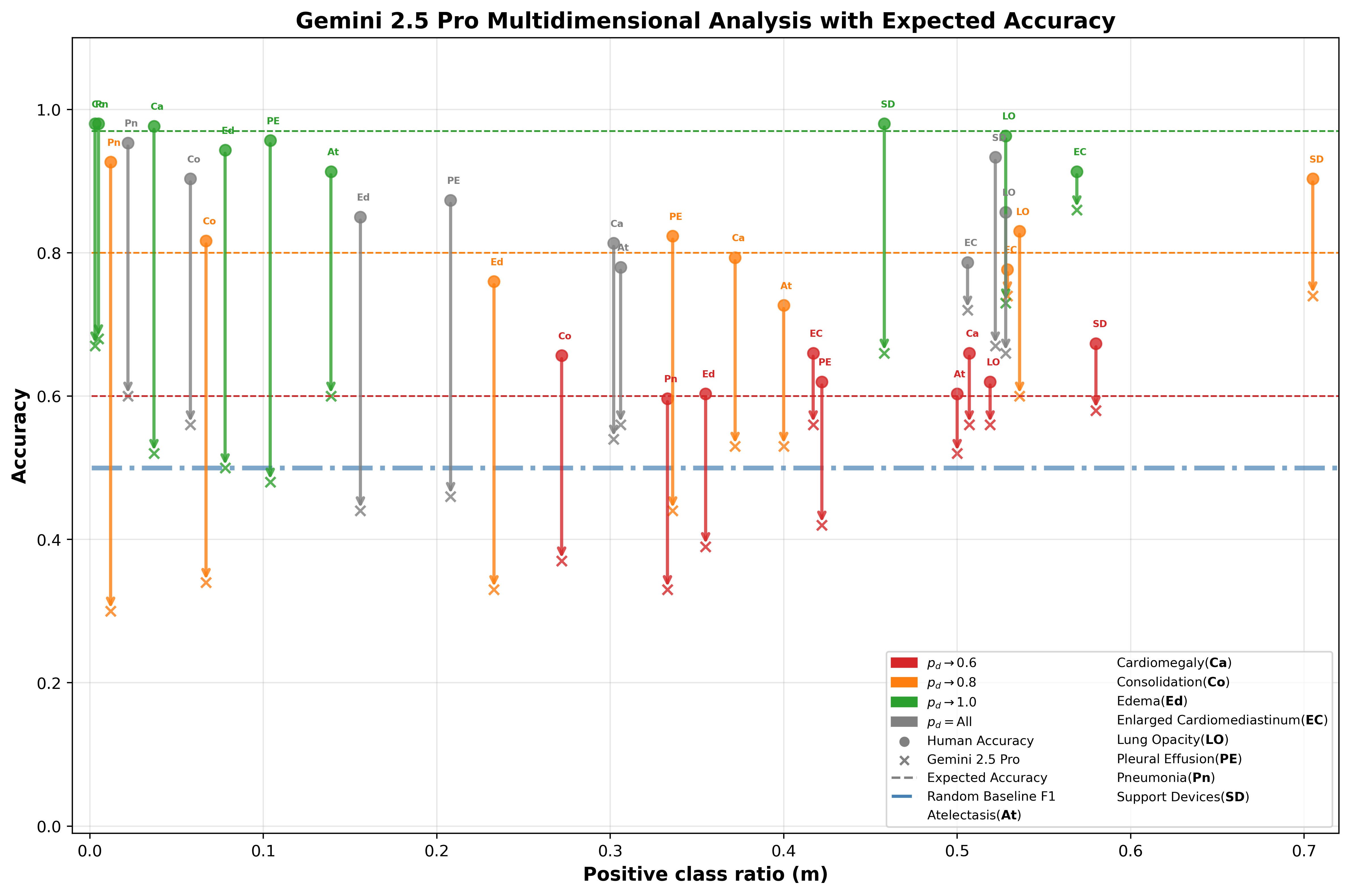}\\
           \includegraphics[width=0.48\linewidth]{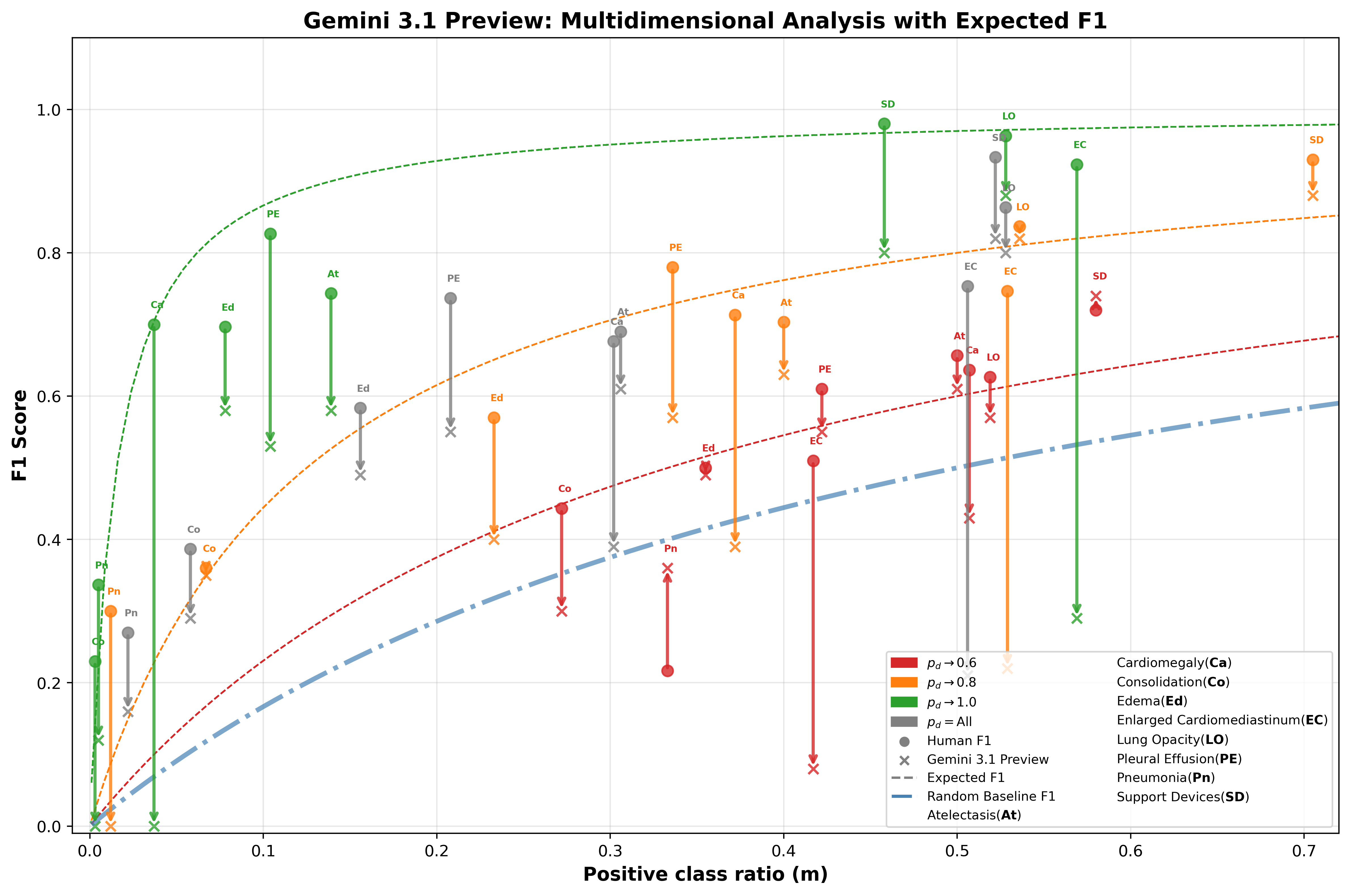}
         \includegraphics[width=0.48\linewidth]{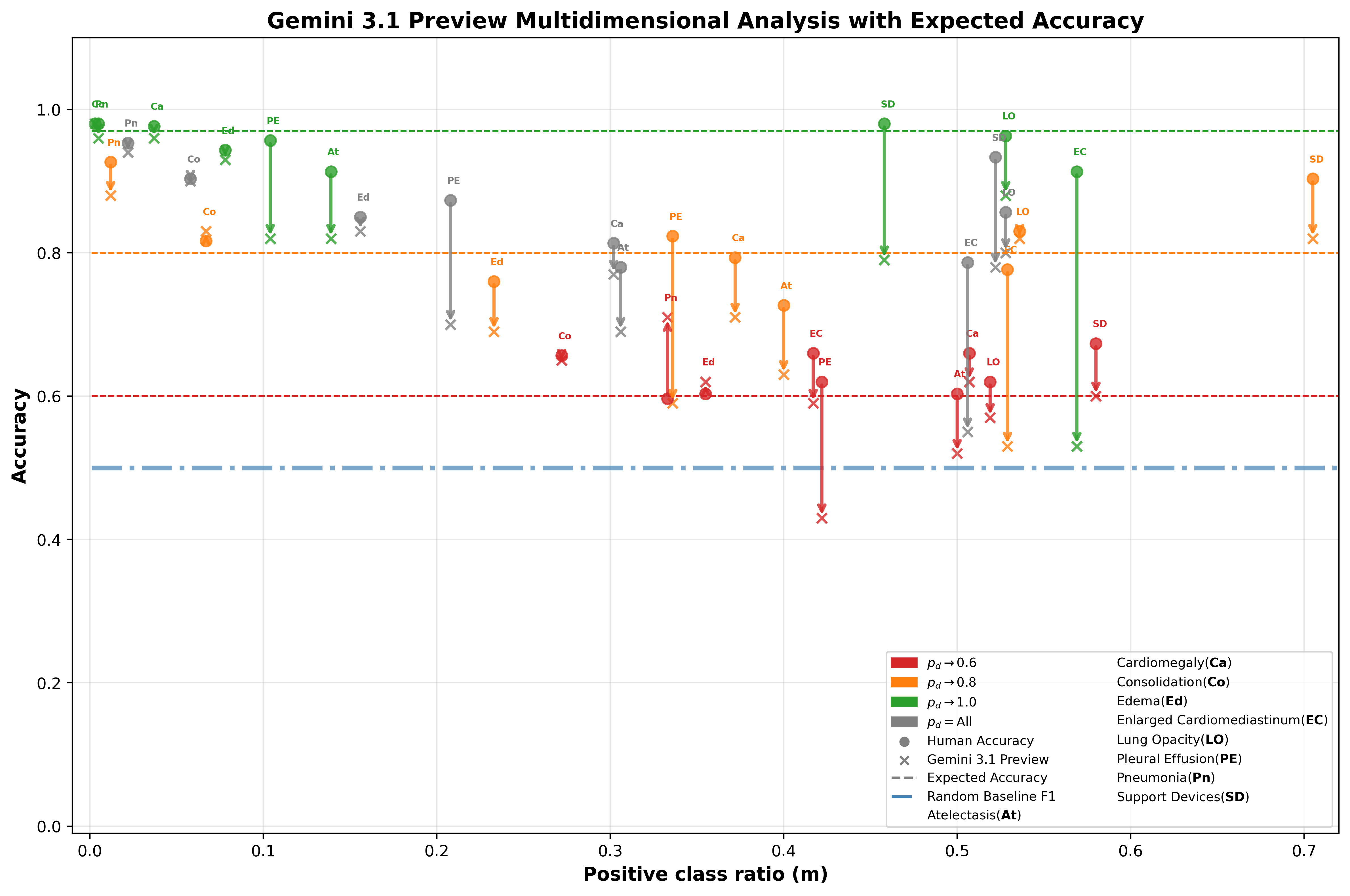}\\

        \caption{Performance of Gemini-2.5, Gemini-3.1 (\textbf{M}) vs.\ human (\textbf{H}) radiologists,  compared against ground truth (majority label across 5 human radiologists). \textbf{H} is the average performance of additional 3 radiologists who are different from the 5 radiologists used to determine the ground truth. The data is stratified by $p_d$ probability of observed agreement among the 5 radiologists for the majority label, corresponding positive class ratio is $m$. Dashed lines are theoretical expected performance at given $p_d$ and $m$  according to equations~\ref{equ:expectedacc} and \ref{equ:equationf1}, ($p_d\rightarrow1.0$ is approximated as $p_d=0.97$).  Humans  are  able to achieve a relatively high F1-score ($> 0.8$) as  $p_d\rightarrow1.0$, whereas  $p_d\rightarrow0.6$ the peak  performance drops close to a random labeller baseline. At low positive class ratio ($m<0.01$), even at $p_d\rightarrow1.0$  (see Pneumonia and Consolidation), F1 is $< 0.35$. $\Delta = H - M$ is the vertical distance  illustrated by the length of the vertical line connecting the human and the model performance.   }
        \label{fig:gemini2.5vshuman}
\end{figure}

\begin{table}[h!]
    \setlength{\tabcolsep}{4pt}
    \centering
    \resizebox{1.0\columnwidth}{!}
    {
\begin{tabular}{p{0.1\linewidth}rrr|rrrrrr|rrrrrr}
\toprule
  &  &  &  & \multicolumn{6}{c|}{F1} & \multicolumn{6}{c}{Accuracy}\\
 \hline
 Pathology & $p_d$ & S & $m$  & AI & $\text{H}'$\ \   & $E$  & $\Delta_{\text{H}-AI}$ & $\Delta_{E-\text{H}}$ & $\Delta_{E-M}$ & AI & $\text{H}$\ \ \  &   $E$  & $\Delta_{\text{H}-AI}$ & $\Delta_{E-\text{H}}$ & $\Delta_{E-M}$\\

\thickhline
\multirow[c]{4}{*}{Atelectasis} & 0.60 & 130 & 0.500 & 0.65 & 0.66$\pm$0.03 & 0.60 & 0.01 & -0.06 & -0.05 & 0.52 & 0.60$\pm$0.02 & 0.60 & 0.08 & -0.00 & 0.08 \\
\cline{2-16} \cline{3-16}
 & 0.80 & 140 & 0.400 & 0.62 & 0.70$\pm$0.03 & 0.76 & 0.08 & 0.06 & 0.14 & 0.53 & 0.73$\pm$0.03 & 0.80 & 0.20 & 0.07 & 0.27 \\
\cline{2-16} \cline{3-16}
 & 1.00 & 230 & 0.139 & 0.40 & 0.74$\pm$0.03 & 0.90 & 0.34 & 0.16 & 0.50 & 0.60 & \mycolorbox{paleblue}{0.91$\pm$0.02} & 0.97 & 0.31 & 0.06 & 0.37 \\
\cline{2-16} \cline{3-16}
 & All & 500 & 0.306 & 0.56 & 0.69$\pm$0.02 & 0.73 & 0.13 & 0.04 & 0.17 & 0.56 & 0.78$\pm$0.02 & 0.83 & 0.22 & 0.05 & 0.27 \\
\thickhline
\multirow[c]{4}{*}{\shortstack[l]{Cardio-\\megaly}} & 0.60 & 142 & 0.507 & 0.69 & 0.64$\pm$0.10 & 0.60 & -0.05 & -0.04 & -0.09 & 0.56 & 0.66$\pm$0.03 & 0.60 & 0.10 & -0.06 & 0.04 \\
\cline{2-16} \cline{3-16}
 & 0.80 & 196 & 0.372 & 0.61 & 0.71$\pm$0.04 & 0.75 & 0.10 & 0.04 & 0.14 & 0.53 & 0.79$\pm$0.03 & 0.80 & 0.26 & 0.01 & 0.27 \\
\cline{2-16} \cline{3-16}
 & 1.00 & 162 & 0.037 & 0.13 & 0.70$\pm$0.12 & 0.71 & 0.57 & 0.01 & 0.58 & 0.52 & \mycolorbox{paleblue}{0.98$\pm$0.02} & 0.97 & 0.46 & -0.01 & 0.45 \\
\cline{2-16} \cline{3-16}
 & All & 500 & 0.302 & 0.56 & 0.68$\pm$0.07 & 0.68 & 0.12 & 0.00 & 0.12 & 0.54 & \mycolorbox{paleblue}{0.81$\pm$0.03} & 0.80 & 0.27 & -0.01 & 0.26 \\
\thickhline
\multirow[c]{4}{*}{\shortstack[l]{Consoli- \\dation}} & 0.60 & 81 & 0.272 & 0.41 & 0.44$\pm$0.07 & 0.45 & 0.03 & 0.01 & 0.04 & 0.37 & 0.66$\pm$0.06 & 0.60 & 0.29 & -0.06 & 0.23 \\
\cline{2-16} \cline{3-16}
 & 0.80 & 89 & 0.067 & 0.14 & 0.36$\pm$0.17 & 0.35 & 0.22 & -0.01 & 0.21 & 0.34 & \mycolorbox{paleblue}{0.82$\pm$0.11} & 0.80 & 0.48 & -0.02 & 0.46 \\
\cline{2-16} \cline{3-16}
 & 1.00 & 330 & 0.003 & 0.00 & 0.23$\pm$0.21 & 0.16 & 0.23 & -0.07 & 0.16 & 0.67 & \mycolorbox{paleblue}{0.98$\pm$0.01} & 0.97 & 0.31 & -0.01 & 0.30 \\
\cline{2-16} \cline{3-16}
 & All & 500 & 0.058 & 0.17 & 0.39$\pm$0.06 & 0.39 & 0.22 & 0.00 & 0.22 & 0.56 & \mycolorbox{paleblue}{0.90$\pm$0.03} & 0.88 & 0.34 & -0.02 & 0.32 \\
\thickhline
\multirow[c]{4}{*}{Edema} & 0.60 & 76 & 0.355 & 0.54 & 0.50$\pm$0.03 & 0.52 & -0.04 & 0.02 & -0.02 & 0.39 & 0.60$\pm$0.09 & 0.60 & 0.21 & -0.00 & 0.21 \\
\cline{2-16} \cline{3-16}
 & 0.80 & 116 & 0.233 & 0.41 & 0.57$\pm$0.07 & 0.65 & 0.16 & 0.08 & 0.24 & 0.33 & 0.76$\pm$0.07 & 0.80 & 0.43 & 0.04 & 0.47 \\
\cline{2-16} \cline{3-16}
 & 1.00 & 308 & 0.078 & 0.24 & 0.70$\pm$0.07 & 0.83 & 0.46 & 0.13 & 0.59 & 0.50 & \mycolorbox{paleblue}{0.94$\pm$0.02} & 0.97 & 0.44 & 0.03 & 0.47 \\
\cline{2-16} \cline{3-16}
 & All & 500 & 0.156 & 0.36 & 0.58$\pm$0.02 & 0.66 & 0.22 & 0.08 & 0.30 & 0.44 & \mycolorbox{paleblue}{0.85$\pm$0.03} & 0.87 & 0.41 & 0.02 & 0.43 \\
\thickhline
\multirow[c]{4}{*}{\shortstack[l]{Enlarged \\Cardiomed- \\iastinum}} & 0.60 & 163 & 0.417 & 0.62 & 0.51$\pm$0.06 & 0.56 & -0.11 & 0.05 & -0.06 & 0.56 & 0.66$\pm$0.02 & 0.60 & 0.10 & -0.06 & 0.04 \\
\cline{2-16} \cline{3-16}
 & 0.80 & 170 & 0.529 & 0.79 & 0.75$\pm$0.04 & 0.81 & -0.04 & 0.06 & 0.02 & 0.74 & 0.78$\pm$0.02 & 0.80 & 0.04 & 0.02 & 0.06 \\
\cline{2-16} \cline{3-16}
 & 1.00 & 167 & 0.569 & \mycolorbox{paleblue}{0.89} & \mycolorbox{paleblue}{0.92$\pm$0.03} & 0.97 & 0.03 & 0.05 & 0.08 & \mycolorbox{paleblue}{0.86} & \mycolorbox{paleblue}{0.91$\pm$0.03} & 0.97 & 0.05 & 0.06 & 0.11 \\
\cline{2-16} \cline{3-16}
 & All & 500 & 0.506 & 0.77 & 0.75$\pm$0.03 & 0.80 & -0.02 & 0.05 & 0.03 & 0.72 & 0.79$\pm$0.02 & 0.79 & 0.07 & 0.00 & 0.07 \\
\thickhline
\multirow[c]{4}{*}{\shortstack[l]{Lung\\Opacity}} & 0.60 & 106 & 0.519 & 0.69 & 0.63$\pm$0.07 & 0.61 & -0.06 & -0.02 & -0.08 & 0.56 & 0.62$\pm$0.03 & 0.60 & 0.06 & -0.02 & 0.04 \\
\cline{2-16} \cline{3-16}
 & 0.80 & 125 & 0.536 & 0.73 & \mycolorbox{paleblue}{0.84$\pm$0.03} & 0.81 & 0.11 & -0.03 & 0.08 & 0.60 & \mycolorbox{paleblue}{0.83$\pm$0.03} & 0.80 & 0.23 & -0.03 & 0.20 \\
\cline{2-16} \cline{3-16}
 & 1.00 & 269 & 0.528 & 0.79 & \mycolorbox{paleblue}{0.96$\pm$0.01} & 0.97 & 0.17 & 0.01 & 0.18 & 0.73 & \mycolorbox{paleblue}{0.96$\pm$0.01} & 0.97 & 0.23 & 0.01 & 0.24 \\
\cline{2-16} \cline{3-16}
 & All & 500 & 0.528 & 0.75 & \mycolorbox{paleblue}{0.86$\pm$0.02} & 0.86 & 0.11 & -0.00 & 0.11 & 0.66 & \mycolorbox{paleblue}{0.86$\pm$0.01} & 0.85 & 0.20 & -0.01 & 0.19 \\
\thickhline
\multirow[c]{4}{*}{\shortstack[l]{Pleural\\Effusion}} & 0.60 & 83 & 0.422 & 0.59 & 0.61$\pm$0.08 & 0.56 & 0.02 & -0.05 & -0.03 & 0.42 & 0.62$\pm$0.11 & 0.60 & 0.20 & -0.02 & 0.18 \\
\cline{2-16} \cline{3-16}
 & 0.80 & 110 & 0.336 & 0.54 & 0.78$\pm$0.07 & 0.73 & 0.24 & -0.05 & 0.19 & 0.44 & \mycolorbox{paleblue}{0.82$\pm$0.07} & 0.80 & 0.38 & -0.02 & 0.36 \\
\cline{2-16} \cline{3-16}
 & 1.00 & 307 & 0.104 & 0.28 & \mycolorbox{paleblue}{0.83$\pm$0.08} & 0.87 & 0.55 & 0.04 & 0.59 & 0.48 & \mycolorbox{paleblue}{0.96$\pm$0.02} & 0.97 & 0.48 & 0.01 & 0.49 \\
\cline{2-16} \cline{3-16}
 & All & 500 & 0.208 & 0.43 & 0.74$\pm$0.08 & 0.72 & 0.31 & -0.02 & 0.29 & 0.46 & \mycolorbox{paleblue}{0.87$\pm$0.05} & 0.87 & 0.41 & -0.00 & 0.41 \\
\thickhline
\multirow[c]{4}{*}{Pneumonia} & 0.60 & 24 & 0.333 & 0.38 & 0.22$\pm$0.08 & 0.50 & -0.16 & 0.28 & 0.12 & 0.33 & 0.60$\pm$0.11 & 0.60 & 0.27 & 0.00 & 0.27 \\
\cline{2-16} \cline{3-16}
 & 0.80 & 82 & 0.012 & 0.03 & 0.30$\pm$0.17 & 0.09 & 0.27 & -0.21 & 0.06 & 0.30 & \mycolorbox{paleblue}{0.93$\pm$0.05} & 0.80 & 0.63 & -0.13 & 0.50 \\
\cline{2-16} \cline{3-16}
 & 1.00 & 394 & 0.005 & 0.02 & 0.34$\pm$0.29 & 0.25 & 0.32 & -0.09 & 0.23 & 0.68 & \mycolorbox{paleblue}{0.98$\pm$0.01} & 0.97 & 0.30 & -0.01 & 0.29 \\
\cline{2-16} \cline{3-16}
 & All & 500 & 0.022 & 0.07 & 0.27$\pm$0.13 & 0.29 & 0.20 & 0.02 & 0.22 & 0.60 & \mycolorbox{paleblue}{0.95$\pm$0.02} & 0.92 & 0.35 & -0.03 & 0.32 \\
\thickhline
\multirow[c]{4}{*}{\shortstack[l]{Support\\Devices}} & 0.60 & 50 & 0.580 & 0.73 & 0.72$\pm$0.04 & 0.64 & -0.01 & -0.08 & -0.09 & 0.58 & 0.67$\pm$0.05 & 0.60 & 0.09 & -0.07 & 0.02 \\
\cline{2-16} \cline{3-16}
 & 0.80 & 105 & 0.705 & \mycolorbox{paleblue}{0.84} & \mycolorbox{paleblue}{0.93$\pm$0.02} & 0.85 & 0.09 & -0.08 & 0.01 & 0.74 & \mycolorbox{paleblue}{0.90$\pm$0.02} & 0.80 & 0.16 & -0.10 & 0.06 \\
\cline{2-16} \cline{3-16}
 & 1.00 & 345 & 0.458 & 0.73 & \mycolorbox{paleblue}{0.98$\pm$0.00} & 0.97 & 0.25 & -0.01 & 0.24 & 0.66 & \mycolorbox{paleblue}{0.98$\pm$0.00} & 0.97 & 0.32 & -0.01 & 0.31 \\
\cline{2-16} \cline{3-16}
 & All & 500 & 0.522 & 0.76 & \mycolorbox{paleblue}{0.93$\pm$0.01} & 0.90 & 0.17 & -0.03 & 0.14 & 0.67 & \mycolorbox{paleblue}{0.93$\pm$0.01} & 0.90 & 0.26 & -0.03 & 0.23 \\
\thickhline

    \end{tabular}
    }
    \caption{Performance on CheXpert using Gemini-2.5 pro}
    \label{tab:appendix:gemini25pro}
\end{table}

\begin{table}[h!]
    \setlength{\tabcolsep}{4pt}
    \centering
    \resizebox{1.0\columnwidth}{!}
    {
\begin{tabular}{p{0.1\linewidth}rrr|rrrrrr|rrrrrr}
\toprule
  &  &  &  & \multicolumn{6}{c|}{F1} & \multicolumn{6}{c}{Accuracy}\\
 \hline
 Pathology & $p_d$ & S & $m$  & AI & $\text{H}$\ \   & $E$  & $\Delta_{\text{H}-AI}$ & $\Delta_{E-\text{H}}$ & $\Delta_{E-M}$ & AI & $\text{H}$\ \ \  &   $E$  & $\Delta_{\text{H}-AI}$ & $\Delta_{E-\text{H}}$ & $\Delta_{E-M}$\\
\thickhline
\multirow[c]{4}{*}{Atelectasis} & 0.60 & 130 & 0.500 & 0.49 & 0.66$\pm$0.03 & 0.60 & 0.17 & -0.06 & 0.11 & 0.54 & 0.60$\pm$0.02 & 0.60 & 0.06 & -0.00 & 0.06 \\
\cline{2-16} \cline{3-16}
 & 0.80 & 140 & 0.400 & 0.64 & 0.70$\pm$0.03 & 0.76 & 0.06 & 0.06 & 0.12 & 0.74 & 0.73$\pm$0.03 & 0.80 & -0.01 & 0.07 & 0.06 \\
\cline{2-16} \cline{3-16}
 & 1.00 & 230 & 0.139 & 0.70 & 0.74$\pm$0.03 & 0.90 & 0.04 & 0.16 & 0.20 & \mycolorbox{paleblue}{0.93} & \mycolorbox{paleblue}{0.91$\pm$0.02} & 0.97 & -0.02 & 0.06 & 0.04 \\
\cline{2-16} \cline{3-16}
 & All & 500 & 0.306 & 0.59 & 0.69$\pm$0.02 & 0.73 & 0.10 & 0.04 & 0.14 & 0.77 & 0.78$\pm$0.02 & 0.83 & 0.01 & 0.05 & 0.06 \\
\thickhline
\multirow[c]{4}{*}{\shortstack[l]{Cardio-\\megaly}} & 0.60 & 142 & 0.507 & 0.59 & 0.64$\pm$0.10 & 0.60 & 0.05 & -0.04 & 0.01 & 0.56 & 0.66$\pm$0.03 & 0.60 & 0.10 & -0.06 & 0.04 \\
\cline{2-16} \cline{3-16}
 & 0.80 & 196 & 0.372 & 0.61 & 0.71$\pm$0.04 & 0.75 & 0.10 & 0.04 & 0.14 & 0.71 & 0.79$\pm$0.03 & 0.80 & 0.08 & 0.01 & 0.09 \\
\cline{2-16} \cline{3-16}
 & 1.00 & 162 & 0.037 & 0.33 & 0.70$\pm$0.12 & 0.71 & 0.37 & 0.01 & 0.38 & \mycolorbox{paleblue}{0.90} & \mycolorbox{paleblue}{0.98$\pm$0.02} & 0.97 & 0.08 & -0.01 & 0.07 \\
\cline{2-16} \cline{3-16}
 & All & 500 & 0.302 & 0.58 & 0.68$\pm$0.07 & 0.68 & 0.10 & 0.00 & 0.10 & 0.73 & \mycolorbox{paleblue}{0.81$\pm$0.03} & 0.80 & 0.08 & -0.01 & 0.07 \\
\thickhline
\multirow[c]{4}{*}{\shortstack[l]{Consoli- \\dation}} & 0.60 & 81 & 0.272 & 0.43 & 0.44$\pm$0.07 & 0.45 & 0.01 & 0.01 & 0.02 & 0.35 & 0.66$\pm$0.06 & 0.60 & 0.31 & -0.06 & 0.25 \\
\cline{2-16} \cline{3-16}
 & 0.80 & 89 & 0.067 & 0.16 & 0.36$\pm$0.17 & 0.35 & 0.20 & -0.01 & 0.19 & 0.30 & \mycolorbox{paleblue}{0.82$\pm$0.11} & 0.80 & 0.52 & -0.02 & 0.50 \\
\cline{2-16} \cline{3-16}
 & 1.00 & 330 & 0.003 & 0.00 & 0.23$\pm$0.21 & 0.16 & 0.23 & -0.07 & 0.16 & 0.75 & \mycolorbox{paleblue}{0.98$\pm$0.01} & 0.97 & 0.23 & -0.01 & 0.22 \\
\cline{2-16} \cline{3-16}
 & All & 500 & 0.058 & 0.21 & 0.39$\pm$0.06 & 0.39 & 0.18 & 0.00 & 0.18 & 0.61 & \mycolorbox{paleblue}{0.90$\pm$0.03} & 0.88 & 0.29 & -0.02 & 0.27 \\
\thickhline
\multirow[c]{4}{*}{Edema} & 0.60 & 76 & 0.355 & 0.48 & 0.50$\pm$0.03 & 0.52 & 0.02 & 0.02 & 0.04 & 0.55 & 0.60$\pm$0.09 & 0.60 & 0.05 & -0.00 & 0.05 \\
\cline{2-16} \cline{3-16}
 & 0.80 & 116 & 0.233 & 0.64 & 0.57$\pm$0.07 & 0.65 & -0.07 & 0.08 & 0.01 & 0.79 & 0.76$\pm$0.07 & 0.80 & -0.03 & 0.04 & 0.01 \\
\cline{2-16} \cline{3-16}
 & 1.00 & 308 & 0.078 & 0.63 & 0.70$\pm$0.07 & 0.83 & 0.07 & 0.13 & 0.20 & \mycolorbox{paleblue}{0.93} & \mycolorbox{paleblue}{0.94$\pm$0.02} & 0.97 & 0.01 & 0.03 & 0.04 \\
\cline{2-16} \cline{3-16}
 & All & 500 & 0.156 & 0.58 & 0.58$\pm$0.02 & 0.66 & 0.00 & 0.08 & 0.08 & \mycolorbox{paleblue}{0.84} & \mycolorbox{paleblue}{0.85$\pm$0.03} & 0.87 & 0.01 & 0.02 & 0.03 \\
\thickhline
\multirow[c]{4}{*}{\shortstack[l]{Enlarged \\Cardiomed- \\iastinum}} & 0.60 & 163 & 0.417 & 0.44 & 0.51$\pm$0.06 & 0.56 & 0.07 & 0.05 & 0.12 & 0.61 & 0.66$\pm$0.02 & 0.60 & 0.05 & -0.06 & -0.01 \\
\cline{2-16} \cline{3-16}
 & 0.80 & 170 & 0.529 & 0.66 & 0.75$\pm$0.04 & 0.81 & 0.09 & 0.06 & 0.15 & 0.71 & 0.78$\pm$0.02 & 0.80 & 0.07 & 0.02 & 0.09 \\
\cline{2-16} \cline{3-16}
 & 1.00 & 167 & 0.569 & \mycolorbox{paleblue}{0.84} & \mycolorbox{paleblue}{0.92$\pm$0.03} & 0.97 & 0.08 & 0.05 & 0.13 & \mycolorbox{paleblue}{0.84} & \mycolorbox{paleblue}{0.91$\pm$0.03} & 0.97 & 0.07 & 0.06 & 0.13 \\
\cline{2-16} \cline{3-16}
 & All & 500 & 0.506 & 0.67 & 0.75$\pm$0.03 & 0.80 & 0.08 & 0.05 & 0.13 & 0.72 & 0.79$\pm$0.02 & 0.79 & 0.07 & 0.00 & 0.07 \\
\thickhline
\multirow[c]{4}{*}{\shortstack[l]{Lung\\Opacity}} & 0.60 & 106 & 0.519 & 0.66 & 0.63$\pm$0.07 & 0.61 & -0.03 & -0.02 & -0.05 & 0.58 & 0.62$\pm$0.03 & 0.60 & 0.04 & -0.02 & 0.02 \\
\cline{2-16} \cline{3-16}
 & 0.80 & 125 & 0.536 & \mycolorbox{paleblue}{0.82} & \mycolorbox{paleblue}{0.84$\pm$0.03} & 0.81 & 0.02 & -0.03 & -0.01 & 0.79 & \mycolorbox{paleblue}{0.83$\pm$0.03} & 0.80 & 0.04 & -0.03 & 0.01 \\
\cline{2-16} \cline{3-16}
 & 1.00 & 269 & 0.528 & \mycolorbox{paleblue}{0.90} & \mycolorbox{paleblue}{0.96$\pm$0.01} & 0.97 & 0.06 & 0.01 & 0.07 & \mycolorbox{paleblue}{0.89} & \mycolorbox{paleblue}{0.96$\pm$0.01} & 0.97 & 0.07 & 0.01 & 0.08 \\
\cline{2-16} \cline{3-16}
 & All & 500 & 0.528 & \mycolorbox{paleblue}{0.83} & \mycolorbox{paleblue}{0.86$\pm$0.02} & 0.86 & 0.03 & -0.00 & 0.03 & \mycolorbox{paleblue}{0.80} & \mycolorbox{paleblue}{0.86$\pm$0.01} & 0.85 & 0.06 & -0.01 & 0.05 \\
\thickhline
\multirow[c]{4}{*}{\shortstack[l]{Pleural\\Effusion}} & 0.60 & 83 & 0.422 & 0.51 & 0.61$\pm$0.08 & 0.56 & 0.10 & -0.05 & 0.05 & 0.45 & 0.62$\pm$0.11 & 0.60 & 0.17 & -0.02 & 0.15 \\
\cline{2-16} \cline{3-16}
 & 0.80 & 110 & 0.336 & 0.63 & 0.78$\pm$0.07 & 0.73 & 0.15 & -0.05 & 0.10 & 0.65 & \mycolorbox{paleblue}{0.82$\pm$0.07} & 0.80 & 0.17 & -0.02 & 0.15 \\
\cline{2-16} \cline{3-16}
 & 1.00 & 307 & 0.104 & 0.55 & \mycolorbox{paleblue}{0.83$\pm$0.08} & 0.87 & 0.28 & 0.04 & 0.32 & \mycolorbox{paleblue}{0.85} & \mycolorbox{paleblue}{0.96$\pm$0.02} & 0.97 & 0.11 & 0.01 & 0.12 \\
\cline{2-16} \cline{3-16}
 & All & 500 & 0.208 & 0.56 & 0.74$\pm$0.08 & 0.72 & 0.18 & -0.02 & 0.16 & 0.74 & \mycolorbox{paleblue}{0.87$\pm$0.05} & 0.87 & 0.13 & -0.00 & 0.13 \\
\thickhline
\multirow[c]{4}{*}{Pneumonia} & 0.60 & 24 & 0.333 & 0.43 & 0.22$\pm$0.08 & 0.50 & -0.21 & 0.28 & 0.07 & 0.33 & 0.60$\pm$0.11 & 0.60 & 0.27 & 0.00 & 0.27 \\
\cline{2-16} \cline{3-16}
 & 0.80 & 82 & 0.012 & 0.03 & 0.30$\pm$0.17 & 0.09 & 0.27 & -0.21 & 0.06 & 0.26 & \mycolorbox{paleblue}{0.93$\pm$0.05} & 0.80 & 0.67 & -0.13 & 0.54 \\
\cline{2-16} \cline{3-16}
 & 1.00 & 394 & 0.005 & 0.02 & 0.34$\pm$0.29 & 0.25 & 0.32 & -0.09 & 0.23 & 0.67 & \mycolorbox{paleblue}{0.98$\pm$0.01} & 0.97 & 0.31 & -0.01 & 0.30 \\
\cline{2-16} \cline{3-16}
 & All & 500 & 0.022 & 0.07 & 0.27$\pm$0.13 & 0.29 & 0.20 & 0.02 & 0.22 & 0.59 & \mycolorbox{paleblue}{0.95$\pm$0.02} & 0.92 & 0.36 & -0.03 & 0.33 \\
\thickhline
\multirow[c]{4}{*}{\shortstack[l]{Support\\Devices}} & 0.60 & 50 & 0.580 & 0.74 & 0.72$\pm$0.04 & 0.64 & -0.02 & -0.08 & -0.10 & 0.60 & 0.67$\pm$0.05 & 0.60 & 0.07 & -0.07 & 0.00 \\
\cline{2-16} \cline{3-16}
 & 0.80 & 105 & 0.705 & \mycolorbox{paleblue}{0.85} & \mycolorbox{paleblue}{0.93$\pm$0.02} & 0.85 & 0.08 & -0.08 & 0.00 & 0.76 & \mycolorbox{paleblue}{0.90$\pm$0.02} & 0.80 & 0.14 & -0.10 & 0.04 \\
\cline{2-16} \cline{3-16}
 & 1.00 & 345 & 0.458 & 0.76 & \mycolorbox{paleblue}{0.98$\pm$0.00} & 0.97 & 0.22 & -0.01 & 0.21 & 0.72 & \mycolorbox{paleblue}{0.98$\pm$0.00} & 0.97 & 0.26 & -0.01 & 0.25 \\
\cline{2-16} \cline{3-16}
 & All & 500 & 0.522 & 0.78 & \mycolorbox{paleblue}{0.93$\pm$0.01} & 0.90 & 0.15 & -0.03 & 0.12 & 0.72 & \mycolorbox{paleblue}{0.93$\pm$0.01} & 0.90 & 0.21 & -0.03 & 0.18 \\
\thickhline

    \end{tabular}
    }
    \caption{Performance of GPT-5.1 on CheXpert }
    \label{tab:gpt5.1}
\end{table}

\end{appendices}

\clearpage
\end{document}